\documentclass[10pt,twocolumn,letterpaper]{article}

\usepackage[pagenumbers]{cvpr} 

\usepackage[normalem]{ulem}

\definecolor{cvprblue}{rgb}{0.21,0.49,0.74}
\usepackage[pagebackref,breaklinks,colorlinks,allcolors=cvprblue]{hyperref}
\usepackage{amsthm}
\usepackage{algorithm}
\usepackage{algpseudocode}
\usepackage{comment}
\usepackage{xcolor}
\usepackage{array}
\usepackage{adjustbox}
\usepackage{booktabs}      
\usepackage{graphicx}      
\usepackage{adjustbox}     
\usepackage{makecell} 
\usepackage{multirow}
\usepackage[table]{xcolor}

\usepackage{cuted}

\newtheorem{theorem}{Theorem}[section]

\theoremstyle{definition}

\def\paperID{2394} 
\def\confName{CVPR}
\def\confYear{2026}

\title{
Learning Eigenstructures of Unstructured Data Manifolds
}

\author{
    Roy Velich$^1$\thanks{Corresponding author: \tt{royve@campus.technion.ac.il}} \;
    Arkadi Piven$^1$ \;
    David Bensa\"id$^1$ \; 
    Daniel Cremers$^{2,3}$ \;
    Thomas Dagès$^{1,2,3}$ \;
    Ron Kimmel$^1$
    \\[1em] 
    \textit{In memory of Haïm Brezis (1944 -- 2024).}
    \vspace{-1em}
}

\begin{document}
\maketitle
\footnotetext[1]{Technion - Israel Institute of Technology}
\footnotetext[2]{Technical University of Munich}
\footnotetext[3]{Munich Center for Machine Learning}

\begin{abstract}

We introduce a novel framework that directly learns a spectral basis for shape and manifold analysis from unstructured data, eliminating the need for traditional operator selection, discretization, and eigensolvers.
Grounded in optimal-approximation theory, we train a network to decompose an implicit approximation operator by minimizing the reconstruction error in the learned basis over a chosen distribution of probe functions. 
For suitable distributions, they can be seen as an approximation of the Laplacian operator and its eigendecomposition, which are fundamental in geometry processing.  
Furthermore, our method recovers in a unified manner not only the spectral basis, but also the implicit metric's sampling density and the eigenvalues of the underlying operator. 
Notably, our unsupervised method makes no assumption on the data manifold, such as meshing or manifold dimensionality, allowing it to scale to arbitrary datasets of any dimension.
On point clouds lying on surfaces in 3D and high-dimensional image manifolds, our approach yields meaningful spectral bases, that can resemble those of the Laplacian, without explicit construction of an operator. 
By replacing the traditional operator selection, construction, and eigendecomposition with a learning-based approach, our framework offers a principled, data-driven alternative to conventional pipelines. This opens new possibilities in geometry processing for unstructured data, particularly in high-dimensional spaces.

\end{abstract}

\section{Introduction}
\label{sec:intro}

Differential geometry is at the core of shape and manifold analysis. 
By defining operators, one can encode the underlying geometry and compute intrinsic and extrinsic quantities, from geodesics and Gaussian curvature to mean curvature and normals. 
The specific choice of operator depends on the task at hand. 
For example, the Laplace-Beltrami operator (LBO)  
captures intrinsic geometric properties of the underlying manifold and governs heat diffusion.
Hamiltonian operators, appearing in the famous Schr\"{o}dinger equation, model wave motion and allow potential-modulated particle dynamics. 
The biharmonic operator encodes fourth-order behavior and is natural for smooth interpolation and shape-aware distances.

In practice, these operators are primarily used for their spectral decompositions, which serve as the computational foundation for geometry processing.
For instance, LBO eigenvalues are intrinsic shape signatures~\cite{reuter2006laplace, kac1966can, protter1987can, gordon1992one, gordon1996you}, while LBO eigenfunctions power heat- and wave-kernel methods for smoothing and propagation~\cite{coifman2006diffusion, hammond2011wavelets}, define feature descriptors~\cite{rustamov2007laplace,sun2009concise,aubry2011wave}, and support matching via functional maps \cite{ovsjanikov2012functional}. 
As such, these eigendecompositions are routinely required in geometry processing.

The standard pipeline for computing 
operator eigendecompositions on discrete data begins by choosing a discrete approximation that preserves key geometric properties such as positive semi-definiteness (PSD), symmetry, consistency, or weak formulations 
\cite{wardetzky2007discrete, pinkall1993computing, sharp2020laplacian}.
Then, explicitly construct a matrix acting as the discretized linear operator and solve a generalized eigenvalue problem using a classical numerical solver. 
Although effective on data sampled from surfaces in 3D, this pipeline
designed for intrinsically two-dimensional manifolds does not scale gracefully to high-dimensional manifolds~\cite{sharp2020laplacian}.

Here, we propose a different route. 
Rather than targeting a specific operator, explicitly discretizing and eigendecomposing it, we learn the spectral decomposition directly from the data. 
The learned spectral decomposition implicitly corresponds to an operator that is reconstructible a posteriori from the basis and associated eigenvalues. 
Building on optimal-approximation theory \cite{aflalo2016best, brezis2017rigidity,aflalo2015optimality}, we find the eigenbasis of an optimal-approximation operator, that minimizes the reconstruction error over a class of probe functions. 
Different probe classes induce different metrics and thus different operators. 
In particular cases, the learned operator approximates the LBO, but the proposed framework generalizes to more operators.
While only the eigenbasis is learned, the eigenvalues are given directly as a by-product of the worst-case reconstruction error. 
This yields a learning methodology for spectral analysis that is data-driven and avoids the need for explicit operator construction.

To summarize, our contributions are as follows:
\begin{itemize}
    \item We learn spectral bases directly from point clouds, bypassing the explicit choice of an operator, its construction, and classical numerical eigensolvers.
    We ground the method in an optimal-basis 
    representation theory and use it to propose
    a unique learning objective.
    \item We demonstrate on surfaces in 3D that the proposed approach provides eigenstructures and estimated metrics performing competitively with oracle discrete LBO baselines, while bypassing explicit operator construction.
    \item We show that the method scales from surfaces to higher-dimensional data manifolds, enabling scalable manifold learning where mesh-based pipelines are inapplicable and common graph-based ones are unreliable.
    \item We will release our code as open source upon acceptance.
\end{itemize}

\section{Related Works}  
\label{sec:related}

Our framework lies at the intersection of operator discretization and neural methods for eigenproblems.

\hfill \break
\noindent \textbf{Discrete operators.}
In geometry processing, discrete operators supply via spectral decomposition \cite{cauchy1829equation} the orthonormal basis that we actually use to process signals on meshes and graphs. 
For instance, filtering \cite{lescoat2020spectral, bensaid2023partial}, diffusion \cite{coifman2006diffusion, sharp2022diffusionnet}, and convolution \cite{bruna2013spectral, levie2021transferability} are usually implemented in this spectral basis.
Notably, the community rarely applies the operator matrix directly.
In practice, only the first eigenvectors are kept, which amounts to projecting a signal onto the lowest-energy subspace defined by the operator.
This acts as a low-pass filter where only the (smooth) low-frequency components are retained.
Because the operator defines what ``energy'', ``smoothness'', and ``frequency'' mean, different operators emphasize different properties of the manifold.
This motivates the importance of choosing operators wisely in the field of geometry processing.

Perhaps the most important example,
the LBO \cite{beltrami1868saggio,rosenberg1997laplacian} is ubiquitous in geometry processing \cite{bronstein2008numerical,wetzler2013laplace,sun2009concise,zhou2025laplace,bracha2020shape,chen2024learning,bracha2024unsupervised,bracha2024wormhole}. Discretizing the weak form gives the cotangent Laplacian \cite{pinkall1993computing,meyer2003visualization} -- the reference in LBO discretization, but alternatives exist \cite{chuang2009estimating,zhang2025topology}. 
Cotangent weights need well-triangulated meshes, i.e.\ Delaunay, to avoid negative edges violating the maximum principle \cite{wardetzky2007discrete}. 
They cannot be directly applied to unstructured data, like point clouds, requiring first wise meshings, e.g.\ with the tufted cover of the Robust Laplacian \cite{sharp2020laplacian} enabling flips to Delaunay triangulations.
While it extends to thin 3D volumes, this approach does not scale beyond surfaces. 
For higher-dimensional manifolds, graph Laplacians \cite{kirchhoff1847ueber,chung1997spectral} are used, but they depend strongly on connectivity, e.g.\ local density, unlike the targeted smooth LBO \cite{levy2006laplace}. 
Constructing a reliable high-dimensional generalization of the LBO is a challenge.

Learning Laplacians has become a popular line of research as the amount of data increases.
Some works suggest learning the operator action \cite{quackenbush2024geometric} but lack eigendecompositions.
Others learn eigenvalues but not eigenvectors \cite{an2025ai}. 
Most learn a Laplacian matrix from data rather than heuristics \cite{pang2024neural, yigit2025lbonet}, but they still explicitly approximate the operator by assembling mass and stiffness matrices and then apply eigensolvers sensitive to these approximations. 
As they rely on triangle-based discretizations, they target surfaces and do not extend to higher-dimensionality. 
For implicit neural representations, the Laplacian can be built via Rayleigh quotients directly \cite{williamson2025neural}. 
However, this assumes a two-dimensional surface and the Euclidean ambient metric, removing adaptivity to other metrics or dimensionalities.
On the theoretical front, Laplacian estimation has progressed \cite{chazal2016data,peoples2024higher,trillos2025minimax,mhaskar2025learning}, yet, translating these insights into practice remains largely unexplored.

Additionally, other discrete operators provide different insights. 
Changing the metric, via scaling (scale-invariant LBO \cite{bracha2020shape,aflalo2013scale,sela2015computational,halimi2018self}), anisotropy \cite{andreux2014anisotropic,boscaini2016anisotropic,boscaini2016learning,rosenberg1997laplacian,raviv2015affine}, or asymmetry \cite{weber2024finsler,dages2025finsler,dages2025metric,barthelme2013natural,ohta2009heat}, changes the LBO and its approximation. 
Beyond LBOs, both intrinsic \cite{rampini2019correspondence,bensaid2023hamiltonian, choukroun2018sparse} and extrinsic \cite{wang2018steklov} alternatives are understudied.


\hfill \break
\noindent \textbf{Eigenproblems and neural networks.}
Recent efforts apply neural networks to operator eigenvalue problems, typically using variational or dynamical formulations. 
In \cite{rowan2025solving}, networks are trained to represent individual Laplacian eigenfunctions as continuous functions on parametrized domains, based on the Rayleigh quotient objective and finding eigenfunctions sequentially via Gram-Schmidt orthogonalization, an idea that can be extended to learning multiple eigenfunctions simultaneously \cite{benshaul2023deep}.
However, this approach is limited to 
simplistic impractical domains, e.g.\ Euclidean.
Another approach reformulates the eigenvalue problem as a fixed point of the operator's semi-group flow, training networks via forward-backward stochastic differential equations to handle high-dimensional problems, up to ten dimensions \cite{han2020solving}.
These methods fundamentally differ from the proposed approach: (1) they require explicit domain parameterization with global coordinates, (2) they assume flat geometry by taking Euclidean gradients via automatic differentiation, and (3) they must be trained from scratch for each specific domain, with no mechanism for generalization across different geometries.
Thus, they cannot directly handle curved manifolds sampled as point clouds without manually specifying metric tensors and managing coordinate singularities.

\section{Method}
\label{sec:method}

\begin{figure*}[t]
    \centering
    \includegraphics[width=\textwidth]{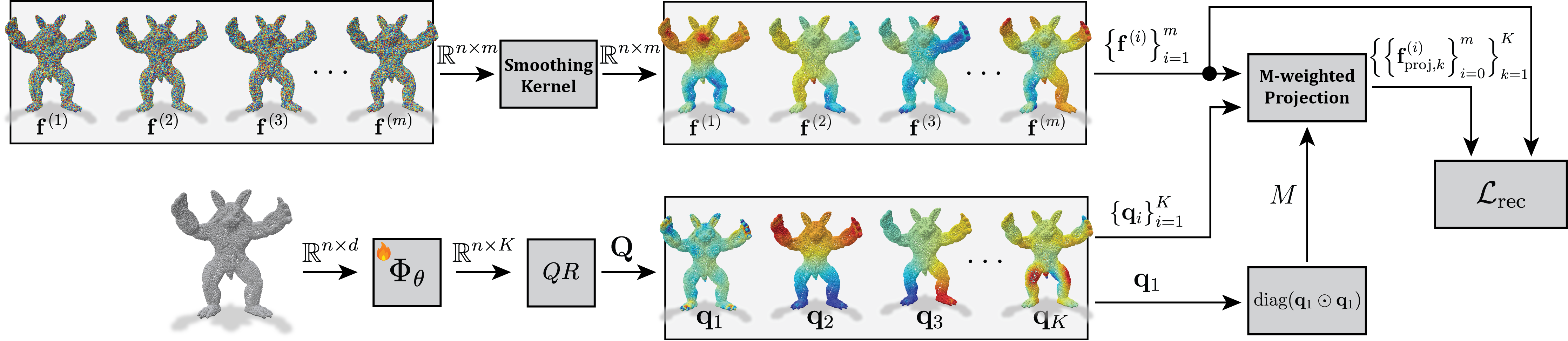}
    \caption{
    Overview of our neural framework to compute spectral bases directly from unstructured point clouds of any dimensionality, based on optimal-approximation theory, without first explicitly choosing, discretely approximating, and eigendecomposing an operator. 
    }
    \label{fig: framework overview}
    \vspace{-1em}
\end{figure*}

\subsection{Foundations}

Denote $\mathcal{M} \subset \mathbb{R}^d$ as a high-dimensional manifold, typically a curved low-dimensional surface embedded in $\mathbb{R}^d$, with functions $f\in\mathcal{F}(\mathcal{M},\mathbb{R})$ and linear operators $L:\mathcal{F}(\mathcal{M},\mathbb{R})\to\mathcal{F}(\mathcal{M},\mathbb{R})$ acting on them. 
When the manifold is discretely sampled by $n$ points, scalar functions can be represented by vectors $f\in \mathbb{R}^n$ and operators as matrices $L\in\mathbb{R}^{n\times n}$. 
Here, we denote by $\langle \cdot, \cdot \rangle_L$ the $L$-weighted inner product $\langle f, g \rangle_L = f^\top L g$ for any {\it symmetric positive definite} (SPD) matrix $L$, with its induced norm $\|f\|_L = \sqrt{\langle f, f \rangle_L} = \sqrt{f^\top L f}$.

\hfill \break
\noindent \textbf{Optimal-approximation theory.}
Given a class $\mathcal{C}$ of signals on the discrete domain $\mathcal{M}$, a fundamental question arises: What is the optimal orthonormal basis for representing functions  $f\in \mathcal{C}$? 
The answer depends on how we characterize the class of signals. 
A key insight \cite{aflalo2016best,brezis2017rigidity} is that many practical signal classes can be characterized by constraints of the form $\|f\|_L \leq 1$, where $L$ is an SPD operator encoding prior knowledge about the signals. 
Remarkably, for any such constraint class, the optimal basis is given by the eigenvectors of $L$ itself.

\begin{theorem}[Min-Max Optimality {\cite[Theorem~2.1]{aflalo2016best}}]
    \label{thm:aflalo_minmax}
    Given a symmetric positive definite operator $L$  with eigenvalues $0 < \lambda_1 \leq \cdots \leq \lambda_n$ and eigenvectors $e_1, \ldots, e_n$, the min-max approximation error
    \begin{eqnarray*}  
        \alpha_k &=& \min_{b=(b_1,\ldots,b_n)} \max_{\|f\|_L \leq 1} \left\|f - \sum_{i=1}^k \langle f, b_i \rangle_L b_i\right\|_L^2,
   \end{eqnarray*}
    where $b$ ranges over orthonormal bases, is minimized by the first $k$ eigenvectors of $L$, i.e.\ $b_i=e_i\;\forall i\le k$, with optimal value $\lambda_{k+1} = \tfrac{1}{\alpha_k}$.
    For simple spectrum, $\lambda_i < \lambda_{i+1} \;\forall i$, the basis $b$ that is a solution for every $k$ is unique up to signs.
\end{theorem}

The key insight is that for any SPD operator $L \in \mathbb{R}^{n \times n}$, the optimal orthonormal basis for the progressive $k$-term approximation, i.e.\ for every $k$, is uniquely given by the eigenvectors of $L$ ordered by increasing eigenvalues. Crucially, the worst approximation error using the first $k$ eigenvectors is $\tfrac{1}{\lambda_{k+1}}$, meaning that the $(k+1)$-th eigenvalue is inversely proportional to the worst maximum reconstruction error.

The min-max formulation (\cref{thm:aflalo_minmax}) optimizes over the set $\mathcal{C}_L = \{f \,;\, \|f\|_L \leq 1\}$.
Instead of minimizing the worst-case reconstruction error, another natural approach to find optimal representations would be to maximize the captured variance. 
This alternative problem leads to Principal Component Analysis (PCA) on the same class of signals $\mathcal{C}_L$.
Remarkably, we obtain the following result when performing PCA on uniformly distributed signals from $\mathcal{C}_L$.

\begin{theorem}[Operator-Bounded PCA {\cite[Section~5]{aflalo2016best}}]
    \label{thm:pca_continuous}
    Given a symmetric positive definite operator $L$ 
    with eigenvalues $0 < \lambda_1 \leq \cdots \leq \lambda_n$, and eigenvectors $e_1, \ldots, e_n$,
    the PCA objective
    \begin{eqnarray*} 
        \min\limits_{b = (b_1,\cdots, b_n)} \mathbb{E}_{f\sim \mathcal{U}\big(\lVert f\rVert_L \le 1\big)}\left(\left\lVert f - \sum\limits_{i=1}^k\langle f,b_i\rangle_L b_i\right\rVert_L^2\right),
    \end{eqnarray*}
    over orthonormal bases $b$ is minimized by the eigenvectors
    of the covariance matrix $R_L=\mathbb{E}_{f\sim\mathcal{U}\big(\{\lVert f \rVert_L\le 1\}\big)}[ff^\top]$, 
    which are the first $k$ eigenvectors of $L$, with eigenvalues (variances) $\lambda_1^{-1} \ge \cdots \ge\lambda_n^{-1}$. In other words:
    $R_L = L^{-1}$.
\end{theorem}
For a reminder why the expectation of the approximation error is PCA, i.e.\ iterative maximisation of the Rayleigh quotient of $R_L$, see \cref{sec: pca alternative formulations}.
Therefore, PCA on $\mathcal{C}_L$
yields the same eigenvectors and order as the min-max solution.
The min-max optimization (\cref{thm:aflalo_minmax}) and its PCA counterpart (\cref{thm:pca_continuous}) are thus equivalent when applied to the same signal class.
These dual formulations have useful practical implications that we exploit in our method.

A natural prior for signals $f$ on a manifold is smoothness, implying that the Dirichlet energy $\lVert \nabla f \rVert_2$ is bounded. 
By Green's identity, $\lVert \nabla f\rVert_2 = \lVert f\rVert_\Delta$, where $\Delta$ is the (discrete) Laplace-Beltrami operator (LBO). 
Thus, bounding the Dirichlet energy leads to choosing $\Delta$ for $L$.
To motivate our method, let us thus focus on the Laplacian operator and its construction. This will enable the derivation of our framework, generalizable both beyond Laplacian operators and to arbitrarily high-dimensional manifolds, from surfaces in $\mathbb{R}^3$ to image datasets.

\hfill \break
\noindent \textbf{Discrete Laplacians.}
The discrete Laplacian is traditionally constructed from two fundamental matrices, the mass matrix $M \in \mathbb{R}^{n \times n}$, which is a positive diagonal matrix encoding the local (metric-dependent) sampling density defining the manifold's Riemannian metric, and the stiffness matrix $S \in \mathbb{R}^{n \times n}$ encoding geometric relationships that discretize the continuous Laplace-Beltrami operator. 
For two-dimensional surfaces, $M$ represents vertex areas and $S$ contains cotangent weights \cite{sorkine2004laplacian, botsch2010polygon, andreux2014anisotropic}. 
For higher-dimensionality, the LBO is harder to approximate, thus $M$ and $S$ are constructed with schemes more sensitive to the connectivity of the relationship graph.
In any case, the matrices $M$ and $S$ enable two common formulations for the discretized Laplacian: the unnormalized Laplacian $L = M^{-1}S$ and the symmetric normalized Laplacian $L_{\text{norm}} = M^{-\frac{1}{2}}SM^{-\frac{1}{2}}$.
While both operators share identical eigenvalues, their eigenvectors differ.
The unnormalized Laplacian's eigenvectors $\mathbf{v}_i \in \mathbb{R}^n$ are $M$-orthogonal, satisfying 
$\langle\mathbf{v}_i, \mathbf{v}_j\rangle_M = \delta_{ij}$,
where $\delta_{ij}=1$ if $i=j$ and $0$ otherwise, whereas the normalized Laplacian's eigenvectors $\mathbf{q}_i \in \mathbb{R}^n$ are Euclidean-orthogonal, with $\langle\mathbf{q}_i, \mathbf{q}_j\rangle = \delta_{ij}$. 
Both sets of eigenvectors are related by $\mathbf{v}_i = M^{-\frac{1}{2}}\mathbf{q}_i$, as $L(M^{-\frac{1}{2}}\mathbf{q}_i) = M^{-\frac{1}{2}}M^{-\frac{1}{2}}S M^{-\frac{1}{2}} \mathbf{q}_i = \lambda_i M^{-\frac{1}{2}}\mathbf{q}_i$. 

On closed connected manifolds\footnote{
This setting is standard. It is both common in practice for meshes, and is systematic for unstructured data like pointclouds, which is what we focus on, where potential manifold boundaries are both ill-defined and not provided. The optimal approximation theory generalizes to this case by focusing on the orthogonal of the nullspace.}, these operators' null-spaces reveal important differences.
While the vector of ones $\mathbf{1}$ lies in the nullspace of $L$, the vector $M^{\frac{1}{2}}\mathbf{1}$ lies in the nullspace of $L_{\text{norm}}$. This means the first normalized eigenvector of $L_{\text{norm}}$ corresponding to eigenvalue zero is proportional to $M^{\frac{1}{2}}\mathbf{1}$, effectively encoding the square root of the sampling density weights (metric-sensitive).
Should we want to learn directly the eigendecomposition of a Laplacian, the normalized formulation 
of $L_{\text{norm}}$
offers a crucial advantage.
Indeed, since the first eigenvector directly encodes the sampling weights 
of  
the discrete metric of the manifold, a trained
neural network predicting the eigenbasis of $L_{\text{norm}}$ simultaneously learns both the spectral decomposition and the manifold metric in a unified manner. 
This eliminates the need for separate modules to predict the mass matrix, simplifying the architecture while maintaining geometric consistency.
Inspired by this remarkable property for Laplacians, which generalizes beyond them (see \cref{sec: Normalized operators and eigenstructures}), we designed our framework to similarly learn a basis, where we decided that the first eigenvector encodes the underlying metric and from which we can explicitly compute the metric-dependent area weights.

Exploiting these insights, we train a neural newtork that predicts a basis for geometry processing directly from unstructured data like point clouds in arbitrary dimension $\mathbb{R}^d$. 
This basis (and associated extracted scalars) yields the eigendecomposition of an implicit operator, which can resemble common operators like the Laplacian, bypassing the need to choose an operator, discretely approximate it via mass and stiffness matrices, and then call eigensolvers.

\subsection{Neural Framework for Direct Spectral Bases}

Our goal is to learn directly a basis for geometry processing on discretely sampled data of arbitrary dimensions, bypassing the need to choose a specific operator, how to discretise it, and calls to sensitive eigensolvers on its disrete approximation.
Should we choose an operator, like the Laplacian, a sub-goal would be to compute its eigendecomposition directly without first discretising it (by combining mass and stiffness matrices) and post hoc eigendecomposition.

Given a point cloud $\mathcal{P}$ discretely sampling a manifold $\mathcal{M}$ with $n$ points $\mathbf{p}_i \in \mathbb{R}^d$ for  $1\le i \le n$, we design a neural network $\Phi_\theta: \mathbb{R}^{n \times d} \rightarrow \mathbb{R}^{n \times K}$ to predict a $K$-dimensional feature vector for each point, where $K$ is the number of basis vectors we wish to predict. 
These per-point predictions form a matrix $\Phi_\theta \left(\mathcal{P}\right) \in \mathbb{R}^{n \times K}$, on which we do QR decomposition, $\Phi_\theta \left(\mathcal{P}\right) = \mathbf{Q}\mathbf{R}$, where $\mathbf{Q} \in \mathbb{R}^{n \times K}$ has orthonormal columns and $\mathbf{R} \in \mathbb{R}^{K \times K}$ is upper triangular.
By analogy with Laplacians, we can interpret $\mathbf{Q} = [\mathbf{q}_1, \mathbf{q}_2, \ldots, \mathbf{q}_K]$,
with $i$-th column $\mathbf{q}_i \in \mathbb{R}^n$, as the first $K$ eigenvectors of a normalized operator, like $L_{\text{norm}}$, corresponding to eigenvalues $\boldsymbol{\lambda} = [\lambda_1, \lambda_2, \ldots, \lambda_K]^\top$, and ordered such that $0 = \lambda_1 \leq \lambda_2 \leq \ldots \leq \lambda_K$. 
For any $k \leq K$, we denote by $\mathbf{Q}_k = [\mathbf{q}_1, \ldots, \mathbf{q}_k] \in \mathbb{R}^{n \times k}$ the matrix containing the first $k\le K$ columns of $\mathbf{Q}$.

This approach radically differs from and short-circuits traditional methods, which require first to choose an operator (e.g.\ $\Delta$), then its approximate discretization (e.g.\ $L_{\text{norm}}$), and then classically eigendecompose it. 
In contrast, our pipeline completely bypasses this altogether by computing a spectral basis directly without first choosing, computing, and eigendecomposing an operator.
Remarkably, we need not learn the metric $M$ separately. Indeed,  with our interpretation, the first eigenvector $\mathbf{q}_1$ directly encodes the mass matrix diagonal, by taking $M = \text{diag}(\mathbf{q}_1 \odot \mathbf{q}_1)$.
We also need not learn separately the eigenvalues by exploiting the min-max theorem (\cref{thm:aflalo_minmax}) linking
the eigenvalues to the maximum approximation error.
During the forward pass, we generate random probe functions for the input point cloud $\mathcal{P}$ and project them onto our predicted truncated bases $\mathbf{Q}_k$ for all $k$.
The maximum reconstruction error across these projections provides an estimate $\alpha_k$, from which we compute $\lambda_{k+1} \approx \tfrac{1}{\alpha_k}$. 
This gives us eigenvalue estimates without requiring additional network parameters.

\subsection{Learning by Optimal Approximations}

Our framework (\cref{fig: framework overview}) operates on a batch of point clouds. Both at train or inference time, we progressively reconstruct probe functions by $M$-orthogonal projections onto the Euclidean-orthogonal estimated bases $\mathbf{Q}_k$ (see how in \cref{sec: Projecting onto Optimal Reconstruction Bases})
for increasing values of $k$ and for each point cloud in the batch, as summarized in \cref{alg:progressive_reconstruction}.

\begin{algorithm}[t]
\caption{Progressive Reconstruction Forward Pass}
\label{alg:progressive_reconstruction}
\begin{algorithmic}[1]
    \Statex \textbf{Input:} Point cloud $\mathcal{P}$ with $n$ points,
    number of probe functions $m$, number of eigenvectors $K$
    \Statex \textbf{Output:} Reconstruction loss $\mathcal{L}_{\text{rec}}$ and maximum-error $e_{\text{max}}\in\mathbb{R}^K$
    \State Generate independently and uniformly $m$ probe functions $\mathbf{f}^{(1)}, \mathbf{f}^{(2)}, \ldots, \mathbf{f}^{(m)} \in \mathbb{R}^n$. By default, do this by iteratively applying Gaussian kernels on the k-nearset neighbor graph to independently and uniformly sampled signals of $\mathbb{R}^n$.
    \State Compute a forward pass on the network $\Phi_\theta (\mathcal{P})$
    \State Compute the QR-decomposition $\Phi_\theta (\mathcal{P}) = \mathbf{Q}\mathbf{R}$
    \For{$k = 1$ to $K$}
        \State $\mathbf{Q}_k \gets \mathbf{Q}$ truncated to the first $k$ columns
        \For{$i = 1$ to $m$}
            \State \parbox[t]{\dimexpr\linewidth-\algorithmicindent}{%
            $\mathbf{f}_{\text{proj},k}^{(i)} \gets$ $M$-projection of 
            $\mathbf{f}^{(i)}$
            onto $\mathbf{Q}_k$%
            }
            \State $e_{k}^{(i)} \gets \|
            \mathbf{f}^{(i)}
            - \mathbf{f}_{\text{proj},k}^{(i)}\|_2^2$
        \EndFor
        \State $i_k^{\text{max}} \gets \mathrm{argmax}_i(e_k^{(i)})$
        \State $e_{\text{max}}.\mathrm{append}(e^{\left(i_k^{\text{max}}\right)}_k)$ 
    \EndFor
    \State $\mathcal{L}_{\text{rec}} \gets \frac{1}{mK} \sum_{i=1}^{m} \sum_{k=1}^{K} e^{\left(i\right)}_k$
\end{algorithmic}
\end{algorithm}

\hfill\break
\noindent\textbf{Training.}
We train our model by learning a basis to optimally reconstruct probe functions.
Depending on the choice of probe function distribution, we will be working implicitly with different operators $L$ in the optimal reconstruction theory (\cref{thm:aflalo_minmax,thm:pca_continuous}), leading to different estimated bases each having their own advantages. 
By default, we take inspiration from the Laplacian, which is the most commonly used operator, yet without computing it. Probe functions for the Laplacian should be smooth, with bounded Dirichlet energy, and uniformly distributed in this bounded set. However, computing the gradient requires the knowledge of the metric, which on unstructured data like high dimensional point clouds is not provided. We relax these constraints by generating probe functions from smoothing random functions with Gaussian kernels on the k-nearest neighbor graph of the data.
The resulting distribution loosely resembles that of the constrained Laplacian, leading to similarly behaved bases, yet we neither aim for exact replica of the Laplacian nor or are we constrained to this operator or this choice of distribution.

Our training loss is based on optimal-reconstruction theory.
Instead of the min-max formulation (\cref{thm:aflalo_minmax}), where for each $k$ only one probe function in the batch (the worst approximated one) would be used to optimize the model, we switch to the equivalent PCA one (\cref{thm:pca_continuous}) averaging out the contribution of all probe functions.
This leads to stabler optimization. 
Our proposed loss is thus the average reconstruction loss $\mathcal{L}_{\text{rec}}$ measuring the average quality of these progressive approximations
\begin{equation}  
    \label{eq: reconstruction loss}
    \mathcal{L}_{\text{rec}} = \frac{1}{mK} \sum_{i=1}^{m} \sum_{k=1}^{K} \|
    \mathbf{f}^{(i)}
    - \mathbf{f}_{\text{proj},k}^{(i)}\|_2^2,
\end{equation}
where 
$\mathbf{f}^{(i)}, \mathbf{f}_{\text{proj},k}^{(i)} \in \mathbb{R}^n$
are the $i$-th probe function and its projection onto the first $k$ estimated basis vectors. 
We train our model by backpropagating through the entire pipeline using $\mathcal{L}_{\text{rec}}$ as the sole training objective, enabling the network to learn the optimal basis
through end-to-end gradient descent-based optimization.
Note that we switched from the $M$-norm to the unweighted Euclidean norm. This hybrid approach, combining $M$-weighted projection with $2$-norm error, creates an unsupervised mechanism for learning 
a density-related mass matrix and improves the training stability (see \cref{sec: Replacing the $M$-norm with the Euclidean norm for Approximation Errors}).

Importantly, we never compute the implicit optimal reconstruction operator nor its eigenvalues during training. Nevertheless, they are easy to compute at inference time.

\hfill\break
\noindent \textbf{Inference.}
A feed-forward pass computes our optimal reconstruction basis for each point cloud in the inference batch.
We can then easily compute the associated implicit optimal reconstruction operator or its eigenvalues for downstream geometric tasks.
Thanks to the min-max formulation of optimal reconstruction theory (\cref{thm:aflalo_minmax}) the eigenvalues can be estimated from the worst-case reconstruction over all the probe functions at each spectral resolution $k\in\{1,\ldots,K\}$ using 
$\lambda_{k+1} = \tfrac{1}{\max_i \|\mathbf{f}^{(i)} - \mathbf{f}_{\text{proj},k}^{(i)}\|_2^2}$.
This provides theoretically-grounded eigenvalue estimates 
directly from the estimated basis rather than in parallel to it.
Given the basis and its associated eigenvalues, we can then optionally explicitly reconstruct the implicit normalized symmetric operator by matrix multiplication $Q_K \Lambda_K  Q_K^T$, or it unnormalized non-symmetric version $M^{-\frac{1}{2}} Q_K \Lambda_K Q_K^T  M^{\frac{1}{2}}$. However, the operator matrix has little use in practice, as even its action is usually computed in its spectral basis. Thus, in our experiments, we never needed to recompute it explicitly.
From the estimated normalized basis $\mathbf{q}_i$, we can also compute an unnormalized spectral basis $\mathbf{v}_i = M^{-\frac{1}{2}}\mathbf{q_i}$, which can be preferable downstream as it is more sampling invariant. For instance, the first vector $\mathbf{v}_1$ is constant regardless of
the sampling of the the underlying smooth manifold.


\section{Experiments}
\label{sec:experiments}

We demonstrate the versatility of our framework to compute spectral bases on arbitrary data. Starting from a sanity check in a toy 1D setting, with points sampled in $[0,1]\subset\mathbb{R}$, we show that we can handle not only traditional 3D point clouds sampled on two-dimensional surfaces in $\mathbb{R}^3$, but also high-dimensional data in $\mathbb{R}^d$ such as image embeddings.
Full details on the data, implementation, and further results are pushed to \cref{sec: additional experimental details sup mat}.

\subsection{Toy 1D Segment Manifold}
\label{sec:1d-sanity}

As a sanity check, we illustrate our method on the simplest geometry: the unit interval $\Omega=[0,1]$.

\begin{figure}[ht]
    \centering
    \includegraphics[width=1\columnwidth]{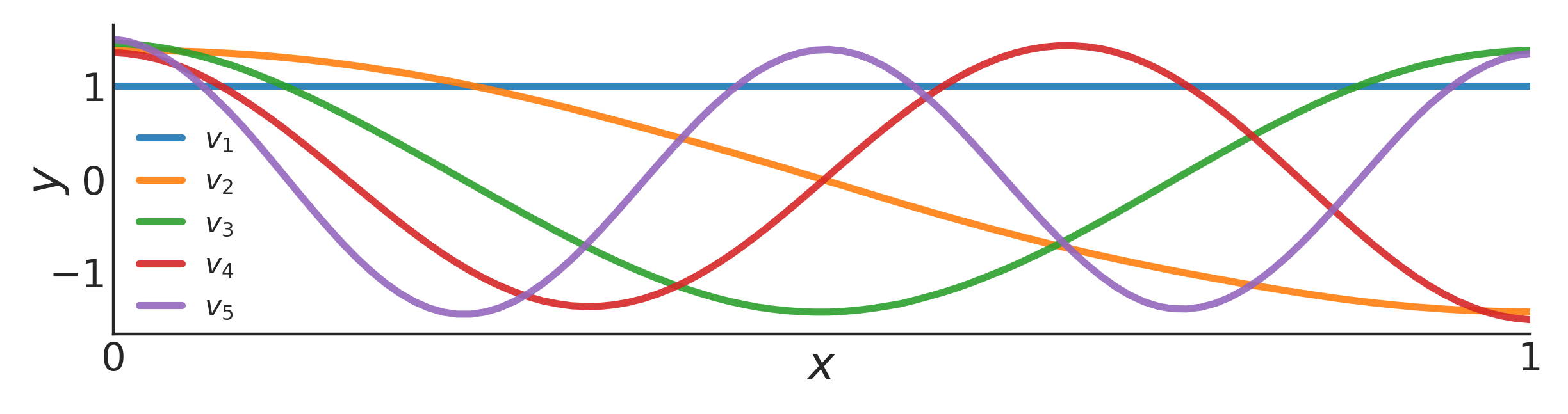}
    \caption{Learned eigenfunctions on [0,1] recover frequency-ordered harmonics resembling the Laplacian's spectrum.}
    \label{fig: 1d_eigenvectors}
\end{figure}

%

\noindent \textbf{Setting.}
We grid sample 100 points on $[0,1]$, forming a point cloud. As in higher dimensions, random probe functions are first smoothened, without boundary constraints.
Our learned extractor $\Phi_\theta$ is a small MLP, which suffices in this single small point cloud setting.

\hfill \break
\noindent \textbf{Results.}
We plot in \cref{fig: 1d_eigenvectors} the first five learned unnormalized spectral basis vectors $\mathbf{v}_i = M^{-\frac{1}{2}}\mathbf{q}_i$ after sign alignment.
The network recovers a Fourier basis-like family with frequency-increasing harmonics: $\mathbf{v}_1$ is constant, while the other $\mathbf{v}_i$ exhibit $i$ half-waves across the interval.
This minimal example showcases how we can compute a frequency-ordered orthonormal basis resembling the Fourier basis, i.e.\ the Laplacian's eigenfunctions, just by optimising the approximation objective on basic probe functions. Importantly, this happens because the metric weights $M$, extracted from the first normalized basis vector $\mathbf{q}_1$, are sensible as they are correlated with intuition.

\subsection{2D Surface and 3D Volume Manifolds}

We here show that our method can learn an approximation of the eigendecomposition of the most prevalent operator in shape analysis: the Laplace-Beltrami operator. 
By learning on single 3D point clouds (overfitting setting), we obtain highly accurate estimates.
By learning on a wide collection of 3D point clouds from various datasets (generalization setting), we may obtain a foundation model that generalizes to unseen point clouds in $\mathbb{R}^3$ without retraining.


\begin{figure}[t]
    \centering
    \includegraphics[width=0.98\columnwidth]{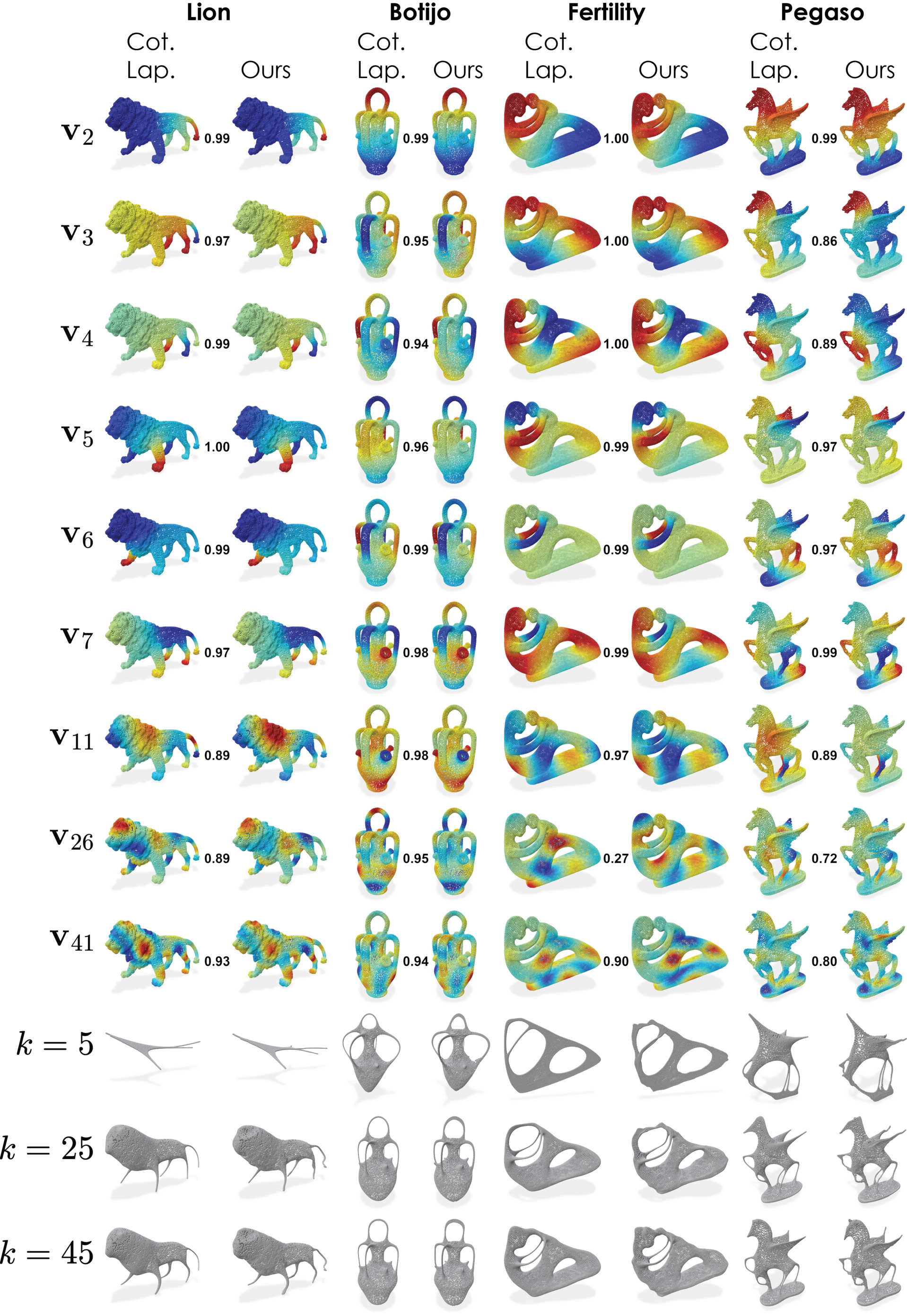}
    \caption{
    Unnormalized spectral basis (top) and $xyz$ reconstruction from $k$ basis vectors (bottom), using either the oracle cotangent Laplacian or our method (overfitting setting). 
    Scalars are cosine similarities between basis vectors. We get similar if not more detailed reconstructions.
    More in \cref{sec: additional experimental details sup mat}.
    }
    \label{fig: 3d point clouds overfitting spectral bases and reconstructions}
    \vspace{-0.5em}
\end{figure}

\begin{figure}[t]
    \centering
    \includegraphics[width=\columnwidth]{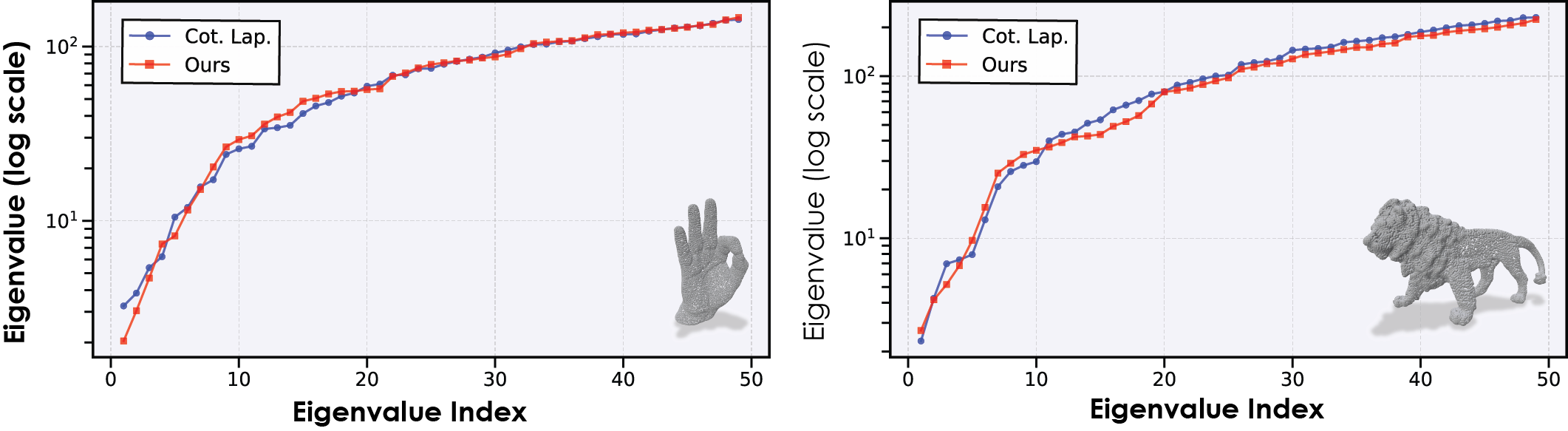}
    \caption{
    Eigenvalues of the oracle cotangent Laplacian and our estimated ones (overfitting setting). More in 
    \cref{sec: additional experimental details sup mat}.
    }
    \label{fig: 3d point clouds overfitting eigenvalue curves}
    \vspace{-0.5em}
\end{figure}

\begin{figure}[t]
    \centering
    \includegraphics[width=\columnwidth]{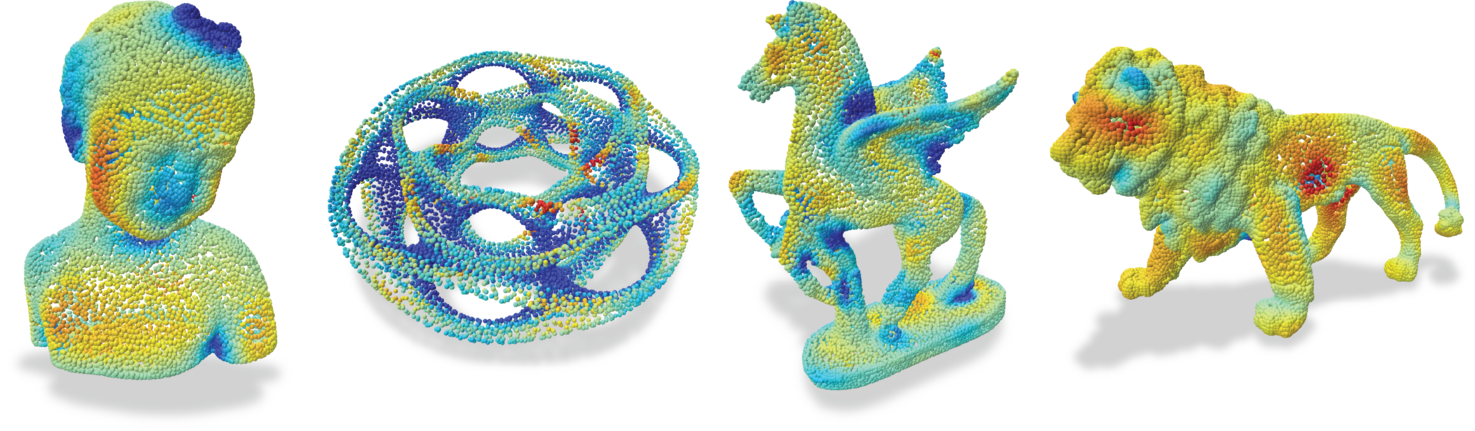}
    \caption{
    Estimated mass metric $M$ from $\mathbf{q}_1$ (overfitting setting).
    }
    \label{fig: 3d point clouds overfitting metric}
    \vspace{-1em}
\end{figure}


\hfill \break
\noindent \textbf{Datasets.}
The shapes in the overfitting setting are from \cite{myles2014robust}.
In the generalization setting, we train our model on a wide collection of surface datasets to ensure broad generalization,  ranging from protein structures to human scans \cite{poulenard2019effective,biasotti2021shrec,koch2019abc,bogo2014faust,li20214dcomplete}.
We evaluate on reference shape analysis benchmarks \cite{sumner2004deformation,myles2014robust,giorgi2007watertight,pickup2014shrec14,lian15nonrigid,lahner2016shrec,melzi2019shrec,dyke2020shrec,thompson2020shrec,zhou2016thingi10k}, spanning many challenges, e.g.\ deformations, topology variations, and different geometric characteristics.
In each dataset, we dropped the mesh connectivity to work only with point clouds. Also, all shapes are scaled to fit within a unit sphere.
We evaluate generalization to new manifold dimensionality by testing on 3D volumetric point clouds, computed by randomly sampling points inside the volume of shapes in \cite{dyke2020shrec}, the model learned on surface point clouds.

\hfill \break
\noindent \textbf{Methods.}
We compare our neural framework (without connectivity) using a transformer \cite{vaswani2017attention} as learned extractor $\Phi_\theta$ against the reference axiomatic oracle: the classical cotangent Laplacian (with oracle mesh connectivity).

\begin{table}[]
    \centering
    \small
    \caption{
    Average cosine similarity between predicted and oracle
    eigenfunctions at different truncation levels $k$, and mean relative eigenvalue discrepancy (overfitting setting).
    More in \cref{sec: additional experimental details sup mat}.
    }
    \resizebox{\columnwidth}{!}{%
    \begin{tabular}{l c c c c c}
        \toprule
            \textbf{Shape} & 
            \textbf{Image}
            & $\boldsymbol{k \leq 10}$ & $\boldsymbol{k \leq 20}$ & $\boldsymbol{k \leq 50}$ & 
            $\boldsymbol{\lambda}$ 
            \textbf{Discrepancy}
            \\
        \midrule

            Armadillo & \adjustbox{valign=c}{\includegraphics[width=0.02\textwidth]{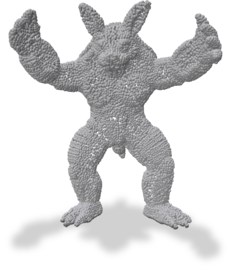}} & 0.968 & 0.967 & 0.773 & 0.200 $\pm$ 0.126 \\
            \midrule
            Bimba & \adjustbox{valign=c}{\includegraphics[width=0.02\textwidth]{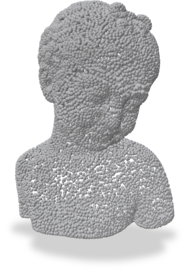}} & 0.964 & 0.945 & 0.822 & 0.093 $\pm$ 0.145 \\
            \midrule
            Botijo & \adjustbox{valign=c}{\includegraphics[width=0.015\textwidth]{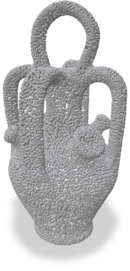}} & 0.972 & 0.955 & 0.813 & 0.153 $\pm$ 0.092 \\
            \midrule
            Elephant & \adjustbox{valign=c}{\includegraphics[width=0.02\textwidth]{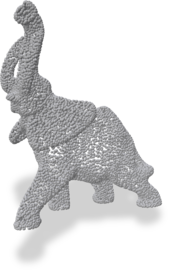}} & 0.979 & 0.866 & 0.687 & 0.105 $\pm$ 0.123 \\
            \midrule
            Fertility & \adjustbox{valign=c}{\includegraphics[width=0.02\textwidth]{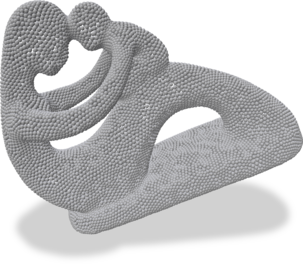}} & 0.874 & 0.866 & 0.720 & 0.083 $\pm$ 0.106 \\
            \midrule
            Kitten & \adjustbox{valign=c}{\includegraphics[width=0.015\textwidth]{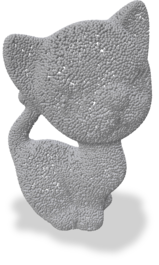}} & 0.993 & 0.988 & 0.981 & 0.088 $\pm$ 0.104 \\
            \midrule
            Laurent Hand & \adjustbox{valign=c}{\includegraphics[width=0.02\textwidth]{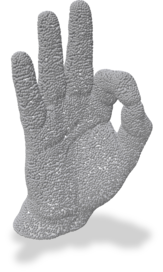}} & 0.823 & 0.696 & 0.568 & 0.066 $\pm$ 0.078 \\
            \midrule
            Lion & \adjustbox{valign=c}{\includegraphics[width=0.02\textwidth]{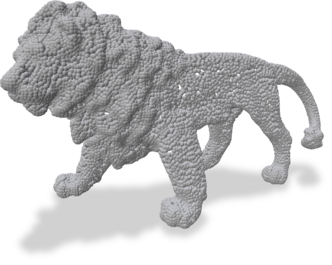}} & 0.951 & 0.908 & 0.822 & 0.067 $\pm$ 0.086 \\
            \midrule
            Pegaso & \adjustbox{valign=c}{\includegraphics[width=0.02\textwidth]{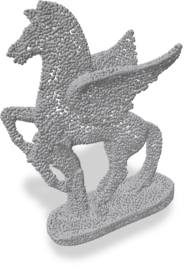}} & 0.932 & 0.797 & 0.544 & 0.142 $\pm$ 0.140 \\

        \bottomrule
    \end{tabular}
    }%
    \label{tab: 3d point clouds overfitting quantitative cosine sim and lambda normalised discrepancy}
    \vspace{-1em}
\end{table}

\hfill \break
\noindent \textbf{Results -- Overfitting setting.}
We plot the learned unnormalized eigenvectors $\mathbf{v}_i$ in \cref{fig: 3d point clouds overfitting spectral bases and reconstructions}. 
Remarkably, our eigenvectors, unsupervisedly learned solely using optimal-approximation theory and without knowledge of the underlying mesh, are almost identical to those computed for the cotangent Laplacian, which relies on the oracle mesh structure, with cosine similarity between them often close to $1$ (see \cref{tab: 3d point clouds overfitting quantitative cosine sim and lambda normalised discrepancy}).
Additionally, the eigenvalues extracted from the worst case errors provide a good approximation to those of the oracle (see \cref{fig: 3d point clouds overfitting eigenvalue curves}). 
These results show that our neural unsupervised method can be used in an overfitting setting to get highly accurate spectral basis estimates that match those of the targeted Laplace-Beltrami oracle.
On some shapes though, e.g.\ Pegaso, higher frequency basis vectors slightly diverge from the oracle. Yet by analyzing shape reconstruction (spectral filtering) results (see \cref{fig: 3d point clouds overfitting spectral bases and reconstructions}), projecting the $xyz$ coordinates to the first $k$ spectral vectors, we see that our model captures additional details lost in the oracle, providing better top $k$ approximation. Thus not only can we imitate the reference oracle method, we can in some cases provide a superior version with better compressed information in the spectral basis.
A cornerstone of our method is the unsupervised extraction of the metric weights $M$ directly from the first estimated normalized vector $\mathbf{q}_1$, without knowledge of the mesh structure. We see in \cref{fig: 3d point clouds overfitting metric} that these estimated area weights are indeed well-behaved.

\begin{figure}[t]
    \centering
    \includegraphics[width=\columnwidth]{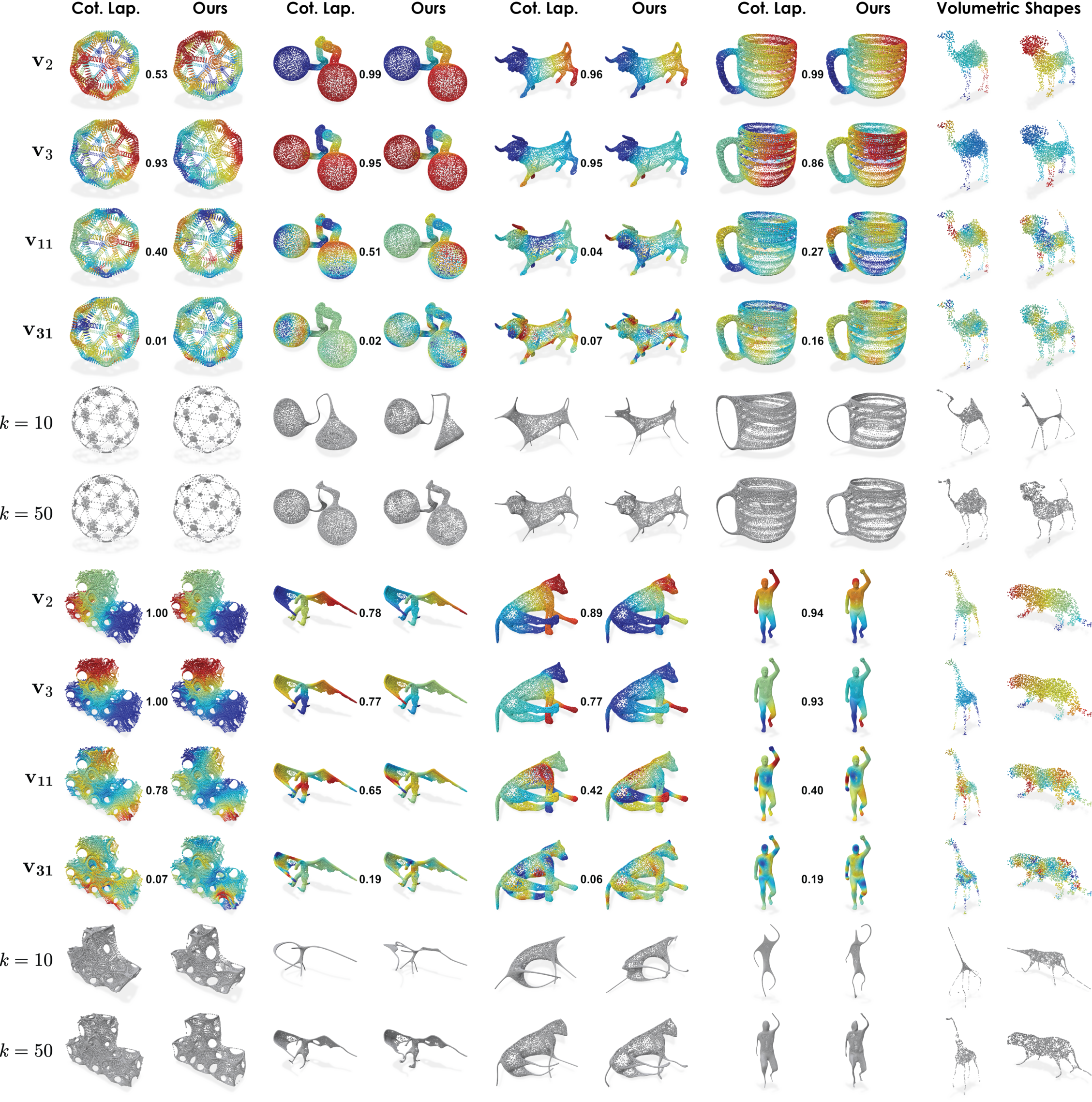}
    \caption{
    Unnormalized spectral basis $\mathbf{v}_1$ on unseen shapes, either surfaces (left) or volumes (right), when the model was trained on a wide collection of surface point clouds (generalization setting). 
    Our model exhibits foundation-level generalization capabilities.
    }
    \label{fig: 3d point clouds generalisation spectral bases}
    \vspace{-1em}
\end{figure}

\hfill \break
\noindent \textbf{Results -- Generalization setting.}
Here, our model is trained on a wide collection of surface point clouds.
We plot learned unnormalized eigenvectors $\mathbf{v}_i$ and filter the $xyz$ coordinates on unseen evaluation shapes, either surfaces or volumes (\cref{fig: 3d point clouds generalisation spectral bases}), which is a type of manifold never seen in training. Our model generalizes well to new eclectic types of shapes, yet with smaller precision than in the overfitting setting, demonstrating that our framework could provide
unsupervised foundation models to compute spectral decompositions generalizing well beyond the training data.

\subsection{High-Dimensional Manifolds}

To demonstrate the generality of our approach, we experiment, beyond the classical 2D surface setting, on manifolds with high intrinsic dimensionality.
Here, each image is a single point on the dataset manifold, with distances between images measured in a pretrained feature space.

\hfill \break
\noindent \textbf{Datasets.}
We use standard image classification datasets: 
STL10 \cite{coates2011analysis},
Imagenette \cite{howard2019imagenette},  
CIFAR100 \cite{krizhevsky2009learning},
and Caltech256 \cite{griffin2007caltech}
having
10, 
10, 
100, 
and 256 classes.
Here, we do not mix datasets, but follow the train-test splits.
As is standard, rather than working on raw pixel images, we use reference feature embeddings,
DINOv2~\cite{oquab2023dinov2} and CLIP~\cite{radford2021learning}, as high-dimensional (yet lower dimensional than the raw pixel image) image coordinates in $\mathbb{R}^d$, with $d\!=\!768$ (resp.\ $512$) for DINOv2 (resp.\ CLIP).
Dataset manifolds are thus mapped to submanifolds of $\mathbb{R}^d$, and image distances are measured on embeddings.

\begin{figure}[t]
    \centering
    \includegraphics[width=0.95\columnwidth]{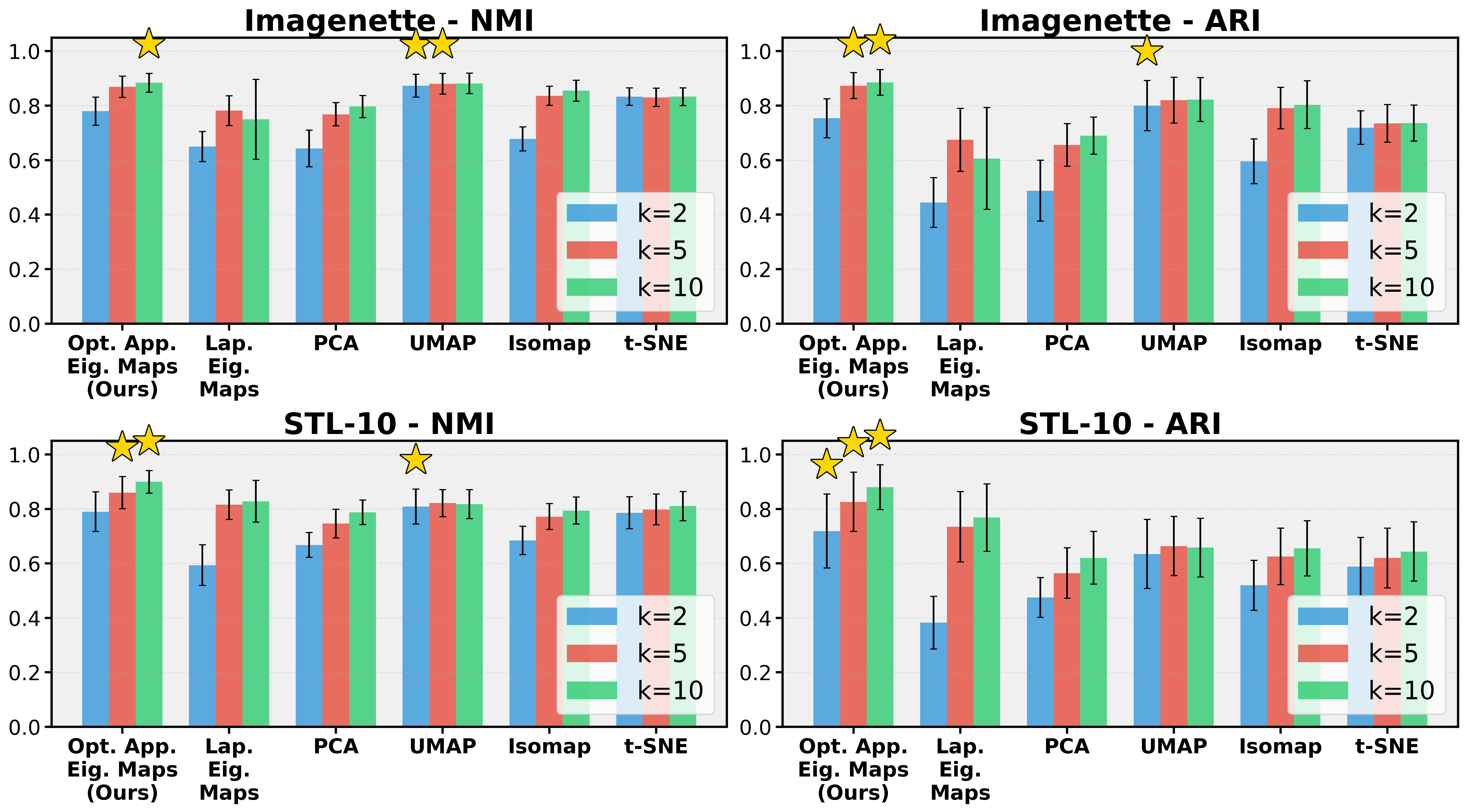}
    \caption{
    Average clustering performance over 50 runs of manifold learning methods on DINOv2 features of random data subsets (1500 images). Higher is better. More in \cref{sec: additional experimental details sup mat}.
    }
    \label{fig: manifold learning nmi ari stl10 imagenette}
    \vspace{-0.5em}
\end{figure}

\begin{figure}[t]
    \centering
    \includegraphics[width=0.8\columnwidth]{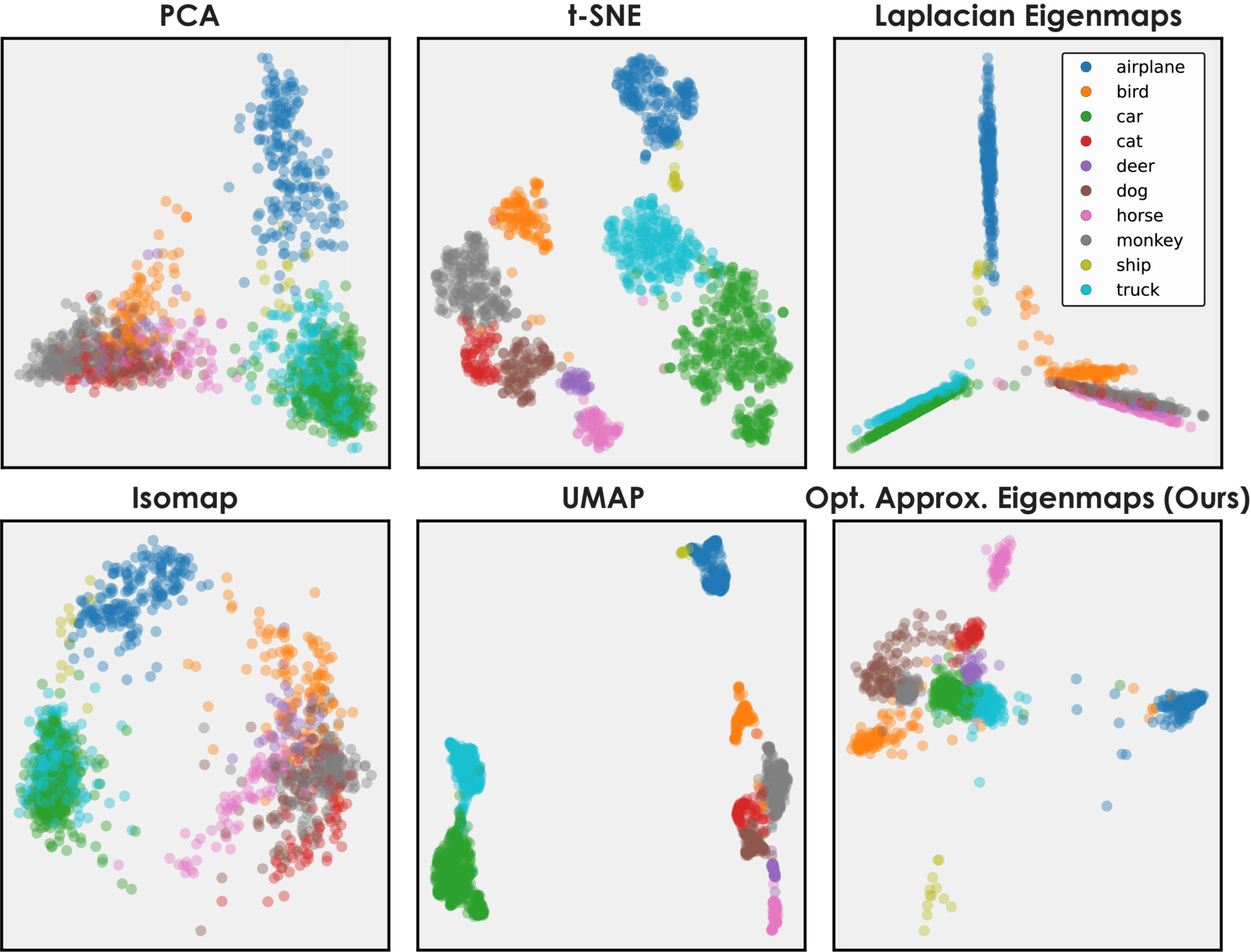}
    \caption{
    Manifold learning visualization of a random subset of STL10 by 2D embedding DINOv2 features. More in \cref{sec: additional experimental details sup mat}.
    }
    \label{fig: manifold learning visualisations}
    \vspace{-1em}
\end{figure}

\hfill \break
\noindent \textbf{Methods.}
Unlike 3D point clouds, high-dimensional data cannot be visualized and analyzed directly. Manifold learning addresses this by finding low-dimensional embeddings in $\mathbb{R}^{k}$, with $k\ll d$, that preserve pairwise dissimilarities to capture manifold structures.
In particular, it enables 2D visualizations ($k=2$) of high-dimensional data.
For a short overview of manifold learning see \cref{fig: manifold learning nmi ari stl10 imagenette}.
For classification data, it is widely accepted that better manifold learning methods provide better clustered embeddings correlating with the class labels. 
This can be both evaluated visually or with metrics such as the Normalized Mutual Information (NMI) \cite{vinh2009information} and Adjusted Rand Index (ARI) \cite{hubert1985comparing}.
Our unnormalized spectral basis $\mathbf{v}_i$ naturally provides $k$-low-dimensional embeddings, which is our proposed manifold learning technique that we name Optimal-Approximation Eigenmaps. It is a direct generalization of Laplacian Eigenmaps \cite{belkin2003laplacian}, which uses the graph Laplacian spectral basis instead.
We compare for various embedding dimensions $k\in\{2,5,10,50\}$ our method against reference manifold learning techniques: PCA \cite{pearson1901pca,hotelling1933analysis}, Isomap \cite{tenenbaum2000global}, Laplacian Eigenmaps \cite{belkin2003laplacian}, t-SNE \cite{maaten2008visualizing,van2014accelerating}, and UMAP \cite{mcinnes2018umap}.

\hfill \break
\noindent \textbf{Results.}
Based on our results (see \cref{fig: manifold learning visualisations,fig: manifold learning nmi ari stl10 imagenette}), our learned spectral basis consistently provides competitive or superior embeddings compared to baselines.
These results showcase the strength of our neural spectral decomposition for capturing the intrinsic geometry of manifolds.
In particular, our superiority compared to the similar Laplacian Eigenmaps suggests that our spectral basis and associated implicit operator are superior to those of the graph Laplacian, which is the standard for high-dimensional data.

\section{Conclusion}

We propose a neural framework to compute directly spectral bases on unstructured data of any dimensionality, bypassing the traditional need to first choose, discretize, and then eigendecompose an operator. Grounded in optimal-approximation theory, we find the orthonormal basis that optimally approximates constrained probe functions, with different probe distributions encoding different implicit operators. From the estimated basis, we can recover directly the manifold metric and compute explicitly the associated eigenvalues of the associated implicit operator. In a wide range of experiments covering one to three-dimensional manifolds, we showed that our method can recover the Laplace-Beltrami operator (LBO), which is the most useful operator in 3D geometry. However, our method scales in practice to high-dimensionality, unlike the LBO, and we show that it provides advantageous spectral representations via manifold learning experiments on image datasets. Our data-driven alternative to conventional pipelines paves the way for new avenues in geometry processing for unstructured data, especially in high-dimensions.

\hfill \break
\noindent \textbf{Limitations and future work.}
Training our method is computationally expensive, 
requiring time and GPUs, 
although this can be improved by optimizing our code with efficient compilation and CUDA implementations.
Due to constraints, we focused on few tasks, yet we intend to explore further downstream applications, especially in high-dimensions, such as using our spectral basis in graph neural networks,  
or learning a more general foundation model that would apply to data of any dimensionality.
At the heart of our method, the choice of probe functions imposes an underlying metric and implicit operator. By default we smoothen random functions on the kNN graph, giving similar results to the Laplacian. Yet, to best approximate it we need to manually tune distribution hyperparameters per shape, as we do in the overfitting setting.
In future work, we will learn probe distributions 
and explore the approximation of more operators beyond the traditional Laplacians.



\clearpage

{
    \small
    \bibliographystyle{ieeenat_fullname}
    \bibliography{main}

@String(CVPR= {IEEE Conf. Comput. Vis. Pattern Recog.})

@String(ICCV= {Int. Conf. Comput. Vis.})

@String(TOG= {ACM Trans. Graph.})

@String(ICLR = {Int. Conf. Learn. Represent.})

@String(AAAI = {AAAI})

@String(CVPR  = {CVPR})

@String(ICCV  = {ICCV})

@String(TOG   = {ACM TOG})

@String(ICLR  = {ICLR})

@article{quackenbush2024geometric,
  title={Geometric neural operators (gnps) for data-driven deep learning in non-euclidean settings},
  author={Quackenbush, B and Atzberger, PJ},
  journal={Machine Learning: Science and Technology},
  volume={5},
  number={4},
  pages={045033},
  year={2024},
  publisher={IOP Publishing}
}

@article{yigit2025lbonet,
  title={LBONet: Supervised Spectral Descriptors for Shape Analysis},
  author={Yigit, Oguzhan and Wilson, Richard C},
  journal={IEEE Transactions on Pattern Analysis and Machine Intelligence},
  year={2025},
  publisher={IEEE}
}

@article{choukroun2018sparse,
    author = {Choukroun, Yoni and Pai, Gautam and Kimmel, Ron},
    title = {Sparse Approximation of 3D Meshes Using the Spectral Geometry of the Hamiltonian Operator},
    year = {2018},
    issue_date = {July 2018},
    publisher = {Kluwer Academic Publishers},
    address = {USA},
    volume = {60},
    number = {6},
    issn = {0924-9907},
    url = {https://doi.org/10.1007/s10851-018-0822-0},
    doi = {10.1007/s10851-018-0822-0}, journal = {J. Math. Imaging Vis.},
    month = jul,
    pages = {941–952},
    numpages = {12}
}

@article{williamson2025neural,
    author = {Williamson, Romy and Mitra, Niloy J.},
    title = {Neural Geometry Processing via Spherical Neural Surfaces},
    journal = {Computer Graphics Forum},
    volume = {44},
    number = {2},
    pages = {e70021},
    doi = {https://doi.org/10.1111/cgf.70021},
    url = {https://onlinelibrary.wiley.com/doi/abs/10.1111/cgf.70021},
    eprint = {https://onlinelibrary.wiley.com/doi/pdf/10.1111/cgf.70021},
    year = {2025}
}

@inproceedings{bracha2020shape,
booktitle = {Eurographics Workshop on 3D Object Retrieval},
editor = {Schreck, Tobias and Theoharis, Theoharis and Pratikakis, Ioannis and Spagnuolo, Michela and Veltkamp, Remco C.},
title = {{Shape Correspondence by Aligning Scale-invariant LBO Eigenfunctions}},
author = {Bracha, Amit and Halimi, Oshri and Kimmel, Ron},
year = {2020},
publisher = {The Eurographics Association},
ISSN = {1997-0471},
ISBN = {978-3-03868-126-7},
DOI = {10.2312/3dor.20201159}
}

@inproceedings{sun2009concise,
  title={A concise and provably informative multi-scale signature based on heat diffusion},
  author={Sun, Jian and Ovsjanikov, Maks and Guibas, Leonidas},
  booktitle={Computer graphics forum},
  volume={28},
  number={5},
  pages={1383--1392},
  year={2009},
  organization={Wiley Online Library}
}

@article{reuter2006laplace,
title = {Laplace–Beltrami spectra as ‘Shape-DNA’ of surfaces and solids},
journal = {Computer-Aided Design},
volume = {38},
number = {4},
pages = {342-366},
year = {2006},
note = {Symposium on Solid and Physical Modeling 2005},
issn = {0010-4485},
doi = {https://doi.org/10.1016/j.cad.2005.10.011},
url = {https://www.sciencedirect.com/science/article/pii/S0010448505001867},
author = {Martin Reuter and Franz-Erich Wolter and Niklas Peinecke}
}

@article{kac1966can,
  title={Can one hear the shape of a drum?},
  author={Kac, Mark},
  journal={The american mathematical monthly},
  volume={73},
  number={4P2},
  pages={1--23},
  year={1966},
  publisher={Taylor \& Francis}
}

@article{protter1987can,
  title={Can one hear the shape of a drum? Revisited},
  author={Protter, Murray H},
  journal={Siam Review},
  volume={29},
  number={2},
  pages={185--197},
  year={1987},
  publisher={SIAM}
}

@article{gordon1992one,
  title={One cannot hear the shape of a drum},
  author={Gordon, Carolyn and Webb, David L and Wolpert, Scott},
  journal={Bulletin of the American Mathematical Society},
  volume={27},
  number={1},
  pages={134--138},
  year={1992}
}

@article{gordon1996you,
  title={You can't hear the shape of a drum},
  author={Gordon, Carolyn and Webb, David},
  journal={American Scientist},
  volume={84},
  number={1},
  pages={46--55},
  year={1996},
  publisher={JSTOR}
}

@inproceedings{bensaid2023hamiltonian,
    author = {Bensa\"{\i}d, David and Bracha, Amit and Kimmel, Ron},
    title = {Partial Shape Similarity by Multi-metric Hamiltonian Spectra Matching},
    year = {2023},
    isbn = {978-3-031-31974-7},
    publisher = {Springer-Verlag},
    address = {Berlin, Heidelberg},
    url = {https://doi.org/10.1007/978-3-031-31975-4_55},
    doi = {10.1007/978-3-031-31975-4_55}, booktitle = {Scale Space and Variational Methods in Computer Vision: 9th International Conference, SSVM 2023, Santa Margherita Di Pula, Italy, May 21–25, 2023, Proceedings},
    pages = {717–729},
    numpages = {13}, location = {Santa Margherita di Pula, Italy} 
}

@article{coifman2006diffusion,
title = {Diffusion maps},
journal = {Applied and Computational Harmonic Analysis},
volume = {21},
number = {1},
pages = {5-30},
year = {2006},
note = {Special Issue: Diffusion Maps and Wavelets},
issn = {1063-5203},
doi = {https://doi.org/10.1016/j.acha.2006.04.006},
url = {https://www.sciencedirect.com/science/article/pii/S1063520306000546},
author = {Ronald R. Coifman and Stéphane Lafon}
}

@article{hammond2011wavelets,
    title = {Wavelets on graphs via spectral graph theory},
    journal = {Applied and Computational Harmonic Analysis},
    volume = {30},
    number = {2},
    pages = {129-150},
    year = {2011},
    issn = {1063-5203},
    doi = {https://doi.org/10.1016/j.acha.2010.04.005},
    url = {https://www.sciencedirect.com/science/article/pii/S1063520310000552},
    author = {David K. Hammond and Pierre Vandergheynst and Rémi Gribonval}
}

@article{aflalo2016best, 
  year     = {2016}, 
  title    = {Best bases for signal spaces}, 
  author   = {Aflalo, Yonathan and Brezis, Haïm and Bruckstein, Alfred and Kimmel, Ron and Sochen, Nir}, 
  journal  = {Comptes Rendus. Mathématique}, 
  issn     = {1631-073X}, 
  doi      = {10.1016/j.crma.2016.10.002}, 
  pages    = {1155--1167}, 
  number   = {12}, 
  volume   = {354}
}

@article{brezis2017rigidity, 
  year     = {2017}, 
  title    = {Rigidity of optimal bases for signal spaces}, 
  author   = {Brezis, Haïm and Gómez-Castro, David}, 
  journal  = {Comptes Rendus. Mathématique}, 
  issn     = {1631-073X}, 
  doi      = {10.1016/j.crma.2017.06.004}, 
  pages    = {780--785}, 
  number   = {7}, 
  volume   = {355}
}

@article{sharp2020laplacian, 
  year     = {2020}, 
  title    = {A Laplacian for Nonmanifold Triangle Meshes}, 
  author   = {Sharp, Nicholas and Crane, Keenan}, 
  journal  = {Computer Graphics Forum}, 
  issn     = {0167-7055}, 
  doi      = {10.1111/cgf.14069}, 
  pages    = {69--80}, 
  number   = {5}, 
  volume   = {39}
}

@article{pang2024neural, 
  year     = {2024}, 
  title    = {Neural Laplacian Operator for 3D Point Clouds}, 
  author   = {Pang, Bo and Zheng, Zhongtian and Li, Yilong and Wang, Guoping and Wang, Peng-Shuai}, 
  journal  = {{ACM} Transactions on Graphics}, 
  issn     = {0730-0301}, 
  doi      = {10.1145/3687901}, 
  pages    = {1--14}, 
  number   = {6}, 
  volume   = {43}
}

@article{vaswani2017attention, 
  year     = {2017}, 
  title    = {Attention Is All You Need}, 
  author   = {Vaswani, Ashish and Shazeer, Noam and Parmar, Niki and Uszkoreit, Jakob and Jones, Llion and Gomez, Aidan N and Kaiser, Lukasz and Polosukhin, Illia}, 
  journal  = {{arXiv}}, 
  doi      = {10.48550/{arXiv}.1706.03762}, 
  eprint   = {1706.03762}
}

@article{sorkine2004laplacian, 
  year     = {2004}, 
  title    = {Laplacian surface editing}, 
  author   = {Sorkine, O and Cohen-Or, D and Lipman, Y and Alexa, M and Rössl, C and Seidel, H -P}, 
  journal  = {Proceedings of the 2004 Eurographics/{ACM} {SIGGRAPH} symposium on Geometry processing}, 
  doi      = {10.1145/1057432.1057456}, 
  pages    = {175--184}
}

@article{levie2021transferability,
  title={Transferability of spectral graph convolutional neural networks},
  author={Levie, Ron and Huang, Wei and Bucci, Lorenzo and Bronstein, Michael and Kutyniok, Gitta},
  journal={Journal of Machine Learning Research},
  volume={22},
  number={272},
  pages={1--59},
  year={2021}
}

@article{botsch2010polygon, 
  year     = {2010}, 
  title    = {Polygon Mesh Processing}, 
  author   = {Botsch, Mario and Kobbelt, Leif and Pauly, Mark and Alliez, Pierre and Levy, Bruno}, 
  doi      = {10.1201/b10688-8}, 
  pages    = {97--122}
}

@article{giorgi2007watertight,
  title={Watertight models track},
  author={Giorgi, Dani{\'e}la and Biasotti, Silvia and Paraboschi, Laura},
  journal={Shape Retrieval Contest},
  year={2007}
}

@inproceedings{pickup2014shrec14,
author = {Pickup, D. and Sun, X. and Rosin, P. L. and Martin, R. R. and Cheng, Z. and Lian, Z. and Aono, M. and Ben Hamza, A. and Bronstein, A. and Bronstein, M. and Bu, S. and Castellani, U. and Cheng, S. and Garro, V. and Giachetti, A. and Godil, A. and Han, J. and Johan, H. and Lai, L. and Li, B. and Li, C. and Li, H. and Litman, R. and Liu, X. and Liu, Z. and Lu, Y. and Tatsuma, A. and Ye, J.},
title = {S{H}{R}{E}{C}'14 track: Shape Retrieval of Non-Rigid 3D Human Models},
booktitle = {Proceedings of the 7th Eurographics workshop on 3D Object Retrieval},
series = {EG 3DOR'14},
year = {2014},
numpages = {10},
publisher = {Eurographics Association}
}

@inproceedings {lian15nonrigid,
booktitle = {Eurographics Workshop on 3D Object Retrieval},
editor = {I. Pratikakis and M. Spagnuolo and T. Theoharis and L. Van Gool and R. Veltkamp},
title = {{Non-rigid 3D Shape Retrieval}},
author = {Lian, Z. and Zhang, J. and Choi, S. and ElNaghy, H. and El-Sana, J. and Furuya, T. and Giachetti, A. and Guler, R. A. and Lai, L. and Li, C. and Li, H. and Limberger, F. A. and Martin, R. and Nakanishi, R. U. and Neto, A. P. and Nonato, L. G. and Ohbuchi, R. and Pevzner, K. and Pickup, D. and Rosin, P. and Sharf, A. and Sun, L. and Sun, X. and Tari, S. and Unal, G. and Wilson, R. C.},
year = {2015},
publisher = {The Eurographics Association},
DOI = {10.2312/3dor.20151064}
}

@inproceedings{lahner2016shrec,
  title={{SHREC}'16: Matching of deformable shapes with topological noise},
  author={L{\"a}hner, Zorah and Rodol{\`a}, Emanuele and Bronstein, Michael M and Cremers, Daniel and Burghard, Oliver and Cosmo, Luca and Dieckmann, Alexander and Klein, Reinhard and Sahillioǧlu, Y and others},
  booktitle={Eurographics Workshop on 3D Object Retrieval, EG 3DOR},
  pages={55--60},
  year={2016},
  organization={Eurographics Association}
}

@inproceedings{melzi2019shrec,
  title={{SHREC}’19: matching humans with different connectivity},
  author={Melzi, Simone and Marin, Riccardo and Rodol{\`a}, Emanuele and Castellani, Umberto and Ren, Jing and Poulenard, Adrien and Ovsjanikov, P and others},
  booktitle={Eurographics Workshop on 3D Object Retrieval},
  pages={1--8},
  year={2019},
  organization={The Eurographics Association}
}

@article{dyke2020shrec,
  title={{SHREC}’20: Shape correspondence with non-isometric deformations},
  author={Dyke, Roberto M and Lai, Yu-Kun and Rosin, Paul L and Zappal{\`a}, Stefano and Dykes, Seana and Guo, Daoliang and Li, Kun and Marin, Riccardo and Melzi, Simone and Yang, Jingyu},
  journal={Computers \& Graphics},
  volume={92},
  pages={28--43},
  year={2020},
  publisher={Elsevier}
}

@article{thompson2020shrec, 
  year     = {2020}, 
  title    = {{SHREC} 2020: Retrieval of digital surfaces with similar geometric reliefs}, 
  author   = {Thompson, Elia Moscoso and Biasotti, Silvia and Giachetti, Andrea and Tortorici, Claudio and Werghi, Naoufel and Obeid, Ahmad Shaker and Berretti, Stefano and Nguyen-Dinh, Hoang-Phuc and Le, Minh-Quan and Nguyen, Hai-Dang and Tran, Minh-Triet and Gigli, Leonardo and Velasco-Forero, Santiago and Marcotegui, Beatriz and Sipiran, Ivan and Bustos, Benjamin and Romanelis, Ioannis and Fotis, Vlassis and Arvanitis, Gerasimos and Moustakas, Konstantinos and Otu, Ekpo and Zwiggelaar, Reyer and Hunter, David and Liu, Yonghuai and Arteaga, Yoko and Luxman, Ramamoorthy}, 
  journal  = {Computers \& Graphics}, 
  issn     = {0097-8493}, 
  doi      = {10.1016/j.cag.2020.07.011}, 
  pages    = {199--218}, 
  volume   = {91}
}

@inproceedings{biasotti2021shrec,
booktitle = {Eurographics Workshop on 3D Object Retrieval},
editor = {Biasotti, Silvia and Dyke, Roberto M. and Lai, Yukun and Rosin, Paul L. and Veltkamp, Remco C.},
title = {{SHREC 2021: Surface-based Protein Domains Retrieval}},
author = {Langenfeld, Florent and Aderinwale, Tunde and Windal, Feryal and Melkemi, Mahmoud and Otu, Ekpo and Zwiggelaar, Reyer and Hunter, David and Liu, Yonghuai and Sirugue, Léa and Nguyen, Huu-Nghia H. and Nguyen, Tuan-Duy H. and Nguyen–Truong, Vinh-Thuyen and Christoffer, Charles and Le, Danh and Nguyen, Hai-Dang and Tran, Minh-Triet and Montès, Matthieu and Shin, Woong-Hee and Terashi, Genki and Wang, Xiao and Kihara, Daisuke and Benhabiles, Halim and Hammoudi, Karim and Cabani, Adnane},
year = {2021},
publisher = {The Eurographics Association},
ISSN = {1997-0471},
ISBN = {978-3-03868-137-3},
DOI = {10.2312/3dor.20211308}
}

@article{myles2014robust, 
  year     = {2014}, 
  title    = {Robust field-aligned global parametrization}, 
  author   = {Myles, Ashish and Pietroni, Nico and Zorin, Denis}, 
  journal  = {{ACM} Transactions on Graphics}, 
  issn     = {0730-0301}, 
  doi      = {10.1145/2601097.2601154}, 
  pages    = {1--14}, 
  number   = {4}, 
  volume   = {33}
}

@article{sumner2004deformation, 
  year     = {2004}, 
  title    = {Deformation transfer for triangle meshes}, 
  author   = {Sumner, Robert W and Popović, Jovan}, 
  journal  = {{ACM} Transactions on Graphics}, 
  issn     = {0730-0301}, 
  doi      = {10.1145/1015706.1015736}, 
  pages    = {399--405}, 
  number   = {3}, 
  volume   = {23}
}

@inproceedings{li20214dcomplete,
  title={4dcomplete: Non-rigid motion estimation beyond the observable surface},
  author={Li, Yang and Takehara, Hikari and Taketomi, Takafumi and Zheng, Bo and Nie{\ss}ner, Matthias},
  booktitle={Proceedings of the IEEE/CVF International Conference on Computer Vision},
  pages={12706--12716},
  year={2021}
}

@InProceedings{koch2019abc,
author = {Koch, Sebastian and Matveev, Albert and Jiang, Zhongshi and Williams, Francis and Artemov, Alexey and Burnaev, Evgeny and Alexa, Marc and Zorin, Denis and Panozzo, Daniele},
title = {ABC: A Big CAD Model Dataset For Geometric Deep Learning},
booktitle = {The IEEE Conference on Computer Vision and Pattern Recognition (CVPR)},
month = {June},
year = {2019}
}

@inproceedings{bogo2014faust,
  title={FAUST: Dataset and evaluation for 3D mesh registration},
  author={Bogo, Federica and Romero, Javier and Loper, Matthew and Black, Michael J},
  booktitle={Proceedings of the IEEE conference on computer vision and pattern recognition},
  pages={3794--3801},
  year={2014}
}

@inproceedings{poulenard2019effective,
  title={Effective rotation-invariant point cnn with spherical harmonics kernels},
  author={Poulenard, Adrien and Rakotosaona, Marie-Julie and Ponty, Yann and Ovsjanikov, Maks},
  booktitle={2019 International Conference on 3D Vision (3DV)},
  pages={47--56},
  year={2019},
  organization={IEEE}
}

@inproceedings{levy2006laplace,
  title={Laplace-beltrami eigenfunctions towards an algorithm that" understands" geometry},
  author={L{\'e}vy, Bruno},
  booktitle={IEEE International Conference on Shape Modeling and Applications 2006 (SMI'06)},
  pages={13--13},
  year={2006},
  organization={IEEE}
}

@inproceedings{rustamov2007laplace,
    booktitle = {Geometry Processing},
    editor = {Alexander Belyaev and Michael Garland},
    title = {{Laplace-Beltrami Eigenfunctions for Deformation Invariant Shape Representation}},
    author = {Rustamov, Raif M.},
    year = {2007},
    publisher = {The Eurographics Association},
    ISSN = {1727-8384},
    ISBN = {978-3-905673-46-3},
    DOI = {/10.2312/SGP/SGP07/225-233}
}

@article{ovsjanikov2012functional, 
  year    = {2012}, 
  title   = {Functional maps: a flexible representation of maps between shapes}, 
  author  = {Ovsjanikov, Maks and Ben-Chen, Mirela and Solomon, Justin and Butscher, Adrian and Guibas, Leonidas}, 
  journal = {{ACM} Transactions on Graphics}, 
  issn    = {0730-0301}, 
  doi     = {10.1145/2185520.2335381}, 
  pages   = {1--11}, 
  number  = {4}, 
  volume  = {31}
}

@article{pinkall1993computing, 
  year     = {1993}, 
  title    = {Computing Discrete Minimal Surfaces and Their Conjugates}, 
  author   = {Pinkall, Ulrich and Polthier, Konrad}, 
  journal  = {Experimental Mathematics}, 
  issn     = {1058-6458}, 
  doi      = {10.1080/10586458.1993.10504266}, 
  pages    = {15--36}, 
  number   = {1}, 
  volume   = {2}
}

@article{meyer2003visualization, 
  year     = {2003}, 
  title    = {Visualization and Mathematics {III}}, 
  author   = {Meyer, Mark and Desbrun, Mathieu and Schröder, Peter and Barr, Alan H}, 
  journal  = {Mathematics and Visualization}, 
  issn     = {1612-3786}, 
  doi      = {10.1007/978-3-662-05105-4_2}, 
  pages    = {35--57}
}

@article{aubry2011wave, 
  year     = {2011}, 
  title    = {The wave kernel signature: A quantum mechanical approach to shape analysis}, 
  author   = {Aubry, Mathieu and Schlickewei, Ulrich and Cremers, Daniel}, 
  journal  = {2011 {IEEE} International Conference on Computer Vision Workshops ({ICCV} Workshops)}, 
  doi      = {10.1109/{ICCVW}.2011.6130444}, 
  pages    = {1626--1633}, 
  volume   = {1}
}

@article{tenenbaum2000global, 
  year     = {2000}, 
  title    = {A Global Geometric Framework for Nonlinear Dimensionality Reduction}, 
  author   = {Tenenbaum, Joshua B and Silva, Vin de and Langford, John C}, 
  journal  = {Science}, 
  issn     = {0036-8075}, 
  doi      = {10.1126/science.290.5500.2319}, 
  pmid     = {11125149}, 
  pages    = {2319--2323}, 
  number   = {5500}, 
  volume   = {290}
}

@article{roweis2000nonlinear,
  title={Nonlinear dimensionality reduction by locally linear embedding},
  author={Roweis, Sam T and Saul, Lawrence K},
  journal={science},
  volume={290},
  number={5500},
  pages={2323--2326},
  year={2000},
  publisher={American Association for the Advancement of Science}
}

@article{belkin2003laplacian, 
  year     = {2003}, 
  title    = {Laplacian Eigenmaps for Dimensionality Reduction and Data Representation}, 
  author   = {Belkin, Mikhail and Niyogi, Partha}, 
  journal  = {Neural Computation}, 
  issn     = {0899-7667}, 
  doi      = {10.1162/089976603321780317}, 
  pages    = {1373--1396}, 
  number   = {6}, 
  volume   = {15}
}

@article{aflalo2015optimality, 
  year     = {2015}, 
  title    = {On the Optimality of Shape and Data Representation in the Spectral Domain}, 
  author   = {Aflalo, Yonathan and Brezis, Haim and Kimmel, Ron}, 
  journal  = {{SIAM} Journal on Imaging Sciences}, 
  doi      = {10.1137/140977680}, 
  pages    = {1141--1160}, 
  number   = {2}, 
  volume   = {8}
}

@inproceedings{lescoat2020spectral,
  title={Spectral mesh simplification},
  author={Lescoat, Thibault and Liu, Hsueh-Ti Derek and Thiery, Jean-Marc and Jacobson, Alec and Boubekeur, Tamy and Ovsjanikov, Maks},
  booktitle={Computer Graphics Forum},
  volume={39},
  number={2},
  pages={315--324},
  year={2020},
  organization={Wiley Online Library}
}

@article{bruna2013spectral,
  title={Spectral networks and locally connected networks on graphs},
  author={Bruna, Joan and Zaremba, Wojciech and Szlam, Arthur and LeCun, Yann},
  journal={International Conference on Learning Representations (ICLR)},
  year={2013}
}

@inproceedings{bensaid2023partial,
  title={Partial matching of nonrigid shapes by learning piecewise smooth functions},
  author={Bensa{\"\i}d, David and Rotstein, Noam and Goldenstein, Nelson and Kimmel, Ron},
  booktitle={Computer Graphics Forum},
  volume={42},
  number={5},
  pages={e14913},
  year={2023},
  organization={Wiley Online Library}
}

@article{sharp2022diffusionnet,
  title={Diffusionnet: Discretization agnostic learning on surfaces},
  author={Sharp, Nicholas and Attaiki, Souhaib and Crane, Keenan and Ovsjanikov, Maks},
  journal={ACM Transactions on Graphics (TOG)},
  volume={41},
  number={3},
  pages={1--16},
  year={2022},
  publisher={ACM New York, NY}
}

@article{rowan2025solving, 
  year     = {2025}, 
  title    = {Solving engineering eigenvalue problems with neural networks using the Rayleigh quotient}, 
  author   = {Rowan, Conor and Evans, John and Maute, Kurt and Doostan, Alireza}, 
  journal  = {{arXiv}}, 
  doi      = {10.48550/{arXiv}.2506.04375}, 
  eprint   = {2506.04375}
}

@article{benshaul2023deep, 
  year     = {2023}, 
  title    = {Deep Learning Solution of the Eigenvalue Problem for Differential Operators}, 
  author   = {Ben-Shaul, Ido and Bar, Leah and Fishelov, Dalia and Sochen, Nir}, 
  journal  = {Neural Computation}, 
  issn     = {0899-7667}, 
  doi      = {10.1162/neco_a_01583}, 
  pmid     = {37037040}, 
  pages    = {1100--1134}, 
  number   = {6}, 
  volume   = {35}
}

@article{han2020solving, 
  year     = {2020}, 
  title    = {Solving high-dimensional eigenvalue problems using deep neural networks: A diffusion Monte Carlo like approach}, 
  author   = {Han, Jiequn and Lu, Jianfeng and Zhou, Mo}, 
  journal  = {Journal of Computational Physics}, 
  issn     = {0021-9991}, 
  doi      = {10.1016/j.jcp.2020.109792}, 
  eprint   = {2002.02600}, 
  pages    = {109792}, 
  volume   = {423}
}

@article{mcinnes2018umap, 
  year     = {2018}, 
  title    = {{UMAP}: Uniform Manifold Approximation and Projection for Dimension Reduction}, 
  author   = {{McInnes}, Leland and Healy, John and Melville, James}, 
  journal  = {{arXiv}}, 
  doi      = {10.48550/{arXiv}.1802.03426}, 
  eprint   = {1802.03426}
}

@inproceedings{andreux2014anisotropic,
  title={Anisotropic Laplace-Beltrami operators for shape analysis},
  author={Andreux, Mathieu and Rodola, Emanuele and Aubry, Mathieu and Cremers, Daniel},
  booktitle={European conference on computer vision},
  pages={299--312},
  year={2014},
  organization={Springer}
}

@inproceedings{boscaini2016anisotropic,
  title={Anisotropic diffusion descriptors},
  author={Boscaini, Davide and Masci, Jonathan and Rodol{\`a}, Emanuele and Bronstein, Michael M and Cremers, Daniel},
  booktitle={Computer Graphics Forum},
  volume={35},
  number={2},
  pages={431--441},
  year={2016},
  organization={Wiley Online Library}
}

@inproceedings{weber2024finsler,
  title={Finsler-{L}aplace-{B}eltrami operators with application to shape analysis},
  author={Weber, Simon and Dag{\`e}s, Thomas and Gao, Maolin and Cremers, Daniel},
  booktitle={Proceedings of the IEEE/CVF Conference on Computer Vision and Pattern Recognition},
  pages={3131--3140},
  year={2024}
}

@article{cauchy1829equation,
  title={Sur l’{\'e}quation {\`a} l’aide de laquelle on d{\'e}termine les in{\'e}galit{\'e}s s{\'e}culaires des mouvements des plan{\`e}tes},
  author={Cauchy, Augustin-Louis},
  journal={Oeuvres Compl{\`e}tes (IIeme S{\'e}rie)},
  volume={9},
  pages={174--195},
  year={1829}
}

@book{beltrami1868saggio,
  title={Saggio di interpretazione della geometria non-euclidea},
  author={Beltrami, Eugenio},
  year={1868},
  publisher={Stab. Tip. De Angelis}
}

@book{rosenberg1997laplacian,
  title={The Laplacian on a Riemannian manifold: an introduction to analysis on manifolds},
  author={Rosenberg, Steven},
  number={31},
  year={1997},
  publisher={Cambridge University Press}
}

@inproceedings{wardetzky2007discrete,
  title={Discrete {L}aplace operators: no free lunch},
  author={Wardetzky, Max and Mathur, Saurabh and K{\"a}lberer, Felix and Grinspun, Eitan},
  booktitle={Symposium on Geometry processing},
  volume={33},
  pages={37},
  year={2007},
  organization={Aire-la-Ville, Switzerland}
}

@article{kirchhoff1847ueber,
  title={Ueber die Aufl{\"o}sung der Gleichungen, auf welche man bei der Untersuchung der linearen Vertheilung galvanischer Str{\"o}me gef{\"u}hrt wird},
  author={Kirchhoff, Gustav},
  journal={Annalen der Physik},
  volume={148},
  number={12},
  pages={497--508},
  year={1847},
  publisher={WILEY-VCH Verlag Leipzig}
}

@book{chung1997spectral,
  title={Spectral graph theory},
  author={Chung, Fan RK},
  volume={92},
  year={1997},
  publisher={American Mathematical Soc.}
}

@article{wang2018steklov,
  title={Steklov spectral geometry for extrinsic shape analysis},
  author={Wang, Yu and Ben-Chen, Mirela and Polterovich, Iosif and Solomon, Justin},
  journal={ACM Transactions on Graphics (TOG)},
  volume={38},
  number={1},
  pages={1--21},
  year={2018},
  publisher={ACM New York, NY, USA}
}

@article{zhou2025laplace,
  title={Laplace-{B}eltrami Operator for Gaussian Splatting},
  author={Zhou, Hongyu and L{\"a}hner, Zorah},
  journal={arXiv preprint arXiv:2502.17531},
  year={2025}
}

@book{bronstein2008numerical,
  title={Numerical geometry of non-rigid shapes},
  author={Bronstein, Alexander M and Bronstein, Michael M and Kimmel, Ron},
  year={2008},
  publisher={Springer Science \& Business Media}
}

@inproceedings{wetzler2013laplace,
  title={The {L}aplace-{B}eltrami operator: a ubiquitous tool for image and shape processing},
  author={Wetzler, Aaron and Aflalo, Yonathan and Dubrovina, Anastasia and Kimmel, Ron},
  booktitle={International Symposium on Mathematical Morphology and Its Applications to Signal and Image Processing},
  pages={302--316},
  year={2013},
  organization={Springer}
}

@inproceedings{chuang2009estimating,
  title={Estimating the Laplace-Beltrami operator by restricting 3d functions},
  author={Chuang, Ming and Luo, Linjie and Brown, Benedict J and Rusinkiewicz, Szymon and Kazhdan, Michael},
  booktitle={Computer graphics forum},
  volume={28},
  number={5},
  pages={1475--1484},
  year={2009},
  organization={Wiley Online Library}
}

@article{zhang2025topology,
  title={Topology-controlled Laplace--Beltrami operator on point clouds based on persistent homology},
  author={Zhang, Ao and Fang, Qing and Zhou, Peng and Fu, Xiao-Ming},
  journal={Graphical Models},
  volume={139},
  pages={101261},
  year={2025},
  publisher={Elsevier}
}

@article{chazal2016data,
  title={Data driven estimation of Laplace-Beltrami operator},
  author={Chazal, Fr{\'e}d{\'e}ric and Giulini, Ilaria and Michel, Bertrand},
  journal={Advances in Neural Information Processing Systems},
  volume={29},
  year={2016}
}

@article{an2025ai,
  title={An AI Approach for Learning the Spectrum of the Laplace-Beltrami Operator},
  author={An, Yulin and del Castillo, Enrique},
  journal={arXiv preprint arXiv:2507.07073},
  year={2025}
}

@article{peoples2024higher,
  title={A Higher Order Local Mesh Method for Approximating Laplacians on Unknown Manifolds},
  author={Peoples, John Wilson and Harlim, John},
  journal={arXiv preprint arXiv:2405.15735},
  year={2024}
}

@article{trillos2025minimax,
  title={Minimax Rates for the Estimation of Eigenpairs of Weighted Laplace-Beltrami Operators on Manifolds},
  author={Trillos, Nicol{\'a}s Garc{\'\i}a and Li, Chenghui and Venkatraman, Raghavendra},
  journal={arXiv preprint arXiv:2506.00171},
  year={2025}
}

@article{mhaskar2025learning,
  title={Learning on manifolds without manifold learning},
  author={Mhaskar, Hrushikesh N and O’Dowd, Ryan},
  journal={Neural Networks},
  volume={181},
  pages={106759},
  year={2025},
  publisher={Elsevier}
}

@article{chen2024learning,
  title={Learning neural operators on riemannian manifolds},
  author={Chen, Gengxiang and Liu, Xu and Meng, Qinglu and Chen, Lu and Liu, Changqing and Li, Yingguang},
  journal={National Science Open},
  volume={3},
  number={6},
  pages={20240001},
  year={2024},
  publisher={China Science Publishing \& Media Ltd. and EDP Sciences}
}

@article{aflalo2013spectral,
  title={Spectral multidimensional scaling},
  author={Aflalo, Yonathan and Kimmel, Ron},
  journal={Proceedings of the National Academy of Sciences},
  volume={110},
  number={45},
  pages={18052--18057},
  year={2013},
  publisher={National Academy of Sciences}
}

@article{maaten2008visualizing,
  title={Visualizing data using t-SNE},
  author={Maaten, Laurens van der and Hinton, Geoffrey},
  journal={Journal of machine learning research},
  volume={9},
  number={Nov},
  pages={2579--2605},
  year={2008}
}

@article{van2014accelerating,
  title={Accelerating t-SNE using tree-based algorithms},
  author={Van Der Maaten, Laurens},
  journal={The journal of machine learning research},
  volume={15},
  number={1},
  pages={3221--3245},
  year={2014},
  publisher={JMLR. org}
}

@inproceedings{coates2011analysis,
  title={An analysis of single-layer networks in unsupervised feature learning},
  author={Coates, Adam and Ng, Andrew and Lee, Honglak},
  booktitle={Proceedings of the fourteenth international conference on artificial intelligence and statistics},
  pages={215--223},
  year={2011},
  organization={JMLR Workshop and Conference Proceedings}
}

@techreport{griffin2007caltech,
  title={Caltech-256 object category dataset},
  author={Griffin, Gregory and Holub, Alex and Perona, Pietro and others},
  year={2007},
  institution={Technical Report 7694, California Institute of Technology Pasadena}
}

@article{oquab2023dinov2,
  title={Dinov2: Learning robust visual features without supervision},
  author={Oquab, Maxime and Darcet, Timoth{\'e}e and Moutakanni, Th{\'e}o and Vo, Huy and Szafraniec, Marc and Khalidov, Vasil and Fernandez, Pierre and Haziza, Daniel and Massa, Francisco and El-Nouby, Alaaeldin and others},
  journal={arXiv preprint arXiv:2304.07193},
  year={2023}
}

@inproceedings{radford2021learning,
  title={Learning transferable visual models from natural language supervision},
  author={Radford, Alec and Kim, Jong Wook and Hallacy, Chris and Ramesh, Aditya and Goh, Gabriel and Agarwal, Sandhini and Sastry, Girish and Askell, Amanda and Mishkin, Pamela and Clark, Jack and others},
  booktitle={International conference on machine learning},
  pages={8748--8763},
  year={2021},
  organization={PmLR}
}

@inproceedings{dages2025finsler,
  title={Finsler Multi-Dimensional Scaling: Manifold Learning for Asymmetric Dimensionality Reduction and Embedding},
  author={Dag{\`e}s, Thomas and Weber, Simon and Lin, Ya-Wei Eileen and Talmon, Ronen and Cremers, Daniel and Lindenbaum, Michael and Bruckstein, Alfred M and Kimmel, Ron},
  booktitle={Proceedings of the Computer Vision and Pattern Recognition Conference},
  pages={25842--25853},
  year={2025}
}

@inproceedings{dages2025metric,
  title={Metric Convolutions: A Unifying Theory to Adaptive Image Convolutions},
  author={Dag{\`e}s, Thomas and Lindenbaum, Michael and Bruckstein, Alfred M},
  booktitle={Proceedings of the IEEE/CVF International Conference on Computer Vision},
  pages={13974--13984},
  year={2025}
}

@article{barthelme2013natural,
  title={A natural Finsler-Laplace operator},
  author={Barthelm{\'e}, Thomas},
  journal={Israel Journal of Mathematics},
  volume={196},
  number={1},
  pages={375--412},
  year={2013},
  publisher={Springer}
}

@article{ohta2009heat,
  title={Heat flow on Finsler manifolds},
  author={Ohta, Shin-ichi and Sturm, Karl-Theodor},
  journal={Communications on Pure and Applied Mathematics: A Journal Issued by the Courant Institute of Mathematical Sciences},
  volume={62},
  number={10},
  pages={1386--1433},
  year={2009},
  publisher={Wiley Online Library}
}

@article{boscaini2016learning,
  title={Learning shape correspondence with anisotropic convolutional neural networks},
  author={Boscaini, Davide and Masci, Jonathan and Rodol{\`a}, Emanuele and Bronstein, Michael},
  journal={Advances in neural information processing systems},
  volume={29},
  year={2016}
}

@article{aflalo2013scale,
  title={Scale invariant geometry for nonrigid shapes},
  author={Aflalo, Yonathan and Kimmel, Ron and Raviv, Dan},
  journal={SIAM Journal on Imaging Sciences},
  volume={6},
  number={3},
  pages={1579--1597},
  year={2013},
  publisher={SIAM}
}

@article{raviv2015affine,
  title={Affine invariant non-rigid shape analysis},
  author={Raviv, Dan and Kimmel, Ron},
  journal={Int. J. Comput. Vision, submitted},
  year={2015}
}

@inproceedings{halimi2018self,
  title={Self functional maps},
  author={Halimi, Oshri and Kimmel, Ron},
  booktitle={2018 International Conference on 3D Vision (3DV)},
  pages={710--718},
  year={2018},
  organization={IEEE}
}

@article{sela2015computational,
  title={Computational caricaturization of surfaces},
  author={Sela, Matan and Aflalo, Yonathan and Kimmel, Ron},
  journal={Computer Vision and Image Understanding},
  volume={141},
  pages={1--17},
  year={2015},
  publisher={Elsevier}
}

@inproceedings{bracha2024unsupervised,
  title={On unsupervised partial shape correspondence},
  author={Bracha, Amit and Dag{\`e}s, Thomas and Kimmel, Ron},
  booktitle={Proceedings of the Asian Conference on Computer Vision},
  pages={4488--4504},
  year={2024}
}

@inproceedings{bracha2024wormhole,
 author = {Bracha, Amit and Dag\`{e}s, Thomas and Kimmel, Ron},
 booktitle = {Advances in Neural Information Processing Systems},
 editor = {A. Globerson and L. Mackey and D. Belgrave and A. Fan and U. Paquet and J. Tomczak and C. Zhang},
 pages = {131247--131277},
 publisher = {Curran Associates, Inc.},
 title = {Wormhole Loss for Partial Shape Matching},
 volume = {37},
 year = {2024}
}

@article{kingma2014adam,
  title={Adam: A method for stochastic optimization},
  author={Kingma, Diederik P},
  journal={arXiv preprint arXiv:1412.6980},
  year={2014}
}

@inproceedings{vinh2009information,
  title={Information theoretic measures for clusterings comparison: is a correction for chance necessary?},
  author={Vinh, Nguyen Xuan and Epps, Julien and Bailey, James},
  booktitle={Proceedings of the 26th annual international conference on machine learning},
  pages={1073--1080},
  year={2009}
}

@article{hubert1985comparing,
  title={Comparing partitions},
  author={Hubert, Lawrence and Arabie, Phipps},
  journal={Journal of classification},
  volume={2},
  number={1},
  pages={193--218},
  year={1985},
  publisher={Springer}
}

@article{pearson1901pca,
  title={On lines and planes of closest fit to systems of points in space},
  author={Pearson, Karl},
  journal={The London, Edinburgh, and Dublin philosophical magazine and journal of science},
  volume={2},
  number={11},
  pages={559--572},
  year={1901},
  publisher={Taylor \& Francis}
}

@article{hotelling1933analysis,
  title={Analysis of a complex of statistical variables into principal components.},
  author={Hotelling, Harold},
  journal={Journal of educational psychology},
  volume={24},
  number={6},
  pages={417},
  year={1933},
  publisher={Warwick \& York}
}

@article{shepard1962analysis,
  title={The analysis of proximities: multidimensional scaling with an unknown distance function. I.},
  author={Shepard, Roger N},
  journal={Psychometrika},
  volume={27},
  number={2},
  pages={125--140},
  year={1962},
  publisher={Springer}
}

@article{kruskal1964multidimensional,
  title={Multidimensional scaling by optimizing goodness of fit to a nonmetric hypothesis},
  author={Kruskal, Joseph B},
  journal={Psychometrika},
  volume={29},
  number={1},
  pages={1--27},
  year={1964},
  publisher={Springer}
}

@article{fischl1999cortical,
  title={Cortical surface-based analysis: II: inflation, flattening, and a surface-based coordinate system},
  author={Fischl, Bruce and Sereno, Martin I and Dale, Anders M},
  journal={Neuroimage},
  volume={9},
  number={2},
  pages={195--207},
  year={1999},
  publisher={Elsevier}
}

@article{wandell2000visualization,
  title={Visualization and measurement of the cortical surface},
  author={Wandell, Brian A and Chial, Suelika and Backus, Benjamin T},
  journal={Journal of cognitive neuroscience},
  volume={12},
  number={5},
  pages={739--752},
  year={2000},
  publisher={MIT Press One Rogers Street, Cambridge, MA 02142-1209, USA journals-info~…}
}

@inproceedings{weinberger2005nonlinear,
  title={Nonlinear dimensionality reduction by semidefinite programming and kernel matrix factorization},
  author={Weinberger, Kilian and Packer, Benjamin and Saul, Lawrence},
  booktitle={International Workshop on Artificial Intelligence and Statistics},
  pages={381--388},
  year={2005},
  organization={PMLR}
}

@article{weinberger2006unsupervised,
  title={Unsupervised learning of image manifolds by semidefinite programming},
  author={Weinberger, Kilian Q and Saul, Lawrence K},
  journal={International journal of computer vision},
  volume={70},
  pages={77--90},
  year={2006},
  publisher={Springer}
}

@inproceedings{brand2005nonrigid,
  title={Nonrigid embeddings for dimensionality reduction},
  author={Brand, Matthew},
  booktitle={European Conference on Machine Learning},
  pages={47--59},
  year={2005},
  organization={Springer}
}

@inproceedings{funke2020low,
  title={Low-Dimensional Statistical Manifold Embedding of Directed Graphs},
  author={Funke, Thorben and Guo, Tian and Lancic, Alen and Antulov-Fantulin, Nino},
  booktitle={8th International Conference on Learning Representations (ICLR 2020)},
  volume={3},
  pages={2018--2035},
  year={2020},
  organization={Curran}
}

@article{donoho2003hessian,
  title={Hessian eigenmaps: {L}ocally linear embedding techniques for high-dimensional data},
  author={Donoho, David L and Grimes, Carrie},
  journal={Proceedings of the National Academy of Sciences},
  volume={100},
  number={10},
  pages={5591--5596},
  year={2003},
  publisher={National Acad Sciences}
}

@article{zhang2006mlle,
  title={{MLLE}: {M}odified locally linear embedding using multiple weights},
  author={Zhang, Zhenyue and Wang, Jing},
  journal={Advances in neural information processing systems},
  volume={19},
  year={2006}
}

@article{zhang2004principal,
  title={Principal manifolds and nonlinear dimensionality reduction via tangent space alignment},
  author={Zhang, Zhenyue and Zha, Hongyuan},
  journal={SIAM journal on scientific computing},
  volume={26},
  number={1},
  pages={313--338},
  year={2004},
  publisher={SIAM}
}

@article{belkin2001laplacian,
  title={Laplacian eigenmaps and spectral techniques for embedding and clustering},
  author={Belkin, Mikhail and Niyogi, Partha},
  journal={Advances in neural information processing systems},
  volume={14},
  year={2001}
}

@article{coifman2005geometric,
  title={Geometric diffusions as a tool for harmonic analysis and structure definition of data: {D}iffusion maps},
  author={Coifman, Ronald R and Lafon, Stephane and Lee, Ann B and Maggioni, Mauro and Nadler, Boaz and Warner, Frederick and Zucker, Steven W},
  journal={Proceedings of the national academy of sciences},
  volume={102},
  number={21},
  pages={7426--7431},
  year={2005},
  publisher={National Acad Sciences}
}

@inproceedings{lin2023hyperbolic,
  title={Hyperbolic diffusion embedding and distance for hierarchical representation learning},
  author={Lin, Ya-Wei Eileen and Coifman, Ronald R and Mishne, Gal and Talmon, Ronen},
  booktitle={International Conference on Machine Learning},
  pages={21003--21025},
  year={2023},
  organization={PMLR}
}

@article{lin2024tree,
  title={Tree-Wasserstein distance for high dimensional data with a latent feature hierarchy},
  author={Lin, Ya-Wei Eileen and Coifman, Ronald R and Mishne, Gal and Talmon, Ronen},
  journal={arXiv preprint arXiv:2410.21107},
  year={2024}
}

@inproceedings{suzuki2019hyperbolic,
  title={Hyperbolic disk embeddings for directed acyclic graphs},
  author={Suzuki, Ryota and Takahama, Ryusuke and Onoda, Shun},
  booktitle={International Conference on Machine Learning},
  pages={6066--6075},
  year={2019},
  organization={PMLR}
}

@article{gou2023discriminative,
  title={Discriminative and geometry-preserving adaptive graph embedding for dimensionality reduction},
  author={Gou, Jianping and Yuan, Xia and Xue, Ya and Du, Lan and Yu, Jiali and Xia, Shuyin and Zhang, Yi},
  journal={Neural Networks},
  volume={157},
  pages={364--376},
  year={2023},
  publisher={Elsevier}
}

@article{venna2010information,
  title={Information retrieval perspective to nonlinear dimensionality reduction for data visualization.},
  author={Venna, Jarkko and Peltonen, Jaakko and Nybo, Kristian and Aidos, Helena and Kaski, Samuel},
  journal={Journal of Machine Learning Research},
  volume={11},
  number={2},
  year={2010}
}

@article{martin2005visualizing,
  title={Visualizing asymmetric proximities with {SOM} and {MDS} models},
  author={Mart{\'\i}n-Merino, Manuel and Mu{\~n}oz, Alberto},
  journal={Neurocomputing},
  volume={63},
  pages={171--192},
  year={2005},
  publisher={Elsevier}
}

@inproceedings{tang2016visualizing,
  title={Visualizing large-scale and high-dimensional data},
  author={Tang, Jian and Liu, Jingzhou and Zhang, Ming and Mei, Qiaozhu},
  booktitle={Proceedings of the 25th international conference on world wide web},
  pages={287--297},
  year={2016}
}

@article{liu2024curvature,
  title={Curvature Augmented Manifold Embedding and Learning},
  author={Liu, Yongming},
  journal={arXiv preprint arXiv:2403.14813},
  year={2024}
}

@article{scholkopf1998nonlinear,
  title={Nonlinear component analysis as a kernel eigenvalue problem},
  author={Sch{\"o}lkopf, Bernhard and Smola, Alexander and M{\"u}ller, Klaus-Robert},
  journal={Neural computation},
  volume={10},
  number={5},
  pages={1299--1319},
  year={1998},
  publisher={MIT Press}
}

@article{schwartz1989numerical,
  author={Schwartz, E.L. and Shaw, A. and Wolfson, E.},
  journal={IEEE Transactions on Pattern Analysis and Machine Intelligence}, 
  title={A numerical solution to the generalized mapmaker's problem: {F}lattening nonconvex polyhedral surfaces}, 
  year={1989},
  volume={11},
  number={9},
  pages={1005-1008},
}

@article{bronstein2006generalized,
  title={Generalized multidimensional scaling: a framework for isometry-invariant partial surface matching},
  author={Bronstein, Alexander M and Bronstein, Michael M and Kimmel, Ron},
  journal={Proceedings of the National Academy of Sciences},
  volume={103},
  number={5},
  pages={1168--1172},
  year={2006},
  publisher={National Academy of Sciences}
}

@article{rosman2010nonlinear,
  title={Nonlinear dimensionality reduction by topologically constrained isometric embedding},
  author={Rosman, Guy and Bronstein, Michael M and Bronstein, Alexander M and Kimmel, Ron},
  journal={International Journal of Computer Vision},
  volume={89},
  number={1},
  pages={56--68},
  year={2010},
  publisher={Springer}
}

@article{pai2022deep,
  title={Deep isometric maps},
  author={Pai, Gautam and Bronstein, Alex and Talmon, Ronen and Kimmel, Ron},
  journal={Image and Vision Computing},
  volume={123},
  pages={104461},
  year={2022},
  publisher={Elsevier}
}

@article{schwartz2019intrinsic,
  title={Intrinsic isometric manifold learning with application to localization},
  author={Schwartz, Ariel and Talmon, Ronen},
  journal={SIAM Journal on Imaging Sciences},
  volume={12},
  number={3},
  pages={1347--1391},
  year={2019},
  publisher={SIAM}
}

@inproceedings{joharinad2025isumap,
  title={IsUMap: Manifold Learning and Data Visualization leveraging Vietoris-Rips Filtrations},
  author={Joharinad, Parvaneh and Fahimi, Hannaneh and Barth, Lukas Silvester and Keck, Janis and Jost, J{\"u}rgen},
  booktitle={Proceedings of the AAAI Conference on Artificial Intelligence},
  volume={39},
  number={17},
  pages={17699--17706},
  year={2025}
}

@article{kim2024inductive,
  title={Inductive Global and Local Manifold Approximation and Projection},
  author={Kim, Jungeum and Wang, Xiao},
  journal={Transactions on Machine Learning Research},
  year={2024}
}

@article{zhou2016thingi10k,
  title={Thingi10k: A dataset of 10,000 3d-printing models},
  author={Zhou, Qingnan and Jacobson, Alec},
  journal={arXiv preprint arXiv:1605.04797},
  year={2016}
}

@inproceedings{rampini2019correspondence,
  title={Correspondence-free region localization for partial shape similarity via hamiltonian spectrum alignment},
  author={Rampini, Arianna and Tallini, Irene and Ovsjanikov, Maks and Bronstein, Alex M and Rodol{\`a}, Emanuele},
  booktitle={2019 International Conference on 3D Vision (3DV)},
  pages={37--46},
  year={2019},
  organization={IEEE}
}

@software{howard2019imagenette,
    title={Imagenette: A Smaller Subset of ImageNet},
    author={Jeremy Howard},
    year={2019},
    month={March},
    publisher = {GitHub},
    url = {https://github.com/fastai/imagenette}
}

@article{krizhevsky2009learning,
  title={Learning multiple layers of features from tiny images},
  author={Krizhevsky, Alex and Hinton, Geoffrey and others},
  year={2009},
  publisher={Toronto, ON, Canada}
}

@software{kaolinLibrary,
      author = {Fuji Tsang, Clement and Shugrina, Maria and Lafleche, Jean Francois and Perel, Or and Loop, Charles and Takikawa, Towaki and Modi, Vismay and Zook, Alexander and Wang, Jiehan and Chen, Wenzheng and Shen, Tianchang and Gao, Jun and Jatavallabhula, Krishna Murthy and Smith, Edward and Rozantsev, Artem and Fidler, Sanja and State, Gavriel and Gorski, Jason and Xiang, Tommy and Li, Jianing and Li, Michael and Lebaredian, Rev},
      title = {Kaolin: A Pytorch Library for Accelerating 3D Deep Learning Research},
      date = {2024-11-20},
      version = {0.18.0},
      url={\url{https://github.com/NVIDIAGameWorks/kaolin}}
}

@article{hendrycks2016gelu,
  title={Gaussian Error Linear Units ({GELU}s)},
  author={Hendrycks, Dan and Gimpel, Kevin},
  journal={arXiv preprint arXiv:1606.08415},
  year={2016}
}

@article{ba2016layer,
  title={Layer normalization},
  author={Ba, Jimmy Lei and Kiros, Jamie Ryan and Hinton, Geoffrey E},
  journal={arXiv preprint arXiv:1607.06450},
  year={2016}
}

@inproceedings{loshchilov2018decoupled,
    title={Decoupled Weight Decay Regularization},
    author={Ilya Loshchilov and Frank Hutter},
    booktitle={International Conference on Learning Representations},
    year={2019},
    url={https://openreview.net/forum?id=Bkg6RiCqY7},
}

@article{eldar1997farthest,
  title={The farthest point strategy for progressive image sampling},
  author={Eldar, Yuval and Lindenbaum, Michael and Porat, Moshe and Zeevi, Yehoshua Y},
  journal={IEEE transactions on image processing},
  volume={6},
  number={9},
  pages={1305--1315},
  year={1997},
  publisher={IEEE}
}

@article{gonzalez1985clustering,
  title={Clustering to minimize the maximum intercluster distance},
  author={Gonzalez, Teofilo F},
  journal={Theoretical computer science},
  volume={38},
  pages={293--306},
  year={1985},
  publisher={Elsevier}
}

@inproceedings{micikevicius2018mixed,
    title={Mixed Precision Training},
    author={Paulius Micikevicius and Sharan Narang and Jonah Alben and Gregory Diamos and Erich Elsen and David Garcia and Boris Ginsburg and Michael Houston and Oleksii Kuchaiev and Ganesh Venkatesh and Hao Wu},
    booktitle={International Conference on Learning Representations},
    year={2018},
    url={https://openreview.net/forum?id=r1gs9JgRZ},
}

@article{falcon2019pytorch,
  title={Pytorch lightning},
  author={Falcon, William A},
  journal={GitHub},
  year={2019}
}

@article{paszke2019pytorch,
  title={Pytorch: An imperative style, high-performance deep learning library},
  author={Paszke, Adam and Gross, Sam and Massa, Francisco and Lerer, Adam and Bradbury, James and Chanan, Gregory and Killeen, Trevor and Lin, Zeming and Gimelshein, Natalia and Antiga, Luca and others},
  journal={Advances in neural information processing systems},
  volume={32},
  year={2019}
}

@article{li2020ddp,
author = {Li, Shen and Zhao, Yanli and Varma, Rohan and Salpekar, Omkar and Noordhuis, Pieter and Li, Teng and Paszke, Adam and Smith, Jeff and Vaughan, Brian and Damania, Pritam and Chintala, Soumith},
title = {PyTorch distributed: experiences on accelerating data parallel training},
year = {2020},
issue_date = {August 2020},
publisher = {VLDB Endowment},
volume = {13},
number = {12},
issn = {2150-8097},
journal = {Proc. VLDB Endow.},
month = aug,
pages = {3005–3018},
numpages = {14}
}

@inproceedings{rosenberg2007v,
  title={V-measure: A conditional entropy-based external cluster evaluation measure},
  author={Rosenberg, Andrew and Hirschberg, Julia},
  booktitle={Proceedings of the 2007 joint conference on empirical methods in natural language processing and computational natural language learning (EMNLP-CoNLL)},
  pages={410--420},
  year={2007}
}

@article{fowlkes1983method,
  title={A method for comparing two hierarchical clusterings},
  author={Fowlkes, Edward B and Mallows, Colin L},
  journal={Journal of the American statistical association},
  volume={78},
  number={383},
  pages={553--569},
  year={1983},
  publisher={Taylor \& Francis}
}
}

\clearpage
\setcounter{page}{1}
\maketitlesupplementary

\appendix

\section{Additional Theory}

\subsection{PCA Alternative Formulations}
\label{sec: pca alternative formulations}

Without loss of generality, assume $\mathbb{E}(f) = 0$, as is the case for our signals uniformly distributed in 
$\mathcal{C}_L = \{f; \lVert f\rVert_L\le 1\}$. 
Denote $C = \mathbb{E}(ff^\top)$ as the covariance matrix of distributed functions $f$. 
In our case for \cref{thm:pca_continuous}, $C = R_L$ as $f\sim\mathcal{U}(\mathcal{S})$. 
The classical iterative optimization formulation for PCA is the following. PCA finds the orthonormal basis $b$ that maximizes iteratively over $k$ the Rayleigh quotient
\begin{equation}
    \max\limits_{\substack{b_k  \text{ s.t. } {\begin{cases}\scriptstyle \lVert b_k\rVert_2=1 \\ \scriptstyle b_k\perp \{b_1, \ldots, b_{k-1}\} \end{cases}}}} b_k^\top C b_k.
\end{equation}
The solution to this Rayleigh optimization, and thus to PCA, is given by the eigenvectors of $C$ in decreasing order of eigenvalues. 

Another alternative formulation exists, where the PCA optimization objective is given as the iterative minimization of the expected squared approximation error, or variance, in a basis $b$
\begin{equation}
    \min\limits_{\substack{b_k  \text{ s.t. } {\begin{cases}\scriptstyle \lVert b_k\rVert_2=1 \\ \scriptstyle b_k\perp \{b_1, \ldots, b_{k-1}\} \end{cases}}}} \mathbb{E}\left(\left\lVert f - \sum\limits_{i=1}^k\langle f,b_i\rangle b_i\right\rVert^2\right).
\end{equation}

Indeed, let us remind the classical link between the two formulations. Denoting $B_k\in\mathbb{R}^{d\times k}$ with columns $b_1,\dots, b_k$, with $B_k^\top B_k = I_k$, the expected squared approximation error can be rewritten as 
\begin{align}
    &\mathbb{E}\left(\left\lVert f - \sum\limits_{i=1}^k\langle f,b_i\rangle b_i\right\rVert^2\right) 
    = \mathbb{E}(\lVert (I - B_kB_k^\top)f \rVert^2)\\
    &\hspace{2em}= \mathrm{tr}(\mathbb{E}(f^\top (I - B_kB_k^\top)^\top (I - B_kB_k^\top) f)) \\
    &\hspace{2em}= \mathrm{tr}(\mathbb{E}(f^\top (I - B_kB_k^\top) f)) \\
    &\hspace{2em}= \mathrm{tr}(\mathbb{E}((I - B_kB_k^\top) ff^\top)) \\
    &\hspace{2em}= \mathrm{tr}((I - B_kB_k^\top) C) \\
    &\hspace{2em}= \mathrm{tr}(C) - \mathrm{tr}(B_kB_k^\top C) \\
    &\hspace{2em}= \mathrm{tr}(C) - \mathrm{tr}(B_k^\top C B_k) \\
    &\hspace{2em}= \mathrm{tr}(C) - \sum\limits_{i=1}^k b_i^\top C b_i
\end{align}
As $C$ is a constant, minimizing the expected squared approximation error is thus the same as maximizing the cumulative sum of Rayleigh quotients of $C$. Working iteratively over $k$, we get that the optimal solution $b$ for both formulations is given by the eigenvectors of $C$ in decreasing eigenvalue order. \qed

\subsection{Normalized Operators and Eigenstructures}
\label{sec: Normalized operators and eigenstructures}

We here generalize the relationship between the normalized symmetric Laplacian compared to the unnormalized one regarding its first eigenvector and how it provides the metric.

For any operator $A$, we can define an analogous normalized operator $A_{\text{norm}} = NA N^{-1}$, where $N$ is a matrix used for normalization. For instance $N = M^{\frac{1}{2}}$ for the normalized Laplacian $L_{\text{norm}}$ and $A = L$. Both operators $A$ and $A_{\text{norm}}$ share the same eigenvalues but different eigenvectors, related by the normalization matrix $N$. Indeed, if $\lambda_i$, $e_i^{\text{norm}}$ are eigenvalues and eigenvectors of $A_{\text{norm}}$, then $A N^{-1}e_i = N^{-1} NAN^{-1}e_i = N^{-1} A_{\text{norm}}e_i = \lambda_i N^{-1}e_i$, meaning that $e_i = N^{-1}e_i^{\text{norm}}$ are the eigenvectors of the unnormalized operator $A$. As such, choosing $N = M^{\frac{1}{2}}$ to be given by the metric as for the normalized Laplacian, the eigenvectors of the normalized Laplacian are related to the unnormalized one by the metric $e_i^{\text{norm}} = M^{\frac{1}{2}}e_i$. Note that $A_{\text{norm}}$ is symmetric if $A = N^{-2}\Sigma$ for symmetric $\Sigma$ and $N$, which is the case for the Laplacians where $\Sigma = S$ and $N = M^{\frac{1}{2}}$ is diagonal, as then $A_{\text{norm}} = N^{-1}\Sigma N^{-1}$, and so by the spectral theorem $e_i^{\text{norm}}$ is Euclidean-orthogonal whereas $e_i$ is $N^\top N$-orthogonal.

Of note, as for the Laplacians, if $e_1 = \mathbf{1}$ is in the nullspace of the unnormalized operator $A$, then $e_1^{\text{norm}} = N\mathbf{1}$ is in the nullspace of $A_{\text{norm}}$. Remarkably, if $N$ is a diagonal matrix (like $M^{\frac{1}{2}}$), if the nullspace is of dimension one, and if we learn to find the nullspace by learning the eigenbasis directly, then we can compute $N$ directly from it as $\text{diag}(N) = e_i^{\text{norm}}$. For most operators of interest, the assumptions on $A$ hold (for connected relationship graphs). This is because operators of interest are usually linear and differential, i.e.\ linear combinations with constant coefficients of the derivative operator $A = \sum_{|k|\le m}\alpha_k \partial^k$, usually without a 0-th order term, and constant functions are zeroed out by derivatives.

As such, our methodology learning a Euclidean orthogonal basis directly is well-suited to capture eigenstructures of implicit operators with Euclidean orthogonal eigenvectors and a single nullspace eigenvector dimension given by a diagonally scaled version of the $\mathbf{1}$ vector. The normalized Laplacian falls into this category, yet it is just one out of the many other interesting operators satisfying these assumptions.

\section{Additional Details on Our Framework}

\subsection{Projecting onto Optimal Reconstruction Bases}
\label{sec: Projecting onto Optimal Reconstruction Bases}

To approximate probe signals, we need to compute projections onto $\mathbf{Q}_k$. 
However, Euclidean, that is uniform, orthogonal projection will not account for the non-uniform geometry of the sampled manifold due to non-uniform sampling density. 
To overcome this issue, we resolve to the $M$-orthogonal projection onto the columns of $\mathbf{Q}_k$. 
As they are the output of a QR-decomposition, these vectors are Euclidean orthogonal and not $M$-orthogonal, yet, we want to
$M$-orthogonally
project onto their span. 
To do this, the $M$-orthogonally projected functions can be rewritten as
\begin{eqnarray}
    \mathbf{f}_{\text{proj},k} &=& \sum_{i=1}^k c_i^{(k)} \mathbf{q}_i \,\,\,= \,\,\,\mathbf{Q}_k \mathbf{c}^{(k)},
\end{eqnarray}
where $\mathbf{c}^{(k)} = [c_1^{(k)}, \ldots, c_k^{(k)}]^\top \in \mathbb{R}^k$ contains the projection coefficients.
The coefficients $\mathbf{c}^{(k)}$ are determined by requiring that the residual $(\mathbf{f} - \mathbf{f}_{\text{proj},k})$ is $M$-orthogonal to all basis vectors
\begin{eqnarray}
    (\mathbf{f} - \mathbf{f}_{\text{proj},k})^\top M \mathbf{q}_i \,=\, 0 \quad \text{ for } \,\, i \,= \,1, \ldots, k.
\end{eqnarray}
This orthogonality condition yields the system
\begin{eqnarray}
    \mathbf{q}_i^\top M \mathbf{f} &=& \sum_{j=1}^k c_j^{(k)} (\mathbf{q}_i^\top M \mathbf{q}_j).
\end{eqnarray}
In matrix form, this becomes
\begin{eqnarray}
    \mathbf{G}^{(k)} \mathbf{c}^{(k)} &=& \mathbf{b}^{(k)},
\end{eqnarray}
where $\mathbf{G}^{(k)} \in \mathbb{R}^{k \times k}$ is the Gram matrix with entries $G_{ij}^{(k)} = \mathbf{q}_i^\top M \mathbf{q}_j$, and $\mathbf{b}^{(k)} \in \mathbb{R}^k$ is a vector with entries $b_i = \mathbf{q}_i^\top M \mathbf{f}$.
Since $\mathbf{G}^{(k)} = \mathbf{Q}_k^\top M \mathbf{Q}_k$ is symmetric positive definite\footnote{As 
the columns of $\boldsymbol{Q}_k$
are linearly independent thanks to the QR-decomposition and $M$ is positive definite.}, we can efficiently solve this linear system using a differentiable solver to obtain the coefficients $\mathbf{c}^{(k)}$\footnote{Avoiding the explicit inversion in the formula $\mathbf{c} = (\mathbf{G}^{(k)})^{-1}\mathbf{b}$.}.
Crucially, $\mathbf{G}^{(k)}\in\mathbb{R}^k\times k$ is 
small,
where $k \ll n$ is the number of eigenvectors used for projection, making the system computationally efficient and independent of the input size $n$. 
Note that $\mathbf{c}^{(k)}$ is not the truncation of $\mathbf{c}^{(K)}$ to its first $k$ terms since $\mathbf{Q}_k$ is not orthogonal for the used $M$-weighted scalar product.
These independent systems are batched and solved concurrently for all $k\le K$.

\subsection{Replacing the $M$-norm with the Euclidean norm for Approximation Errors}
\label{sec: Replacing the $M$-norm with the Euclidean norm for Approximation Errors}

While the theoretical framework calls for the $M$-norm in the error computation (in both \cref{thm:aflalo_minmax,thm:pca_continuous}), we deliberately use the $2$-norm in our loss (\cref{eq: reconstruction loss}). This hybrid approach, using $M$-weighted projection with $2$-norm error, creates an unsupervised mechanism for learning 
a density-related
mass matrix. Since the $2$-norm weights all points equally regardless of sampling density, regions with different sampling densities contribute differently to the loss, implicitly encouraging the network to learn mass elements 
connected to the ambient space's Euclidean-induced volume (or area on surface manifolds) of the samples.
We also experimented with using the $M$-norm for error computation as prescribed by theory. However, this resulted in unstable training, numerical issues (NaNs), 
and even collapses of the metric to $0$ to naively minimise $\mathcal{L}_{\text{rec}}$. This confirms
the necessity of our hybrid approach.

\section{Additional Experimental Details and Results}
\label{sec: additional experimental details sup mat}

\subsection{Toy 1D Segment Manifold}

In this toy setting we work on a single point cloud.
We uniformly grid sample $n=100$ points $x_i=\tfrac{i}{n-1}$ for $i=0,\dots,n-1$. Probe signals $\mathbf{f}_i$ are generated by drawing i.i.d.\ noise $\mathbf{g}_i\in\mathbb{R}^n$ and applying a few steps of 1D Gaussian smoothing on the kNN graph with 16 neighbors, yielding smooth probes
$\mathbf{f}_i$.
Our 
lightweight MLP consists in 
3 hidden layers with width 64 and ReLU activations. We only compute the first $K=5$ basis vectors.
Wet train
over batches of $m = 500 000$ probe functions using the Adam \cite{kingma2014adam} optimizer with an initial learning rate of $\eta = 10^{-2}$, a step scheduler with parameter $\gamma=0.1$ at $30\%$ and $70\%$.

\subsection{2D Surface and 3D Volume Manifolds}

\subsubsection{Datasets}

\paragraph{Overfitting setting.}
We use shapes from \cite{myles2014robust} commonly used for benchmarking geometry processing works.
We discard the mesh connectivity to work with point clouds.

\paragraph{Generalization setting.}
In order to train a foundation model for estimating spectral bases on 3D pointclouds, we unsupervisedly trained our model on many datasets covering a large spectrum of types of shapes: SURREAL~\cite{poulenard2019effective} for synthetic human bodies, SHREC'21 Protein Domains~\cite{biasotti2021shrec} for biological structures, ABC~\cite{koch2019abc} for CAD models, FAUST~\cite{bogo2014faust} for real human scans, and DeformingThings4D~\cite{li20214dcomplete} for dynamic non-rigid surfaces.
Evaluation is performed on well-established evaluation benchmarks covering many challenging settings: 
DefTransfer
\cite{sumner2004deformation}, 
MPZ14
\cite{myles2014robust},
SHREC'07 Watertight~\cite{giorgi2007watertight}, 
SHREC'14 Human Models~\cite{pickup2014shrec14}, 
SHREC'15 Non-Rigid~\cite{lian15nonrigid}, 
SHREC'16 Topological Noise 
(TopKids) \cite{lahner2016shrec}, 
SHREC'19 Humans~\cite{melzi2019shrec}, 
SHREC'20 Non-Isometric (SHREC'20-NI) \cite{dyke2020shrec}, 
SHREC'20 Geometric Reliefs (SHREC'20-GR) \cite{thompson2020shrec}, 
and Thingi10K \cite{zhou2016thingi10k}.
All shapes in the datasets, both in train and evaluation, originally come with mesh connectivity. When using our method, we remove all this oracle connectivity information to keep only unstructured point clouds.
We also evaluate, but not train, on volumetric point clouds generated from surface meshes in SHREC'20 Non-Isometric \cite{dyke2020shrec}. 
To generate this data, we perform volumetric sampling using 
{Kaolin}'s
\footnote{\url{https://github.com/NVIDIAGameWorks/kaolin}} 
\cite{kaolinLibrary} GPU-accelerated approach. Random candidate points are uniformly sampled within the mesh's axis-aligned bounding box (with 5\% padding), then for each point we compute the unsigned distance to the nearest mesh surface and determine inside/outside classification via ray casting\footnote{Shooting a ray and counting mesh intersections, with odd counts indicating interior points.}. Points classified as interior are retained and subsampled to the target count, optionally combined with surface vertices to form the final volumetric point cloud.

\subsubsection{Implementation -- Generalization Setting}

The following implementation details describe our generalization model that learns to predict a spectral basis resembling Laplacian eigenfunctions across diverse 3D shapes, rather than overfitting to a single geometry. The model is trained on a large-scale dataset combining multiple datasets to generalize to unseen shapes at test time.

\paragraph{Network Architecture.} Our 3D model processes point clouds through a three-stage pipeline. First, raw 3D coordinates are passed through a preprocessing MLP with architecture [3 $\rightarrow$ 64 $\rightarrow$ 256 $\rightarrow$ 1024] using GELU activations \cite{hendrycks2016gelu} and layer normalization \cite{ba2016layer}. The resulting 1024-dimensional features are then processed by an 8-layer Transformer encoder with 32 attention heads, where each point attends to all other points in the point cloud via global self-attention. The Transformer uses $d_{\text{model}} = 1024$, feedforward dimension of 1024, GELU activations, and processes sequences in batch-first format. Finally, a lightweight output MLP [1024 $\rightarrow$ 256 $\rightarrow$ 50] with GELU activations and layer normalization predicts the 50 feature vectors for each point (prior to QR decomposition to get the estimated normalized spectral basis).

\paragraph{Training Configuration.} We optimize with AdamW \cite{loshchilov2018decoupled} (learning rate $5\times 10^{-4}$) and cosine annealing schedule decaying to $1 \times 10^{-5}$ over 100 epochs. We apply gradient clipping at 0.01 and accumulate gradients over 2 batches. The training set consists of approximately 200000 3D shapes (file sizes 0.2-7MB) from diverse datasets
with a batch size of 30.

\paragraph{Data Processing.} Each mesh is sampled to 1500 points using Farthest Point Sampling (FPS) \cite{gonzalez1985clustering,eldar1997farthest}, with random initialization during training and fixed initialization during validation to ensure reproducibility. We generate 500 scalar fields per shape by sampling uniform random values in the range [-1, 1] at each vertex. Before computing the reconstruction loss, scalar fields are smoothed for 20 iterations using Gaussian kernel convolution with $\sigma=0.1$ over a $k$-nearest neighbor graph with $k=15$. The kNN graph construction uses Euclidean distance (not cosine similarity) and includes self-loops. Oracle cotangent Laplacian eigendecompositions are precomputed and cached to accelerate the validation phase, where we monitor cosine similarity between predicted and oracle eigenvectors.

\paragraph{Computational Resources.} All experiments were conducted on 8 GPUs (type NVIDIA A100-SXM4-40GB) with automatic mixed precision \cite{micikevicius2018mixed} training enabled through PyTorch Lightning \cite{paszke2019pytorch,falcon2019pytorch}.

\subsubsection{Implementation -- Overfitting setting}

The following implementation details describe the configuration for overfitting the eigendecomposition of a specific 3D shape. This setup employs a smaller architecture to achieve maximum accuracy for a single target geometry.

\paragraph{Network Architecture.} The architecture for single-shape overfitting uses a more compact design compared to the generalization model. Raw 3D coordinates are processed through a preprocessing MLP with architecture [3 $\rightarrow$ 32 $\rightarrow$ 128 $\rightarrow$ 256] using GELU activations and layer normalization. The core network consists of an 8-layer Transformer encoder with 8 attention heads (compared to 32 for generalization), $d_{\text{model}} = 256$, and feedforward dimension of 256. The output head is a simple two-layer MLP [256 $\rightarrow$ 50] with GELU activation and layer normalization that directly predicts the 50 Laplacian eigenvectors.

\paragraph{Training Configuration.} We optimize with AdamW (learning rate 0.001) and cosine annealing schedule decaying to $1 \times 10^{-5}$ over 5000 epochs, training for a total of 12,000 epochs. Gradient clipping is applied at 0.01. The training consists of a single mesh with batch size 1, allowing the model to specialize completely to the target geometry.

\paragraph{Data Processing.} The single training mesh is sampled to 4000 points using Farthest Point Sampling (FPS) with random initialization at each iteration to provide variation during training. We use a higher point count compared to the generalization setting (4000 vs 1500) since batch size 1 requires significantly less GPU memory, allowing for denser point sampling. We generate 1500 scalar fields per shape by sampling uniform random values in the range [-1, 1] at each vertex. Before computing the reconstruction loss, each scalar field is smoothed for 40 iterations using Gaussian kernel convolution over a $k$-nearest neighbor graph with $k=10$, where $\sigma$ is randomly sampled from the range [0.01, 0.2] independently for each scalar field. We employ this range of smoothing scales rather than a fixed value because we empirically observed improved results across various shapes. We speculate that sampling different $\sigma$ values allows the network to sense the shape's manifold at multiple scales, which helps generate a statistically sufficient number of smooth functions that are smooth with respect to the surface metric -- a requirement for optimal approximation. Empirically, we found that using a range of $\sigma$ values does not contribute significantly to the generalization case, suggesting this multi-scale sensing is particularly beneficial when overfitting to a specific geometry. The kNN graph uses Euclidean distance and does not include self-loops. Validation is performed every 200 epochs, with model checkpoints saved at the same frequency.

\paragraph{Computational Resources.} Experiments were conducted on a single NVIDIA GPU using PyTorch Lightning. The compact architecture and small batch size make this configuration accessible for consumer-grade hardware such as an RTX 3090, requiring significantly fewer computational resources than the generalization model.

\paragraph{Eigenvalues at Inference Time.} Our model is designed to compute a spectral basis. However, we can recover without additional cost as a byproduct the associated eigenvalues of the implicit optimal-reconstruction operator. It is given by the invert of the worse approximation error. Depending on how we generate probe functions at inference time, we get different errors and thus different eigenvalues. Our methodology to generate probe functions depends on three hyperparameters: the number of nearest neighbors $k$ in the kNN graph, the number of smoothing iterations, and the smoothing scale $\sigma$. By default, we propose to take at inference time the fixed hyperparameters $k = 70$, $48$ iterations, and $\sigma = 0.101$. These numbers gave us the minimal mean relative eigenvalue discrepancy with the oracle cotangent Laplacian's eigenvalues across all tested shapes. However, we also manually tuned these hyperparameters per shape and provide in \cref{tab: overfitting tuned hyperparameters per shape} the best ones we found. Unless specified otherwise, presented results used the tuned version rather than the fixed one.
In future work we intend to provide an automatic mechanism for finding the best hyperparameters per shape. For instance, as probe functions depend differentiably on $\sigma$, it could be learned via descent strategies.

\begin{table}[h]
\centering
\caption{Optimal hyperparameters for generating probe functions in the single-shape overfitting setting when estimating eigenvalues. For each shape, we report the k-nearest neighbors ($k$), number of smoothing iterations, and Gaussian kernel bandwidth ($\sigma$) that yielded the best alignment with the oracle cotangent Laplacian's eigenvalues.}
\label{tab: overfitting tuned hyperparameters per shape}
\begin{tabular}{lccc}
\toprule
\textbf{Shape} & {$\boldsymbol{k}$} & \textbf{Iterations} & {$\boldsymbol{\sigma}$} \\
\midrule
Armadillo & 45 & 111 & 0.194 \\
Beetle  & 70 & 120 & 0.120 \\
Bimba & 33 & 57 & 0.101 \\
Botijo & 27 & 93 & 0.120 \\
David   & 33 & 48  & 0.138 \\
Dente   & 21 & 75  & 0.101 \\
Dragon  & 70 & 48 & 0.269 \\
Elephant & 57 & 75 & 0.157 \\
Eros    & 27 & 102 & 0.138 \\
Fertility & 63 & 40 & 0.176 \\
Heptoroid & 39 & 40  & 0.213 \\
Horse   & 70 & 75  & 0.194 \\
Kitten & 27 & 66 & 0.120 \\
Knot    & 15 & 66  & 0.101 \\
Laurent Hand & 21 & 102 & 0.101 \\
Lion & 39 & 111 & 0.176 \\
Master Cylinder & 15 & 102 & 0.269 \\
Pegaso & 70 & 120 & 0.213 \\
Teddy & 45 & 48 & 0.138 \\
Woodenfish & 70 & 120 & 0.232 \\
Wrench  & 70 & 120 & 0.082 \\
\bottomrule
\end{tabular}
\end{table}



\subsubsection{Additional Results}

Due to page-length constraints, we here provide additional results to those in the main paper.

\begin{figure*}[t]
    \centering
    \includegraphics[width=0.95\textwidth]{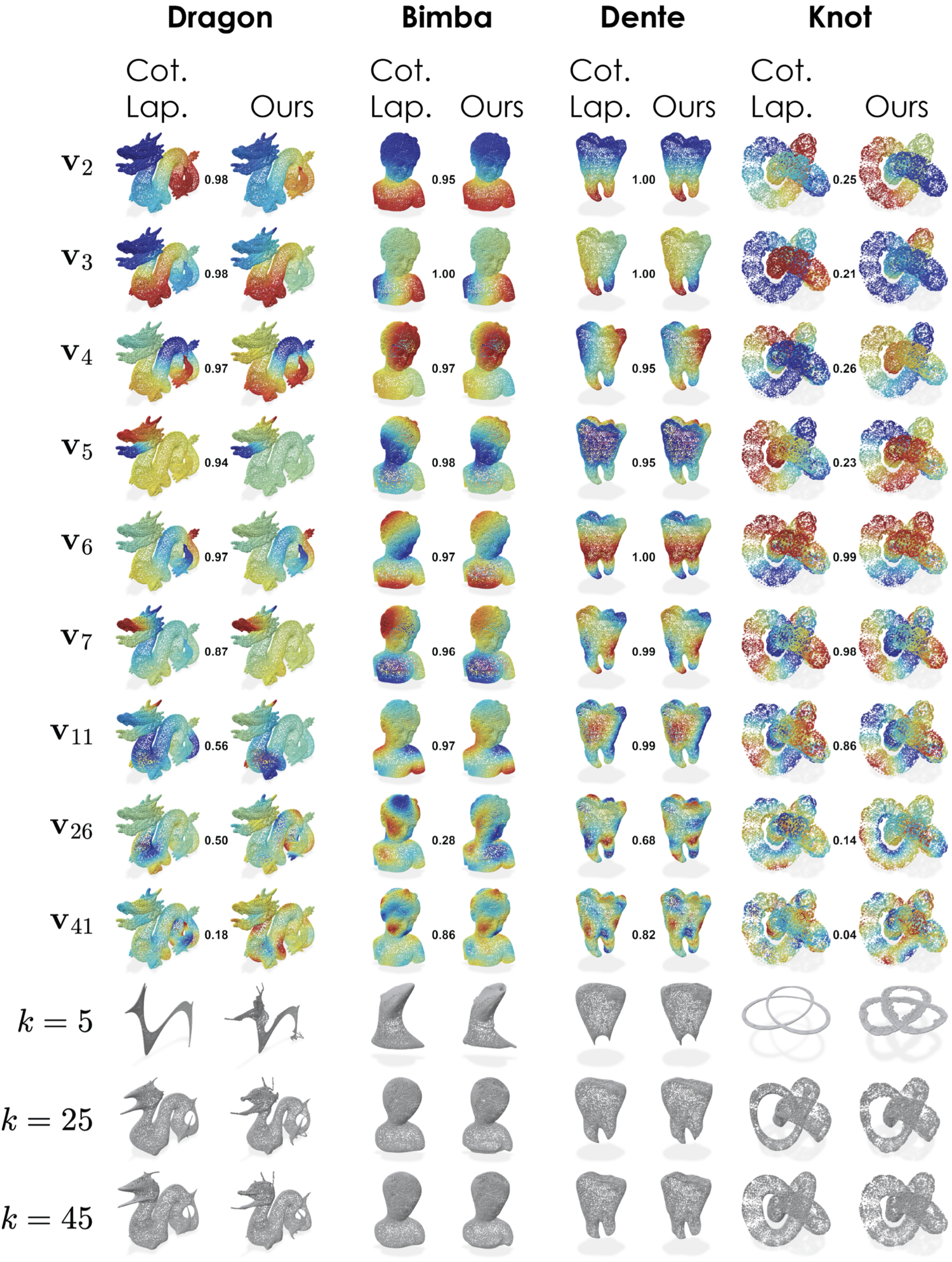}
    \caption{
    Unnormalized spectral basis (top) and $xyz$ reconstruction from $k$ basis vectors (bottom), using either the oracle cotangent Laplacian or our method (overfitting setting). 
    Scalars are cosine similarities between basis vectors.
    }
    \label{fig: supp_overfit_eigvec_and_filtering_1}
\end{figure*}

\begin{figure*}[t]
    \centering
    \includegraphics[width=0.95\textwidth]{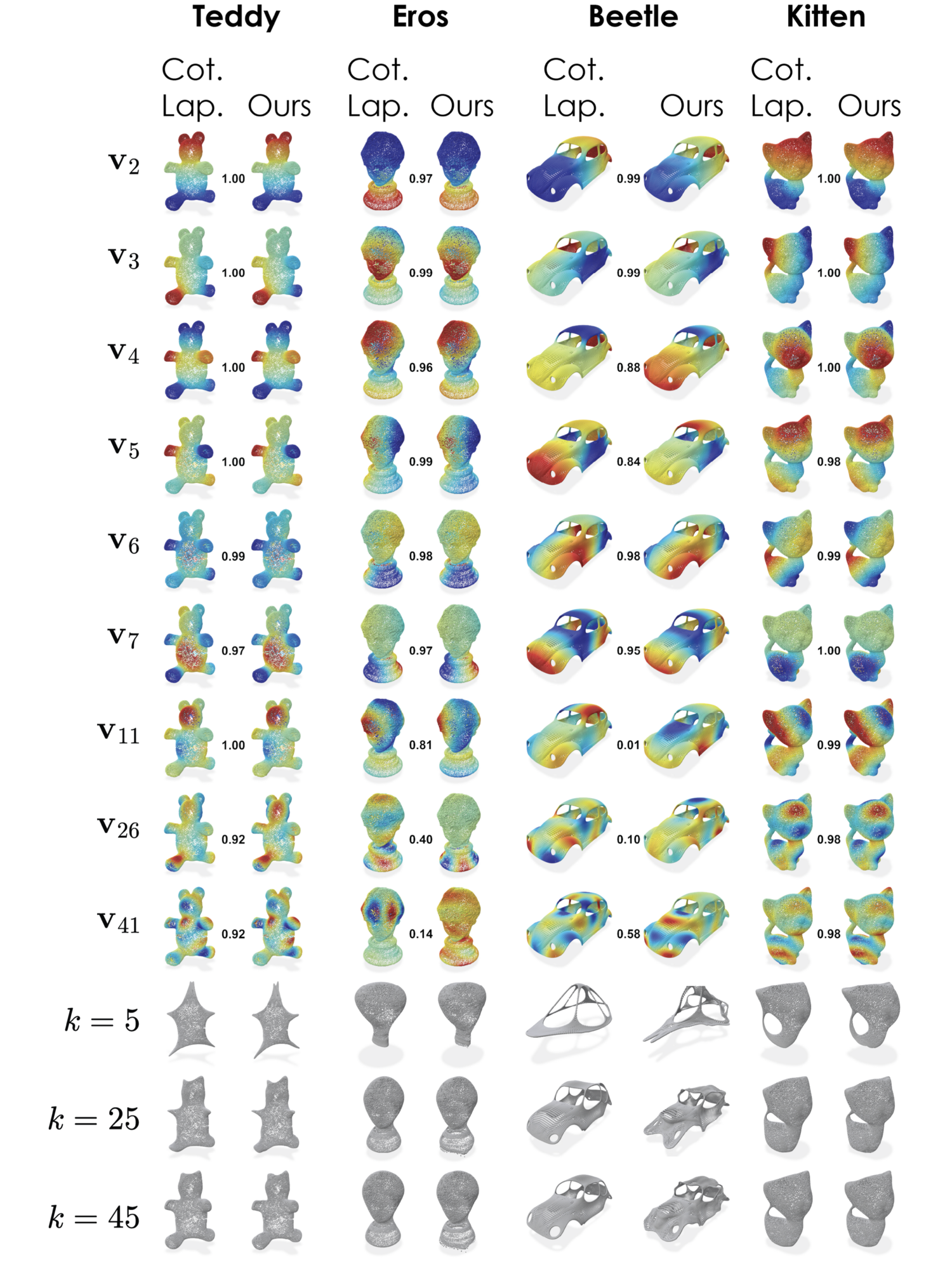}
    \caption{
    Unnormalized spectral basis (top) and $xyz$ reconstruction from $k$ basis vectors (bottom), using either the oracle cotangent Laplacian or our method (overfitting setting). 
    Scalars are cosine similarities between basis vectors.
    }
    \label{fig: supp_overfit_eigvec_and_filtering_2}
\end{figure*}

\begin{figure*}[t]
    \centering
    \includegraphics[width=0.9\textwidth]{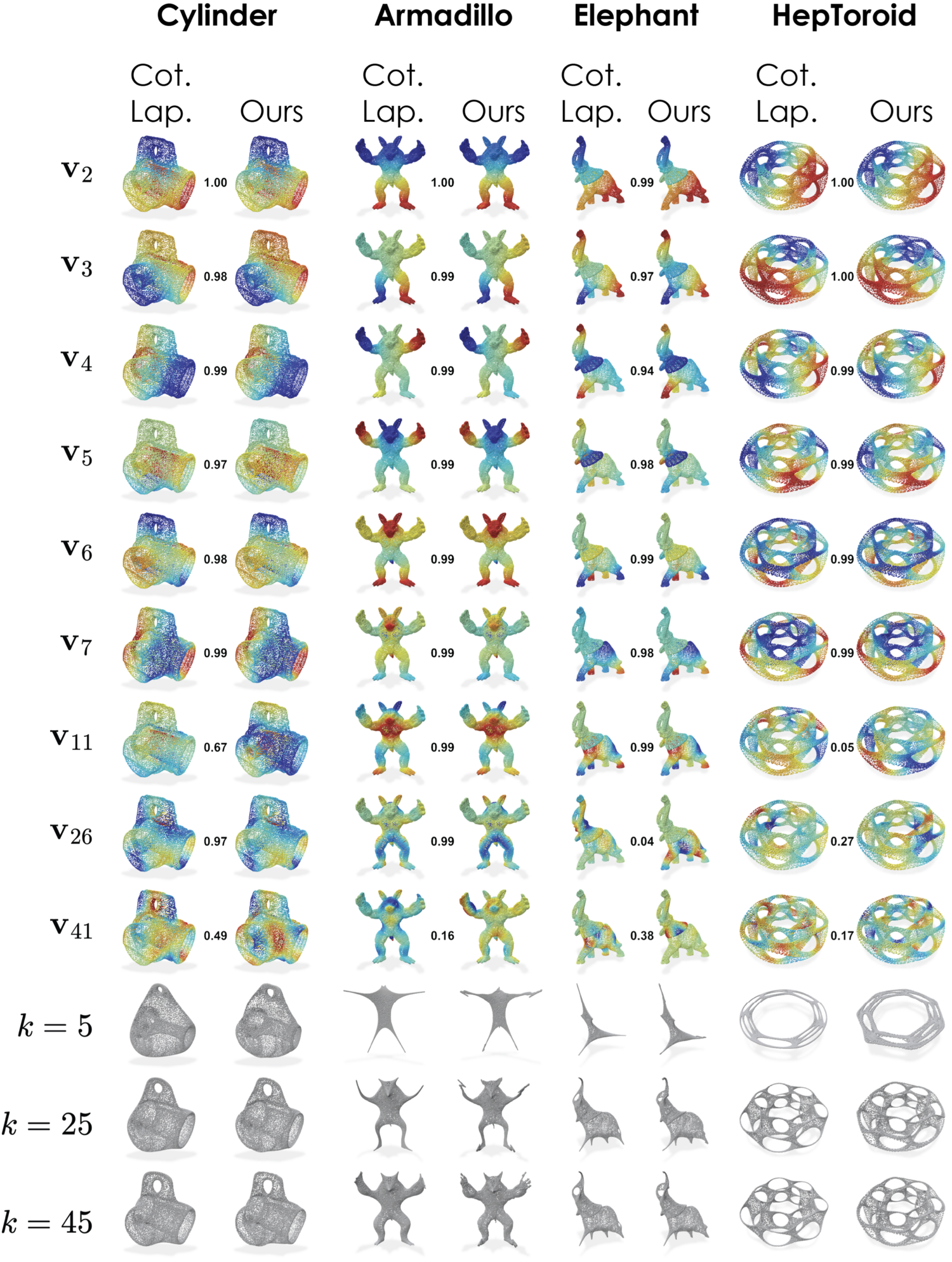}
    \caption{
    }
    Unnormalized spectral basis (top) and $xyz$ reconstruction from $k$ basis vectors (bottom), using either the oracle cotangent Laplacian or our method (overfitting setting). 
    Scalars are cosine similarities between basis vectors.
    \label{fig: supp_overfit_eigvec_and_filtering_3}
\end{figure*}

\begin{figure*}[t]
    \centering
    \includegraphics[width=0.95\textwidth]{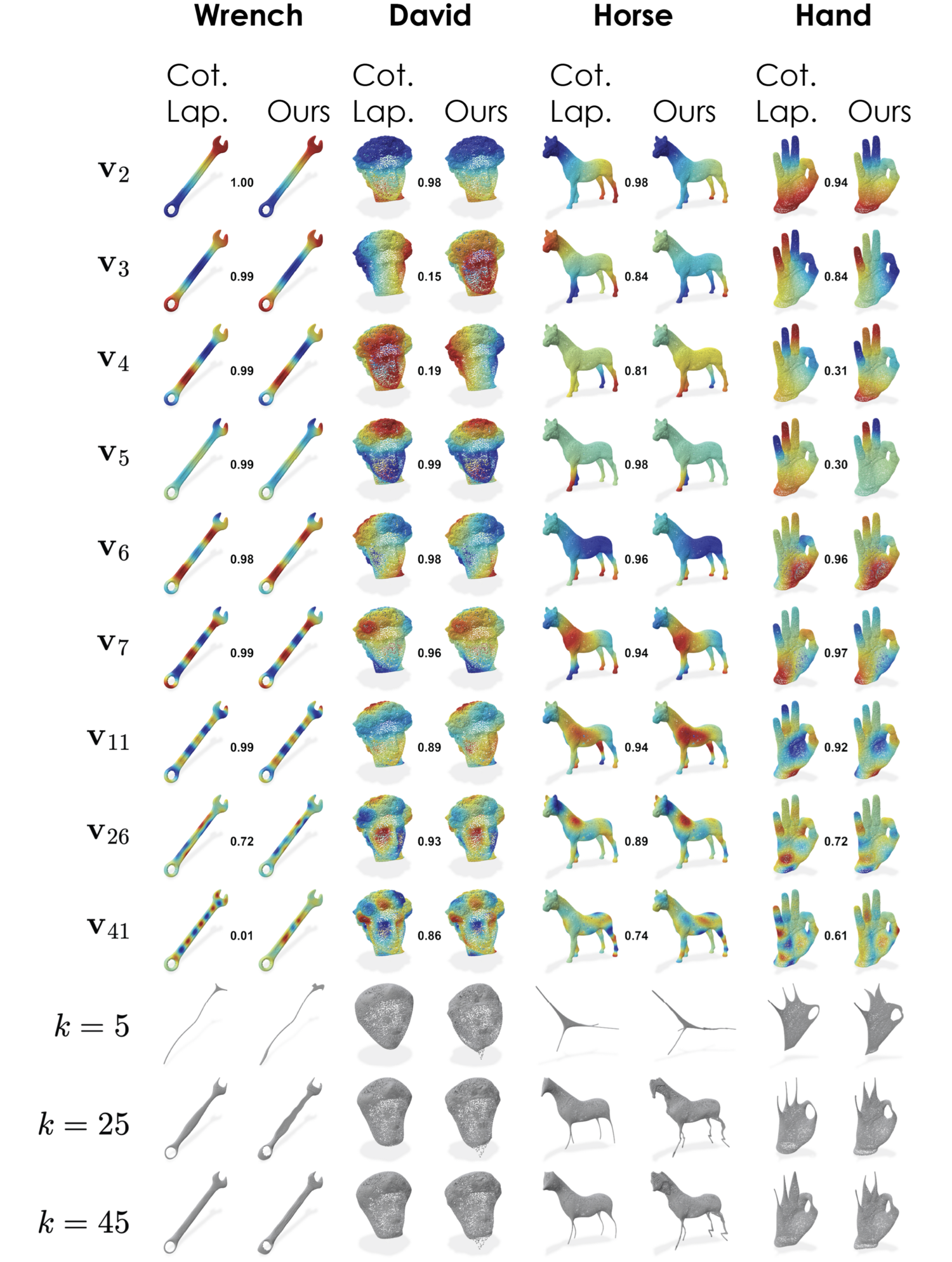}
    \caption{
    Unnormalized spectral basis (top) and $xyz$ reconstruction from $k$ basis vectors (bottom), using either the oracle cotangent Laplacian or our method (overfitting setting). 
    Scalars are cosine similarities between basis vectors.
    }
    \label{fig: supp_overfit_eigvec_and_filtering_4}
\end{figure*}

\begin{figure*}[t]
    \centering
    \includegraphics[width=0.72\textwidth]{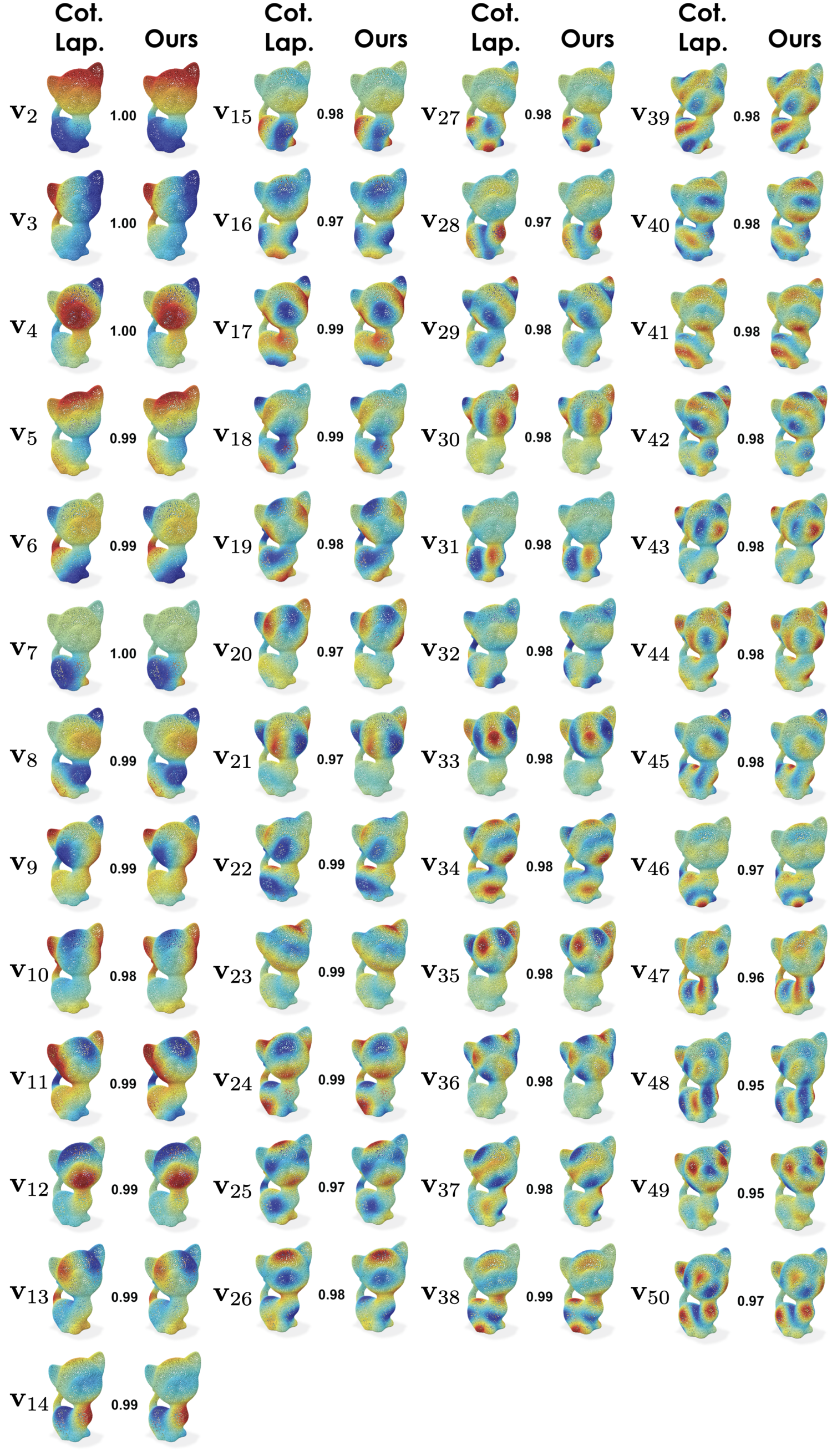}
    \caption{
    Unnormalized spectral basis using either the oracle cotangent Laplacian or our method (overfitting setting). 
    Scalars are cosine similarities between basis vectors.
    }
    \label{fig: supp_overfit_eigvec_cat}
\end{figure*}

\begin{figure*}[t]
    \centering
    \includegraphics[width=0.72\textwidth]{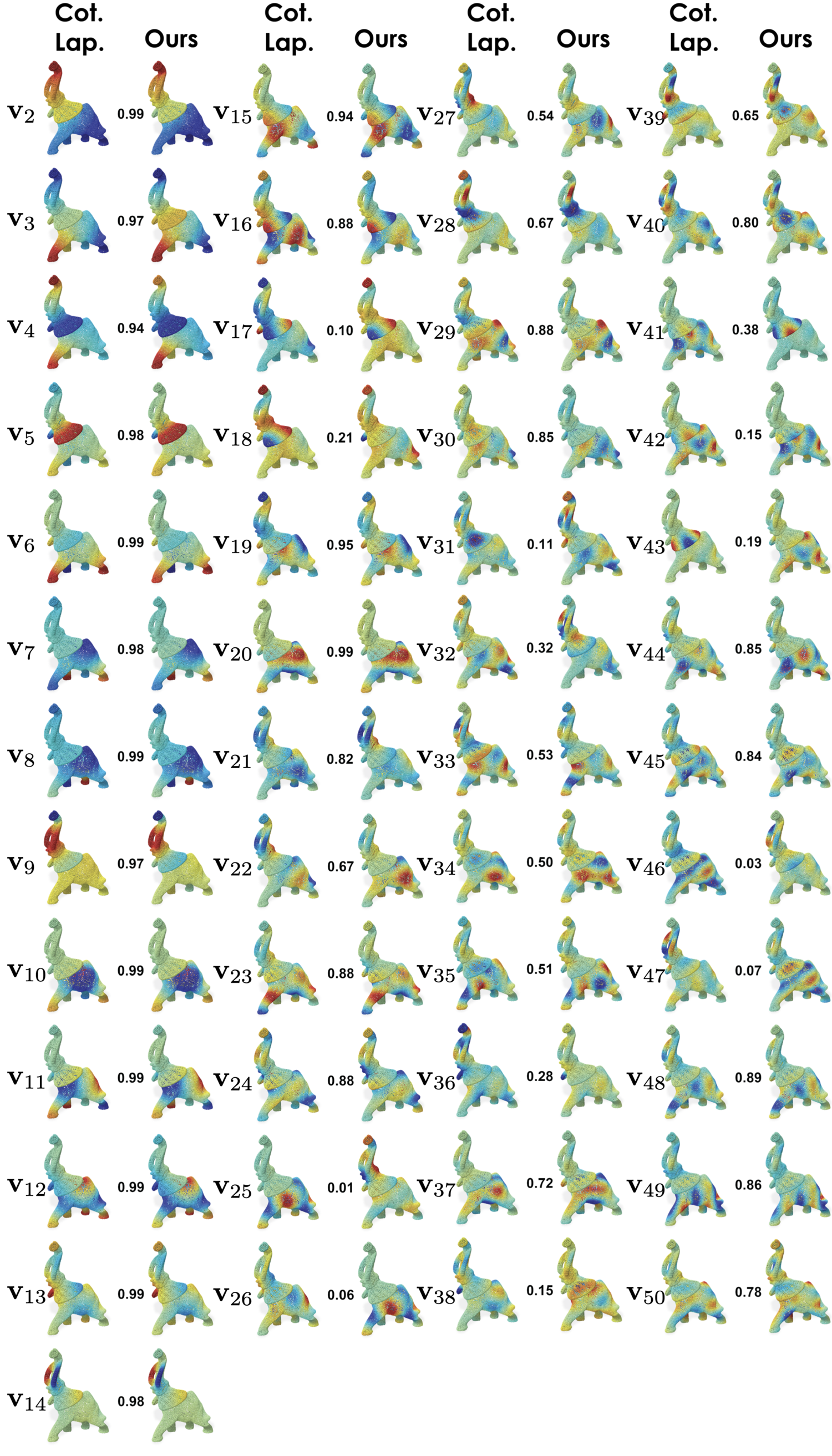}
    \caption{
    Unnormalized spectral basis using either the oracle cotangent Laplacian or our method (overfitting setting). 
    Scalars are cosine similarities between basis vectors.
    }
    \label{fig: supp_overfit_eigvec_elephant}
\end{figure*}

\begin{figure*}[t]
    \centering
    \includegraphics[width=0.95\textwidth]{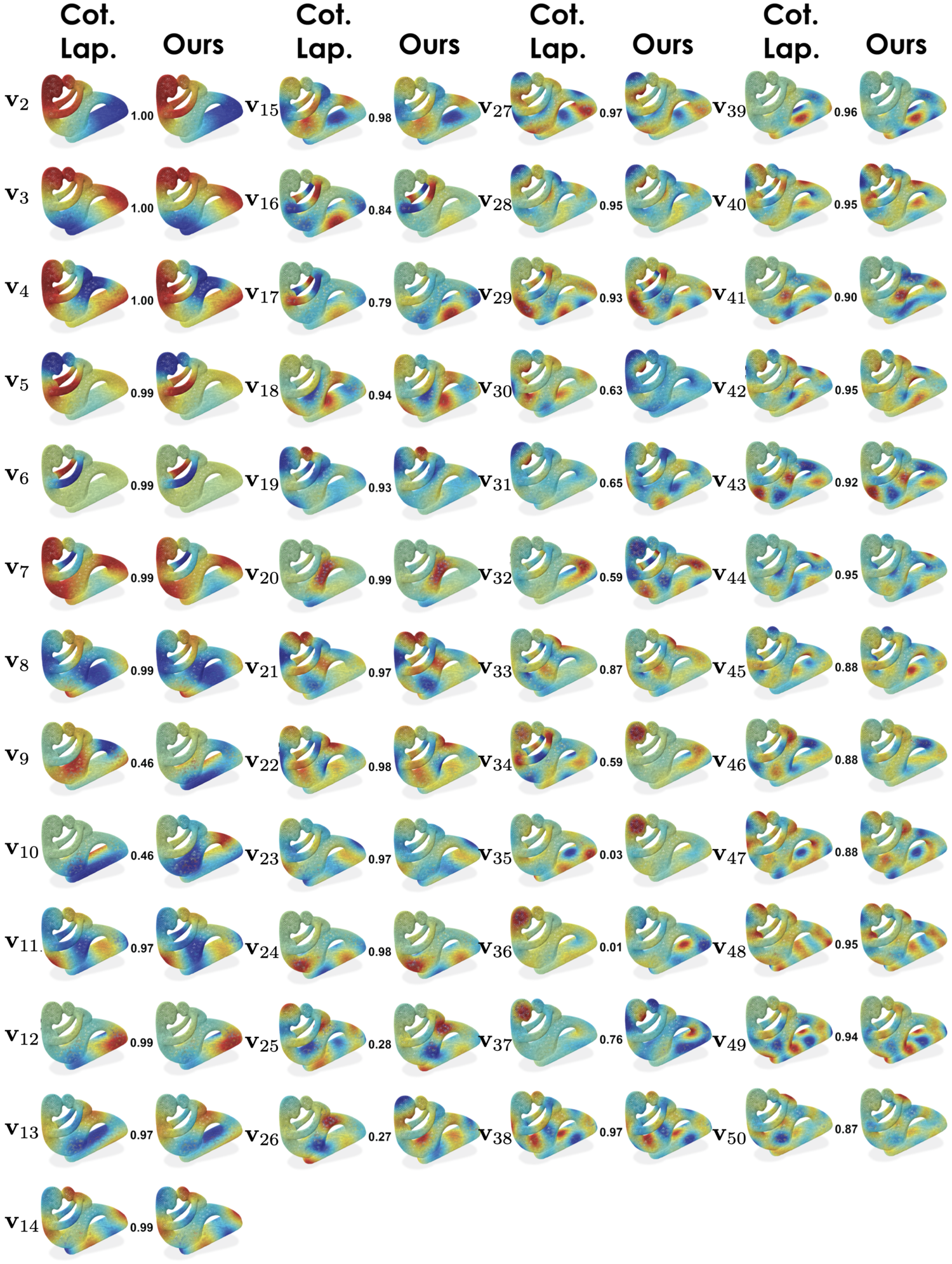}
    \caption{
    Unnormalized spectral basis using either the oracle cotangent Laplacian or our method (overfitting setting). 
    Scalars are cosine similarities between basis vectors.
    }
    \label{fig: supp_overfit_eigvec_fertility}
\end{figure*}

\begin{figure*}[t]
    \centering
    \includegraphics[width=0.75\textwidth]{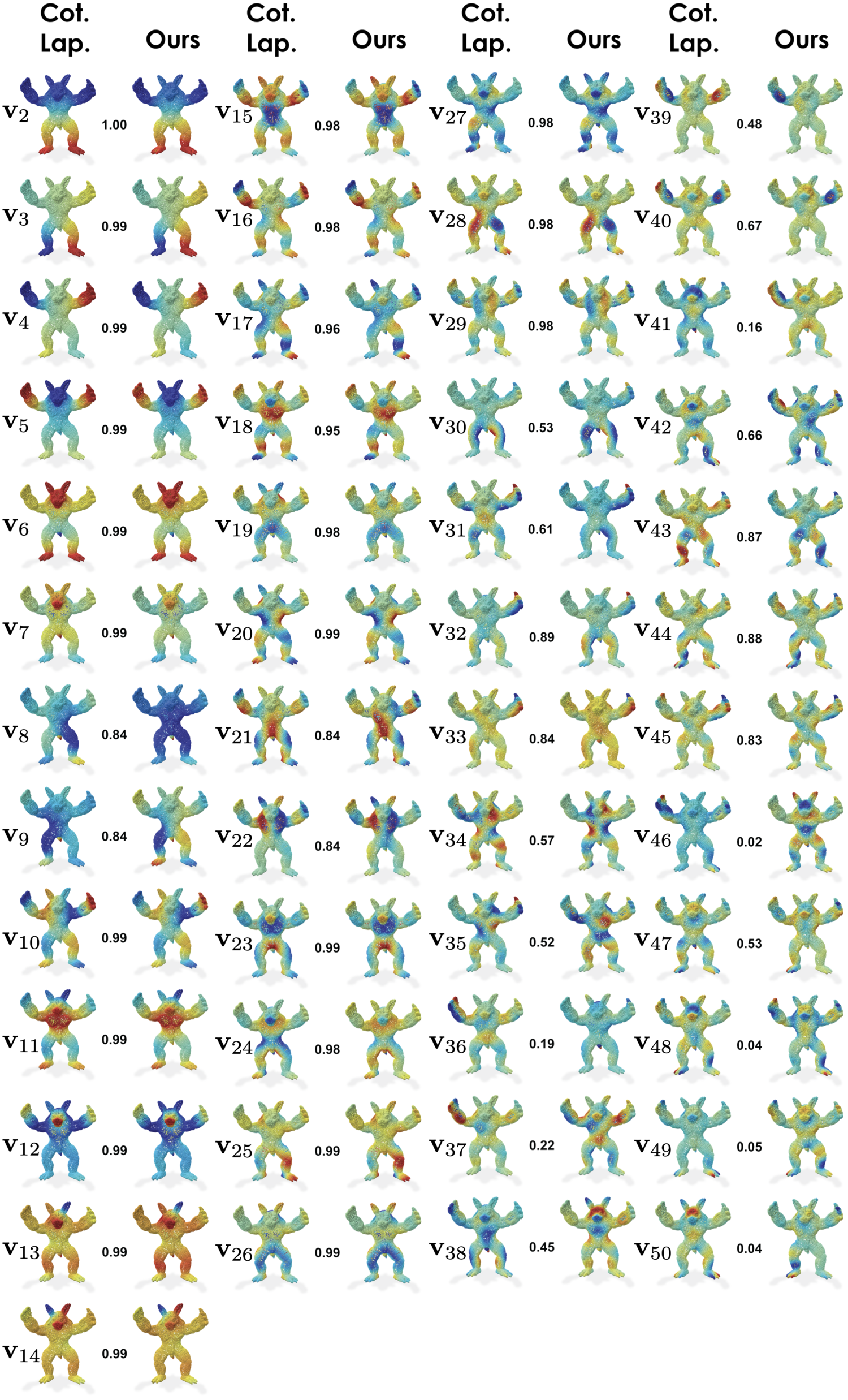}
    \caption{
    Unnormalized spectral basis using either the oracle cotangent Laplacian or our method (overfitting setting). 
    Scalars are cosine similarities between basis vectors.
    }
    \label{fig: supp_overfit_eigvec_armadillo}
\end{figure*}

\begin{figure*}[t]
    \centering
    \includegraphics[width=\textwidth]{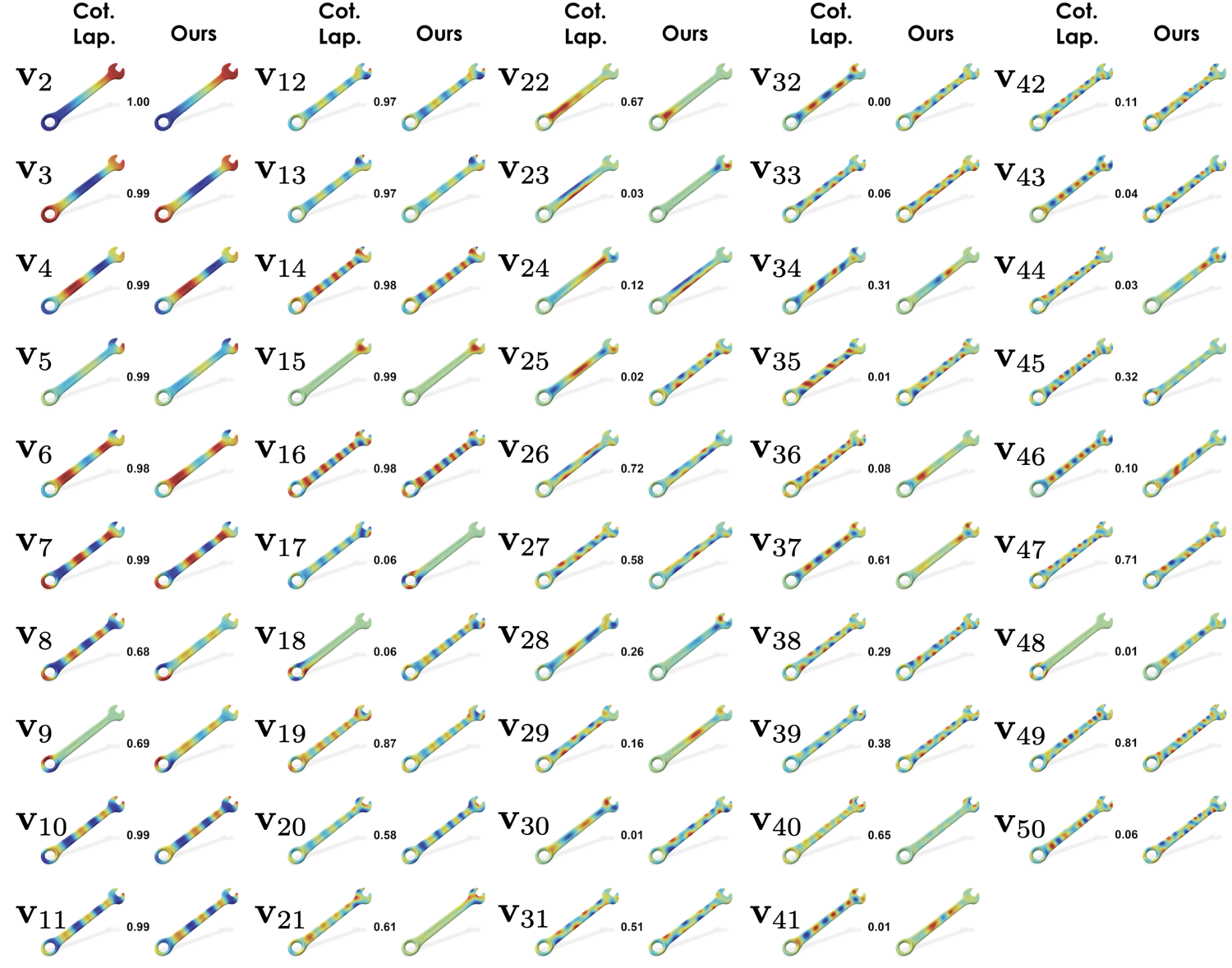}
    \caption{
    Unnormalized spectral basis using either the oracle cotangent Laplacian or our method (overfitting setting). 
    Scalars are cosine similarities between basis vectors.
    }
    \label{fig: supp_overfit_eigvec_wrench}
\end{figure*}

\begin{figure*}[t]
    \centering
    \includegraphics[width=\textwidth]{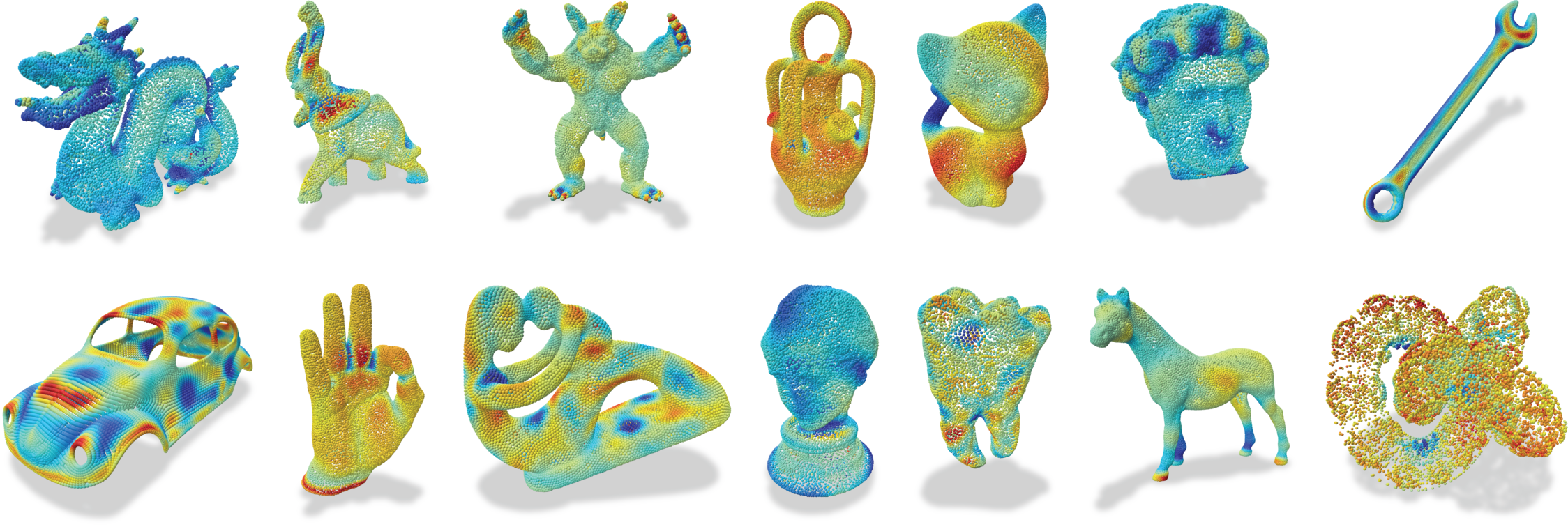}
    \caption{
    Estimated mass metric $M$ from $\mathbf{q}_1$ (overfitting setting).
    }
    \label{fig:supp-overfit-metric}
\end{figure*}

\begin{figure*}[t]
    \centering
    \includegraphics[width=\textwidth]{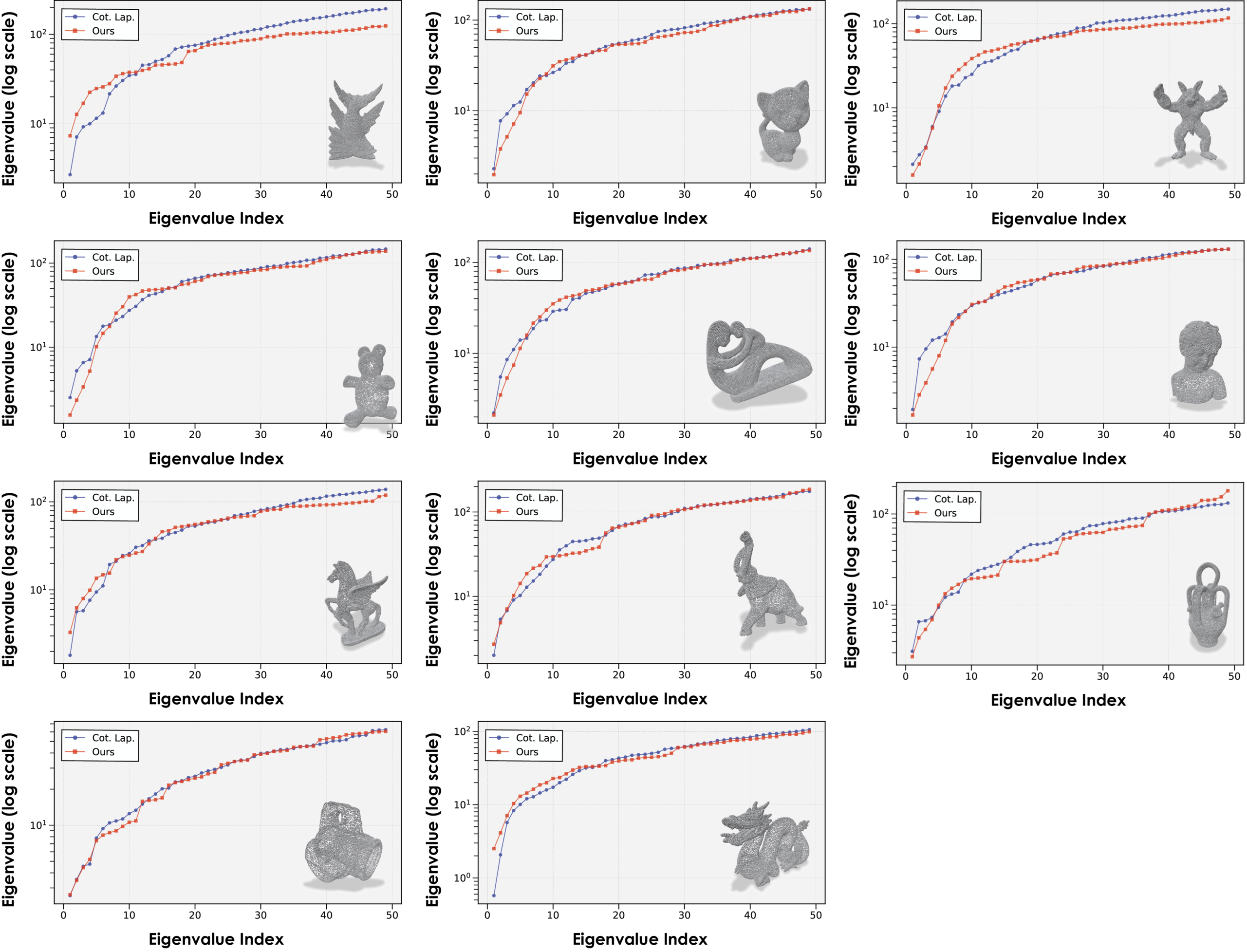}
    \caption{
    Eigenvalues of the oracle cotangent Laplacian and our estimated ones (overfitting setting, fine-tuned hyperparameter configuration for generating probe functions per shape).
    }
    \label{fig: supp_overfit_eigenvalues tuned hyperparameters}
\end{figure*}

\begin{figure*}[t]
    \centering
    \includegraphics[width=\textwidth]{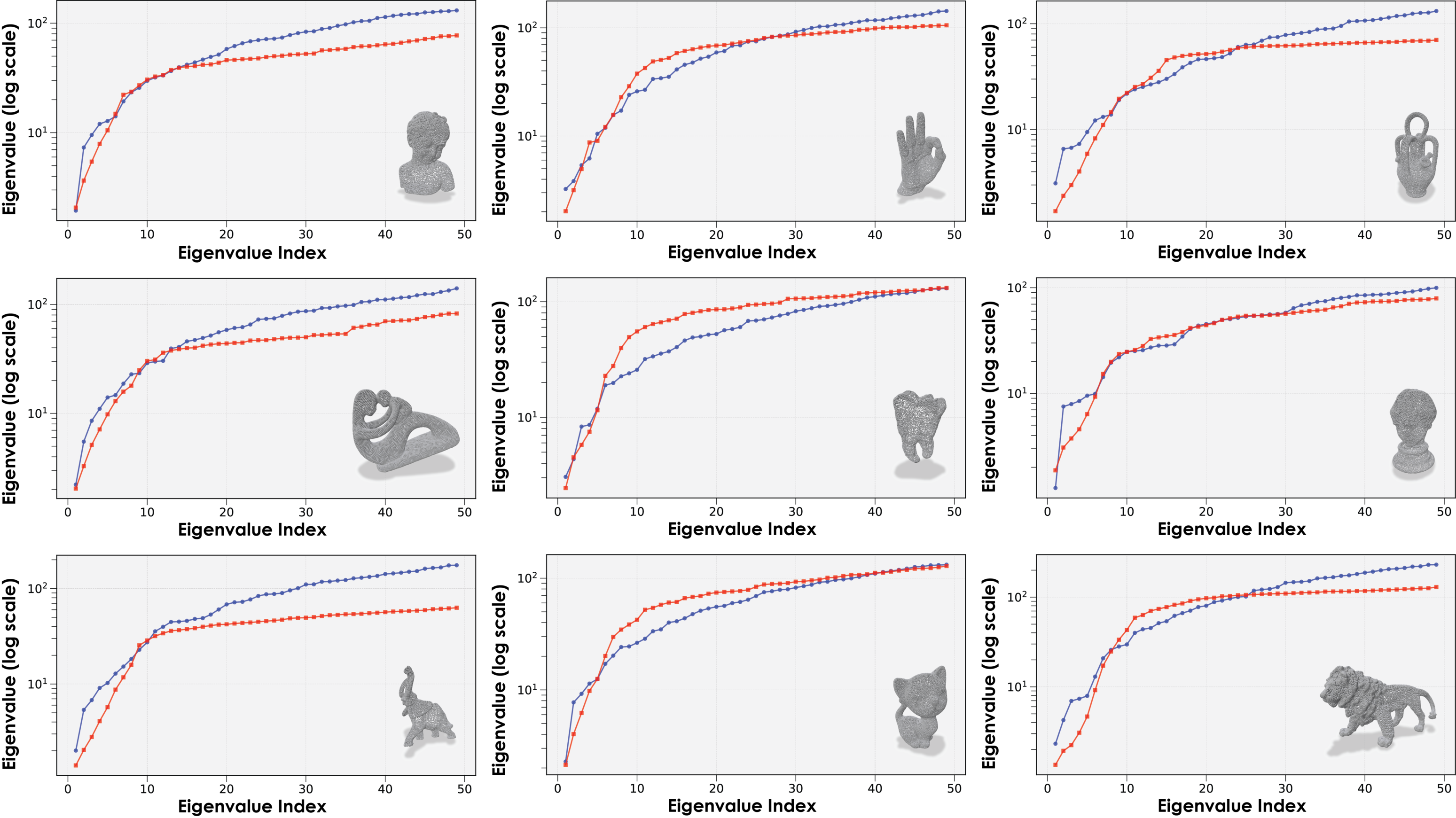}
    \caption{
    Eigenvalues of the oracle cotangent Laplacian and our estimated ones (overfitting setting, 
    fixed hyperparameter configuration for generating probe functions across all shapes).
    }
    \label{fig: supp_overfit_eigenvalues fixed hyperparameters}
\end{figure*}

\paragraph{Overfitting setting.} 
In \cref{fig: supp_overfit_eigvec_and_filtering_1,fig: supp_overfit_eigvec_and_filtering_2,fig: supp_overfit_eigvec_and_filtering_3,fig: supp_overfit_eigvec_and_filtering_4}, we provide additional visualizations of the learned unnormalized spectral basis $\mathbf{v}_i$ and obtained spectral filtering, generalizing \cref{fig: 3d point clouds overfitting spectral bases and reconstructions} to other shapes.
We also plot on a few shapes in \cref{fig: supp_overfit_eigvec_cat,fig: supp_overfit_eigvec_elephant,fig: supp_overfit_eigvec_fertility,fig: supp_overfit_eigvec_armadillo,fig: supp_overfit_eigvec_wrench} all the first 50 spectral basis vectors (excluding the first constant one $\mathbf{v}_1$), i.e.\ $\mathbf{v}_2,\ldots, \mathbf{v}_{50}$.
In \cref{fig: supp_overfit_eigenvalues tuned hyperparameters}, we plot eigenvalue curves on other shapes than in \cref{fig: 3d point clouds overfitting eigenvalue curves} in the tuned hyperparameter setting for generating inference probe functions, along with those obtained in the fixed hyperparameter setting in \cref{fig: supp_overfit_eigenvalues fixed hyperparameters}.
In \cref{fig:supp-overfit-metric}, we display more estimated metrics $M$ from $\mathbf{q}_1$ on many more shapes than in \cref{fig: 3d point clouds overfitting metric}.
In \cref{tab: sup mat 3d point clouds overfitting quantitative cosine sim and lambda normalised discrepancy}, we give further quantitative results between our estimated eigendecomposition and the oracle cotangent one\footnote{Using the ground-truth mesh information that our point cloud method does not have access to.} for the average cosine similarity and the eigenvalue discrepancy between both bases, extending \cref{tab: 3d point clouds overfitting quantitative cosine sim and lambda normalised discrepancy}.
These results supplement those in the main manuscript demonstrating how well we are able to recover the oracle Laplacian spectral decomposition, both the basis and associated eigenvalues, with our direct neural method without any knowledge on the underlying mesh structure. While extremely similar at low frequencies, differences emerge at higher frequencies, revealing that our method can account for different details from the oracle, while still preserving the same overall (i.e.\ low-pass) information of the shape manifold. Unlike the eigenfunctions, recovering the same eigenvalues as those of the oracle Laplacian is trickier and often requires carefully tuning hyperparameters to generate the appropriate probe functions at test time.

\begin{table}[t]
    \centering
    \caption{
    Average cosine similarity between predicted and oracle
    eigenfunctions at different truncation levels $k$, and mean relative eigenvalue discrepancy (overfitting setting).
    }
    \small
    \resizebox{\columnwidth}{!}{%
    \begin{tabular}{l c c c c c}
        \toprule
            \textbf{Shape} & 
            \textbf{Image}
            & $\boldsymbol{k \leq 10}$ & $\boldsymbol{k \leq 20}$ & $\boldsymbol{k \leq 50}$ & 
            $\boldsymbol{\lambda}$ 
            \textbf{Discrepancy}
            \\
        \midrule

            Beetle & \adjustbox{valign=c}{\includegraphics[width=0.02\textwidth]{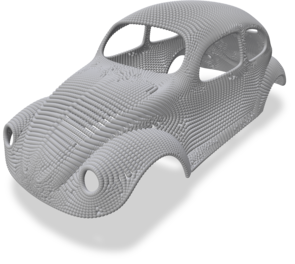}} & 0.855 & 0.657 & 0.450 & 0.514 $\pm$ 0.706 \\
            \midrule
            David & \adjustbox{valign=c}{\includegraphics[width=0.02\textwidth]{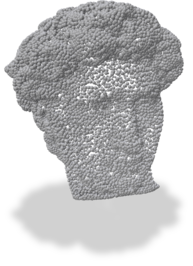}} & 0.803 & 0.879 & 0.823 & 0.098 $\pm$ 0.126 \\
            \midrule
            Dente & \adjustbox{valign=c}{\includegraphics[width=0.015\textwidth]{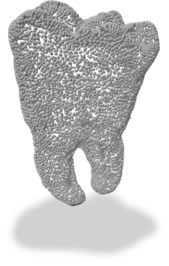}} & 0.981 & 0.976 & 0.910 & 0.106 $\pm$ 0.150 \\
            \midrule
            Dragon & \adjustbox{valign=c}{\includegraphics[width=0.02\textwidth]{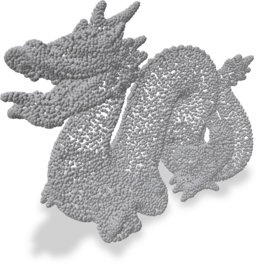}} & 0.879 & 0.827 & 0.529 & 0.197 $\pm$ 0.481 \\
            \midrule
            Eros & \adjustbox{valign=c}{\includegraphics[width=0.02\textwidth]{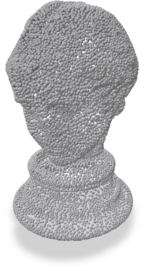}} & 0.952 & 0.891 & 0.640 & 0.083 $\pm$ 0.156 \\
            \midrule
            Heptoroid & \adjustbox{valign=c}{\includegraphics[width=0.02\textwidth]{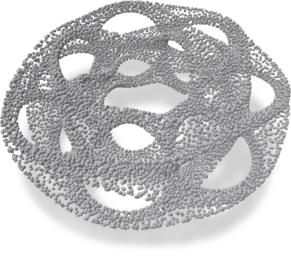}} & 0.796 & 0.757 & 0.509 & 0.182 $\pm$ 0.185 \\
            \midrule
            Horse & \adjustbox{valign=c}{\includegraphics[width=0.02\textwidth]{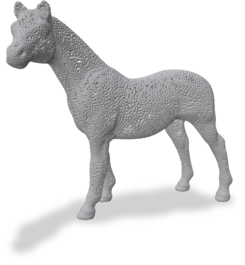}} & 0.927 & 0.908 & 0.718 & 0.107 $\pm$ 0.128 \\
            \midrule
            Knot & \adjustbox{valign=c}{\includegraphics[width=0.015\textwidth]{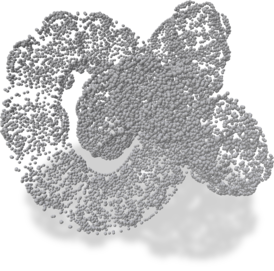}} & 0.536 & 0.687 & 0.504 & 0.127 $\pm$ 0.279 \\
            \midrule
            Wrench & \adjustbox{valign=c}{\includegraphics[width=0.015\textwidth]{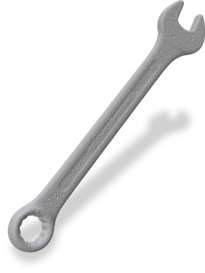}} & 0.938 & 0.817 & 0.476 & 0.694 $\pm$ 0.222 \\
            \midrule
            Master Cylinder & \adjustbox{valign=c}{\includegraphics[width=0.015\textwidth]{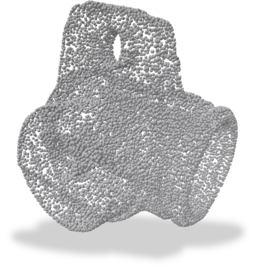}} & 0.948 & 0.923 & 0.779 & 0.051 $\pm$ 0.045 \\
            \midrule
            Teddy & \adjustbox{valign=c}{\includegraphics[width=0.015\textwidth]{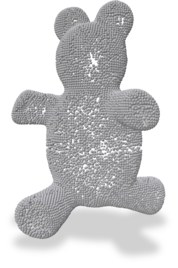}} & 0.987 & 0.981 & 0.912 & 0.109 $\pm$ 0.146 \\
        \bottomrule
    \end{tabular}
    }%
    \label{tab: sup mat 3d point clouds overfitting quantitative cosine sim and lambda normalised discrepancy}
\end{table}

\paragraph{Generelization setting.}
In \cref{tab: 3D generalisation inference results cosine sim}, we provide quantitative results comparing our predicted and oracle eigenfunctions on each evaluation dataset.
Our generalization method quantitatively demonstrates potential for designing a foundation model to estimate Laplacian spectral bases. A proper foundational model would be achieved by greatly scaling up the amount of training data, as is usually done for foundation models in other fields like image or text processing. In our limited training setting, yet still large for initial insights, our model is able to accurately estimate the first Fourier eigenfunctions, which corresponds to low-pass information and thus to the most important global information on the manifold. As expected, performance is naturally lower than in the single-shape overfitting setting.

\subsection{High-Dimensional Manifolds}

\subsubsection{Short Manifold Learning Overview}

Manifold learning is a classical task \cite{shepard1962analysis,kruskal1964multidimensional,fischl1999cortical,wandell2000visualization} in data analysis and computer vision. It consists in finding low-dimensional representations capturing the manifold structure of high-dimensional data. In practice, methods aim to find very low-dimensional embeddings in $\mathbb{R}^{k}$, with $k\ll d$ (where $d$ is the original high dimension), by preserving pairwise dissimilarities based on distances. The wide collection of approaches is usually grouped based on whether the targeted preserved structures of the manifold are local \cite{roweis2000nonlinear,donoho2003hessian,zhang2006mlle,belkin2001laplacian,belkin2003laplacian,zhang2004principal,coifman2005geometric,coifman2006diffusion,lin2023hyperbolic,lin2024tree,suzuki2019hyperbolic, gou2023discriminative,martin2005visualizing,tang2016visualizing,mcinnes2018umap,venna2010information,liu2024curvature}, global \cite{weinberger2005nonlinear,weinberger2006unsupervised,brand2005nonrigid,funke2020low}, or a combination of both \cite{pearson1901pca,hotelling1933analysis,scholkopf1998nonlinear,schwartz1989numerical,tenenbaum2000global,maaten2008visualizing,bronstein2006generalized,rosman2010nonlinear,aflalo2013spectral,pai2022deep,schwartz2019intrinsic,bracha2024wormhole,dages2025finsler,joharinad2025isumap,kim2024inductive}.

\subsubsection{Implementation}
\label{sec:image_manifold_impl_details}

For manifold learning experiments on image datasets, we work on the embedded feature manifold provided by pretrained vision models. Thus, each image is mapped to a high-dimensional\footnote{Yet lower-dimensional than the original image dimensions.} feature space, and we learn the spectral decomposition of the resulting manifold structure across four benchmark datasets: Caltech256, CIFAR100, Imagenette, and STL-10.

\paragraph{Feature Extraction.} We experiment with two pretrained vision models as feature extractors: CLIP\footnote{\url{https://huggingface.co/openai/clip-vit-base-patch32}} producing 512-dimensional embeddings, and DINOv2\footnote{\url{https://huggingface.co/facebook/dinov2-small}} producing 384-dimensional embeddings. These frozen feature extractors map images to points on a learned manifold, where geometric relationships reflect semantic similarities. The extracted embeddings are used directly as spatial coordinates for kNN graph construction with cosine similarity.

\paragraph{Network Architecture and Hyperparameters.} 
In \cref{tab:image_experiments}, we detail the architecture and probe function generation hyperparameters for each experiment. For CLIP features, we use a preprocessing MLP [512 $\rightarrow$ 256], followed by a 6-layer Transformer encoder with 8 attention heads, $d_{\text{model}} = 256$, feedforward dimension of 256, dropout of 0.1, and GELU activations. The output head is [256 $\rightarrow$ 64 $\rightarrow$ 11] with layer normalization. For DINOv2 features, we use a more compact architecture: preprocessing MLP [384 $\rightarrow$ 128], 6-layer Transformer with 4-8 attention heads depending on the dataset, $d_{\text{model}} = 128$, and a simpler output head [128 $\rightarrow$ 11].

\paragraph{Training Configuration.} We optimize with AdamW (learning rate 0.0005) and accumulate gradients over 16 batches. Gradient clipping is applied at 0.01. Training uses distributed data parallel (DDP) \cite{li2020ddp} strategy across multiple GPUs. Each dataset samples 5000 images uniformly, with deterministic sampling for validation and random sampling for training. We predict 11 eigenvectors for spectral clustering evaluation at $k \in \{2, 5, 10\}$. We generate 300 probe functions per manifold by sampling uniform random values in [-1, 1] at each point.

\paragraph{Evaluation Setting.} We compare our method against classical manifold learning approaches including PCA, UMAP, t-SNE, Isomap, and Laplacian Eigenmaps. For evaluation, we use class-weighted sampling to create challenging, non-uniform manifold distributions. Specifically, we sample 1500 images using random class weights with entropy in the range [0.01, 0.1], ensuring that the sampled manifolds have imbalanced class distributions. This sampling strategy tests the robustness of each method to realistic, non-uniform data distributions across image classes. Due to the randomness of the sampling and of some of the methods, including ours due to random probe functions, we perform 32 runs for each method on each dataset.

\paragraph{Sensitivity to Hyperparameters.} Similar to the single-shape overfitting experiments in the 3D case, we observe that results in the image manifold setting are sensitive to the choice of probe function smoothing hyperparameters ($k$, $\sigma$, iterations). The parameters in Table~\ref{tab:image_experiments} were tuned per dataset to achieve good performance. In future work, these probe function parameters could potentially be learned during training rather than manually tuned. We emphasize that the current results showcase the potential of our method, but more optimized tuning should yield better results. As such, we believe the full potential of our approach for image manifold learning has yet to be realized.

\paragraph{Computational Resources.} All experiments were conducted on 8 GPUs (type NVIDIA A100-SXM4-40GB) using PyTorch Lightning with DDP training strategy.

\begin{table}[h]
\centering
\caption{Architecture and probe function generation hyperparameters for image manifold experiments. All experiments use 6 Transformer layers, 11 predicted eigenvectors, 5000 sampled images, and 300 probe functions. The kNN graph always uses cosine similarity with self-loops enabled.}
\label{tab:image_experiments}
\resizebox{\columnwidth}{!}{
\begin{tabular}{lcccccc}
\toprule
\textbf{Dataset} & \textbf{Feature Extractor} & \textbf{$d_{\text{model}}$} & \textbf{Attn. Heads} & \textbf{$k$} & \textbf{$\sigma$} & \textbf{Iterations} \\
\midrule
Caltech256 & CLIP & 256 & 8 & 10 & 0.04 & 20 \\
Caltech256 & DINOv2 & 128 & 8 & 15 & 0.15 & 40 \\
CIFAR100 & CLIP & 256 & 8 & 5 & 0.04 & 20 \\
CIFAR100 & DINOv2 & 128 & 4 & 30 & 0.20 & 30 \\
Imagenette & CLIP & 256 & 8 & 20 & 0.08 & 20 \\
Imagenette & DINOv2 & 128 & 4 & 30 & 0.10 & 30 \\
STL-10 & CLIP & 256 & 8 & 35 & 0.08 & 20 \\
STL-10 & DINOv2 & 128 & 4 & 20 & 0.10 & 20 \\
\bottomrule
\end{tabular}
}
\end{table}

\subsubsection{Results}

We evaluate our method on four image datasets (Caltech256, CIFAR100, Imagenette, and STL-10) using both CLIP and DINOv2 embeddings, comparing against classical manifold learning methods: PCA, UMAP, t-SNE, Isomap, and Laplacian Eigenmaps. Results are reported for spectral clustering with $k \in \{2, 5, 10\}$ clusters across seven standard clustering metrics: Normalized Mutual Information (NMI) \cite{vinh2009information}, Adjusted Rand Index (ARI) \cite{hubert1985comparing}, Completeness \cite{rosenberg2007v}, Adjusted Mutual Information (AMI) \cite{vinh2009information}, Homogeneity \cite{rosenberg2007v}, V-Measure \cite{rosenberg2007v}, and Fowlkes-Mallows Index (FMI) \cite{fowlkes1983method}.

\paragraph{Quantitative Results.} 
In \cref{tab:dimred_stl10_dino,tab:dimred_stl10_clip,tab:dimred_imagenette_dino,tab:dimred_imagenette_clip,tab:dimred_caltech256_dino,tab:dimred_caltech256_clip,tab:dimred_cifar100_dino,tab:dimred_cifar100_clip} we present comprehensive results, averaged over our 32 runs, across all datasets and feature extractors. Note that UMAP and t-SNE often get better clustering scores, which is unsurprising as these methods are designed to handle classification datasets by overclustering their embeddings (better scores yet often artificial separation between clusters). Our method (Optimal Approximation Eigenmaps) demonstrates competitive performance across most settings. Our approach generally outperforms classical methods like PCA, Isomap, and most importantly, shows superior performance to its vanilla analog, Laplacian Eigenmaps, in most cases.
On CIFAR100 with CLIP embeddings, our method achieves particularly strong results at $k=10$ ($\text{NMI}=0.641$, $\text{ARI}=0.257$), matching or exceeding Laplacian Eigenmaps while substantially outperforming PCA and Isomap. Similarly, on Caltech256 with DINOv2 features at $k=5$, we achieve $\text{NMI}=0.714$ and $\text{ARI}=0.246$, demonstrating effective learned spectral representations. The method shows consistent improvements as the number of clusters increases from 2 to 10, suggesting that higher-dimensional embeddings better capture the manifold structure.

\begin{figure*}[t]
    \centering
    \includegraphics[width=0.72\textwidth]{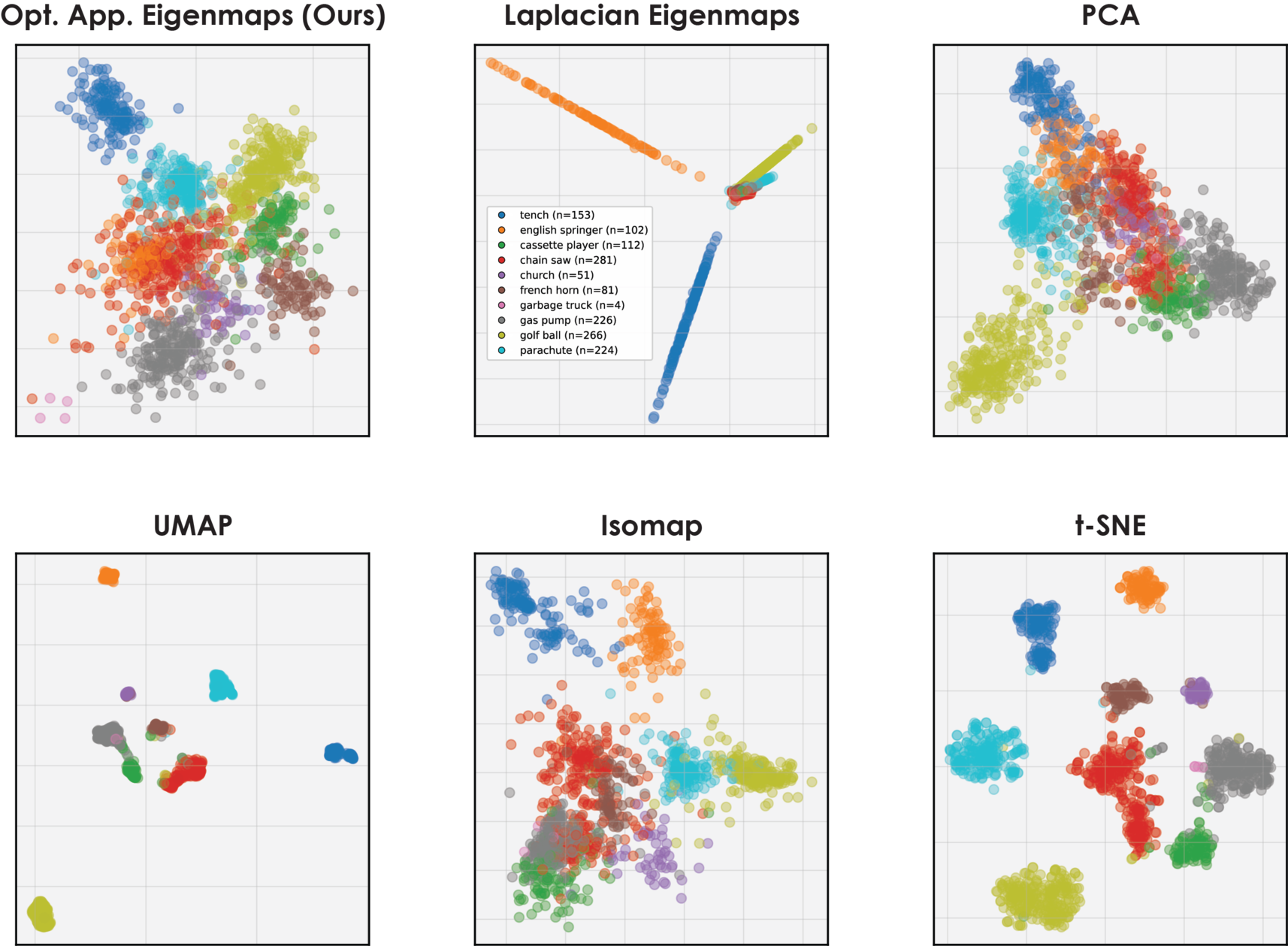}
    \caption{2D visualization of a random subset of the Imagenette dataset using CLIP embeddings. Points are colored by ground truth class labels. Our learned spectral basis (Optimal Approximation Eigenmaps) produces meaningful clusters that separate the image manifold according to semantic content, with less overlap than in PCA and Isomap, while accurately preserving the smooth manifold structure without artificially overclustering it as in t-SNE, Umap, and Laplacian Eigenmaps.}
    \label{fig: supp_imagenette_clip_cluster1}
\end{figure*}

\begin{figure*}[t]
    \centering
    \includegraphics[width=0.72\textwidth]{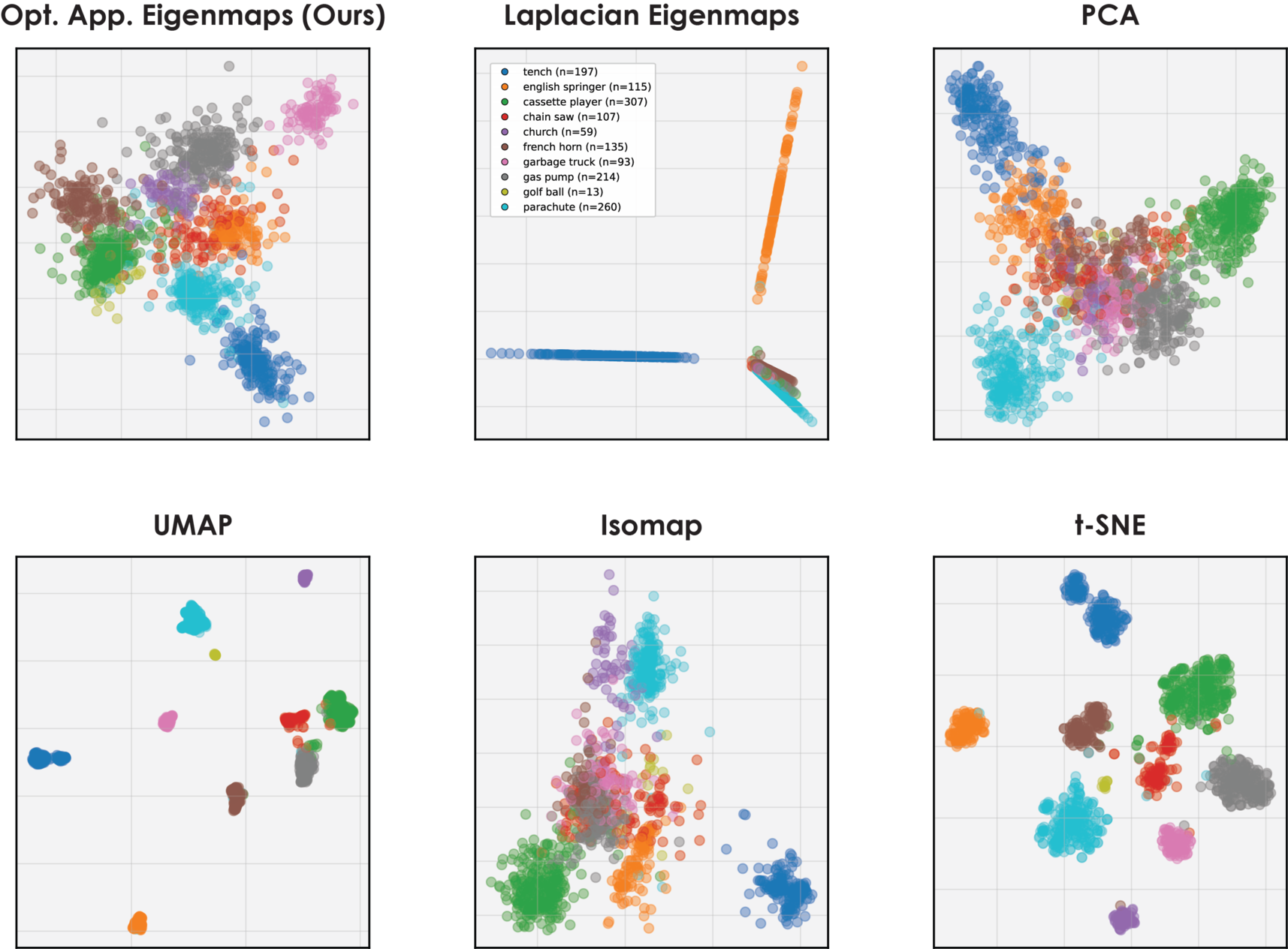}
    \caption{2D visualization of another random subset of the Imagenette dataset using CLIP embeddings.
    }
    \label{fig: supp_imagenette_clip_cluster2}
\end{figure*}

\begin{figure*}[t]
    \centering
    \includegraphics[width=0.72\textwidth]{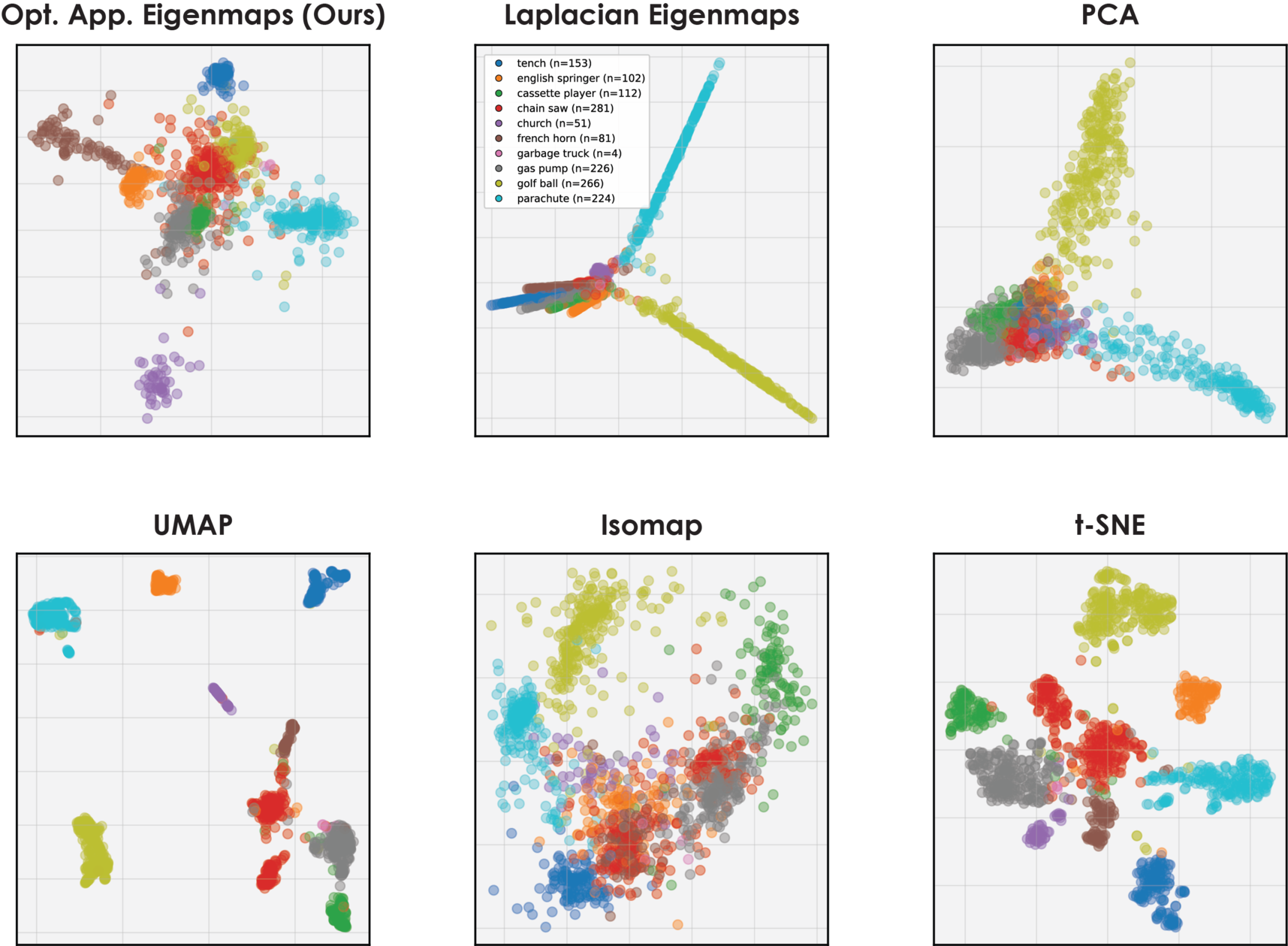}
    \caption{2D visualization of a random subset of the Imagenette dataset using DINOv2 embeddings. 
    }
    \label{fig: supp_imagenette_dino_cluster1}
\end{figure*}

\begin{figure*}[t]
    \centering
    \includegraphics[width=0.72\textwidth]{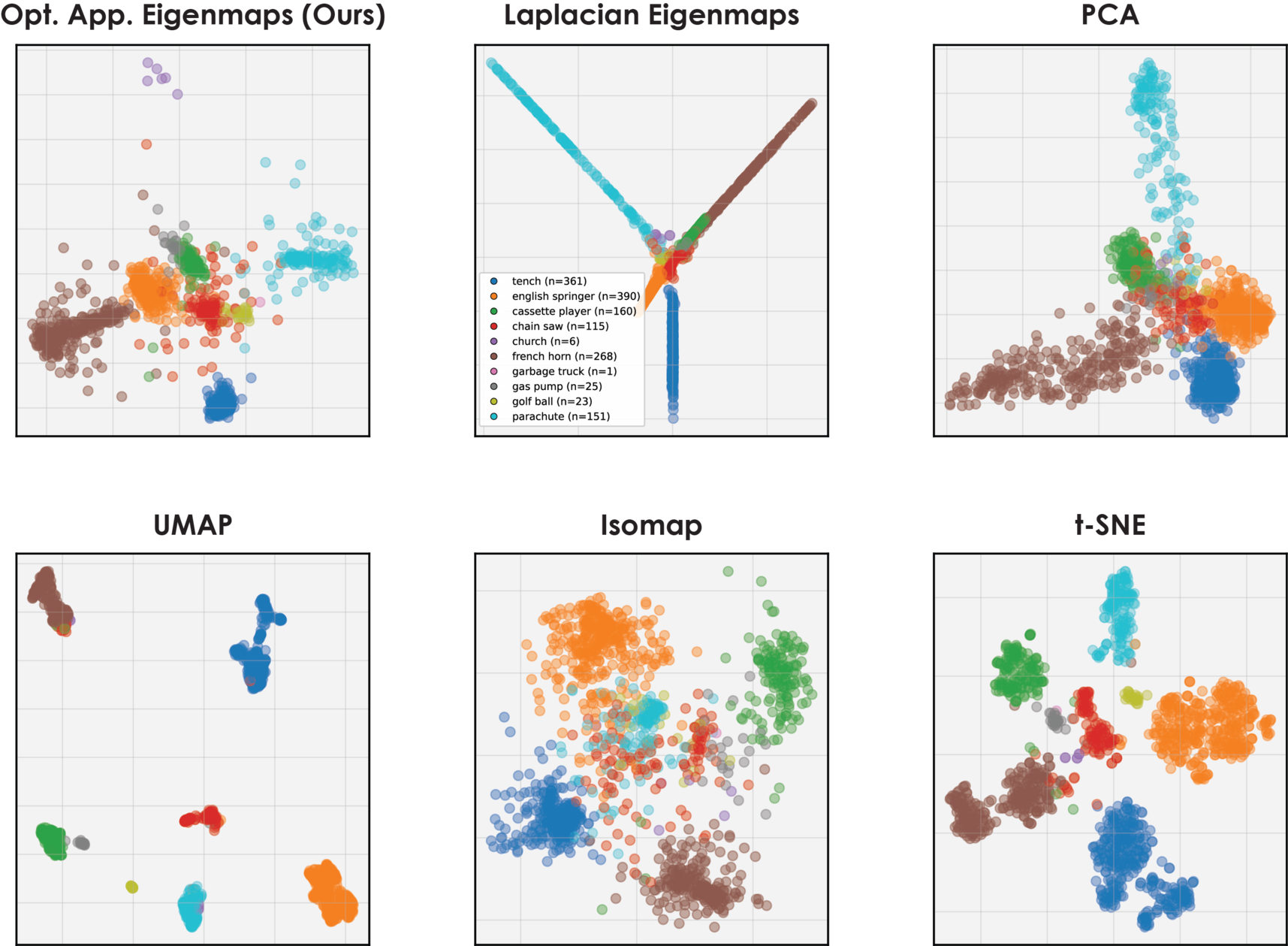}
    \caption{2D visualization of another random subset of the Imagenette dataset using DINOv2 embeddings. 
    }
    \label{fig: supp_imagenette_dino_cluster2}
\end{figure*}

\begin{figure*}[t]
    \centering
    \includegraphics[width=0.72\textwidth]{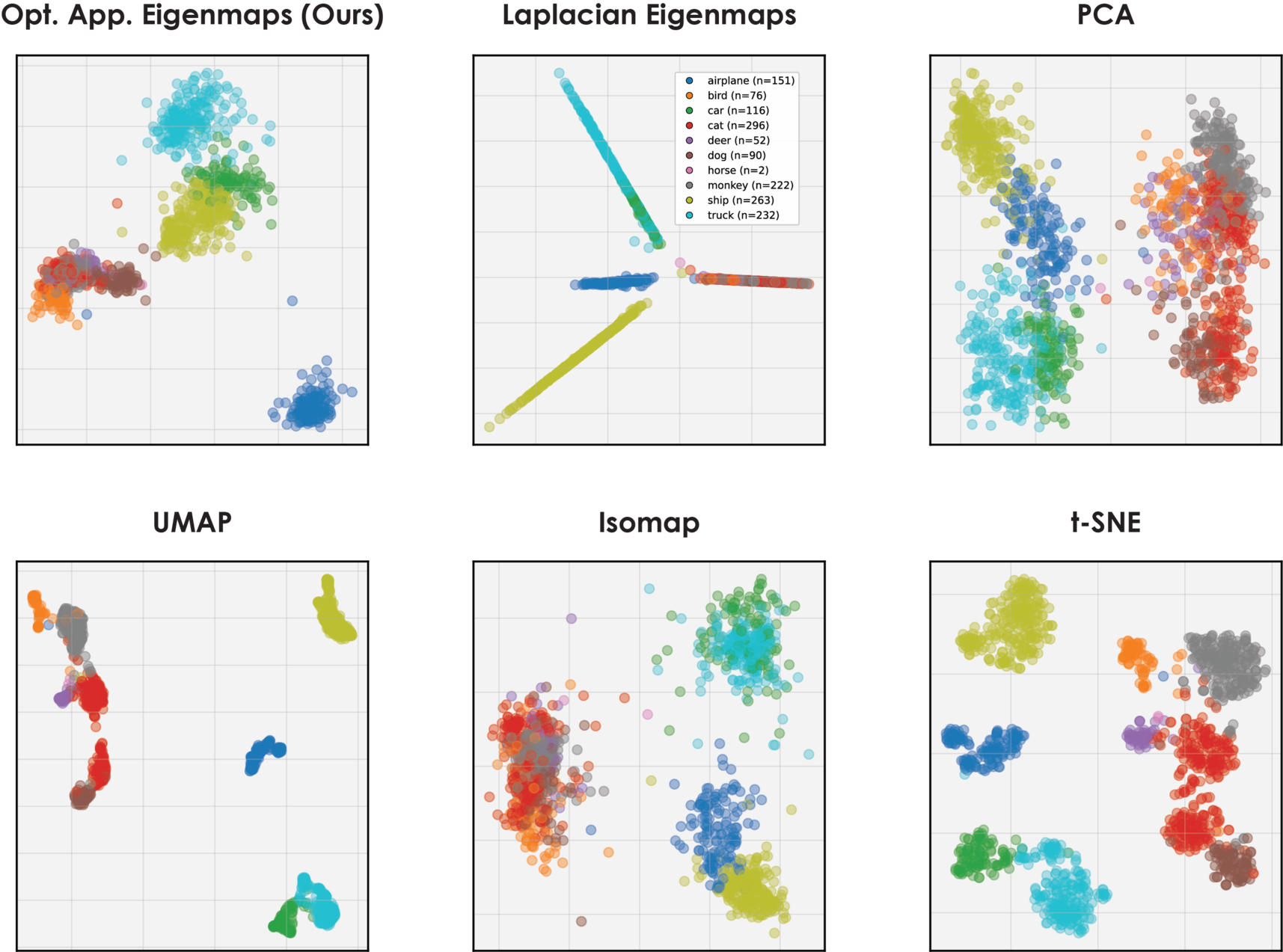}
    \caption{2D visualization of a random subset of the STL-10 dataset using CLIP embeddings. 
    }
    \label{fig: supp_stl10_clip_cluster1}
\end{figure*}

\begin{figure*}[t]
    \centering
    \includegraphics[width=0.72\textwidth]{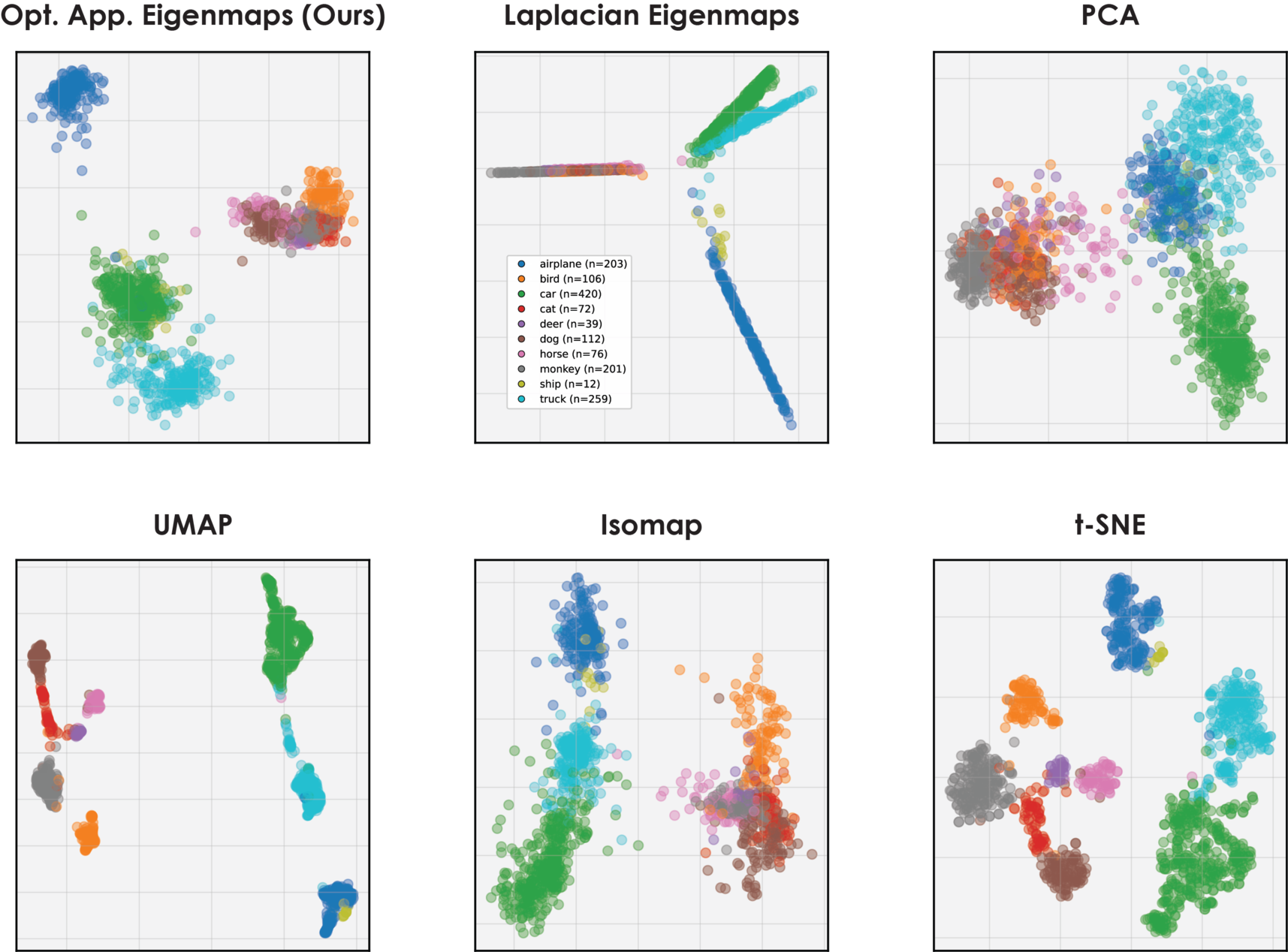}
    \caption{2D visualization of another random subset of the STL-10 dataset using CLIP embeddings. 
    }
    \label{fig: supp_stl10_clip_cluster2}
\end{figure*}

\begin{figure*}[t]
    \centering
    \includegraphics[width=0.72\textwidth]{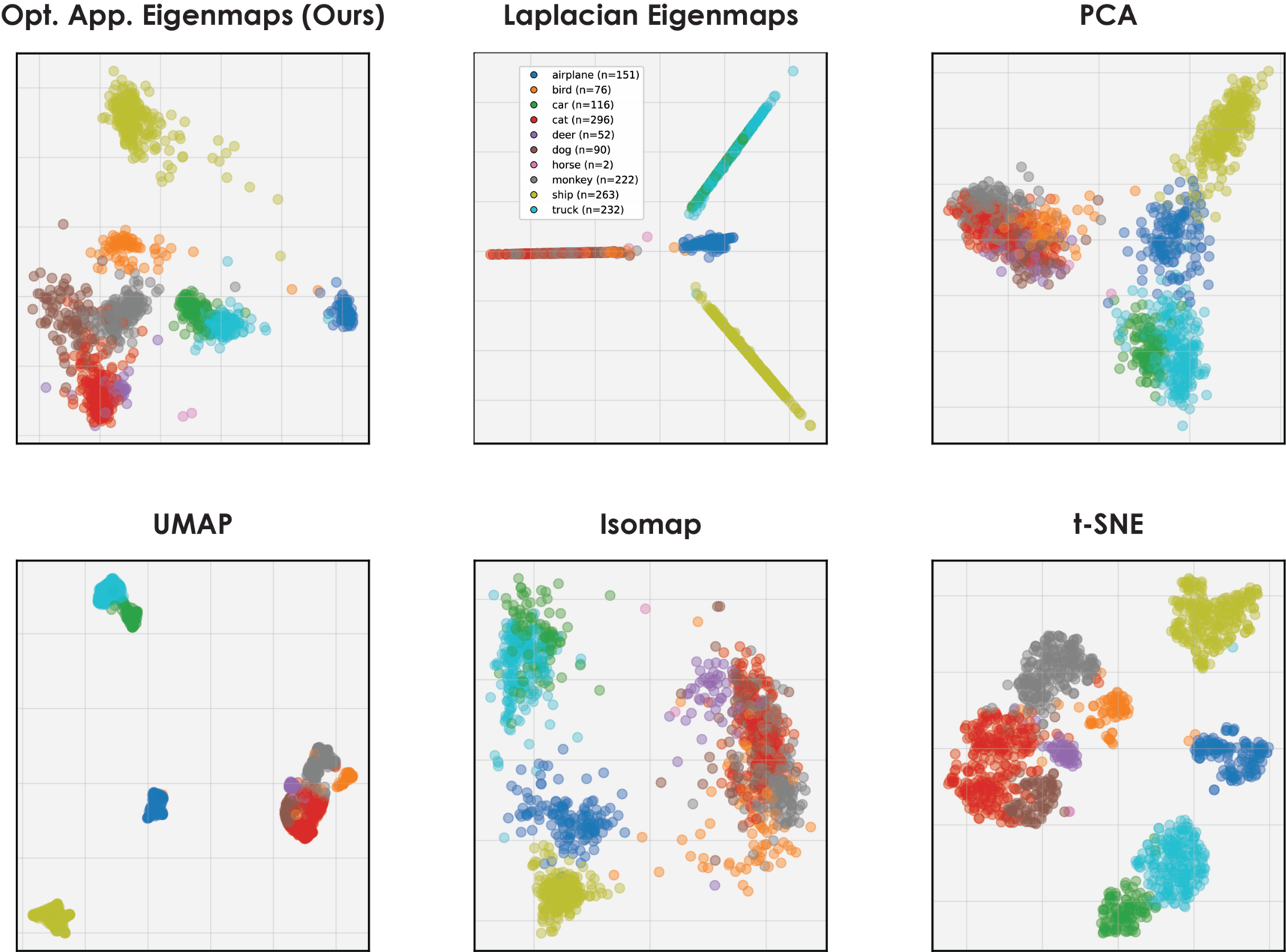}
    \caption{2D visualization of a random subset of the STL-10 dataset using DINOv2 embeddings. 
    }
    \label{fig: supp_stl10_dino_cluster1}
\end{figure*}

\begin{figure*}[t]
    \centering
    \includegraphics[width=0.72\textwidth]{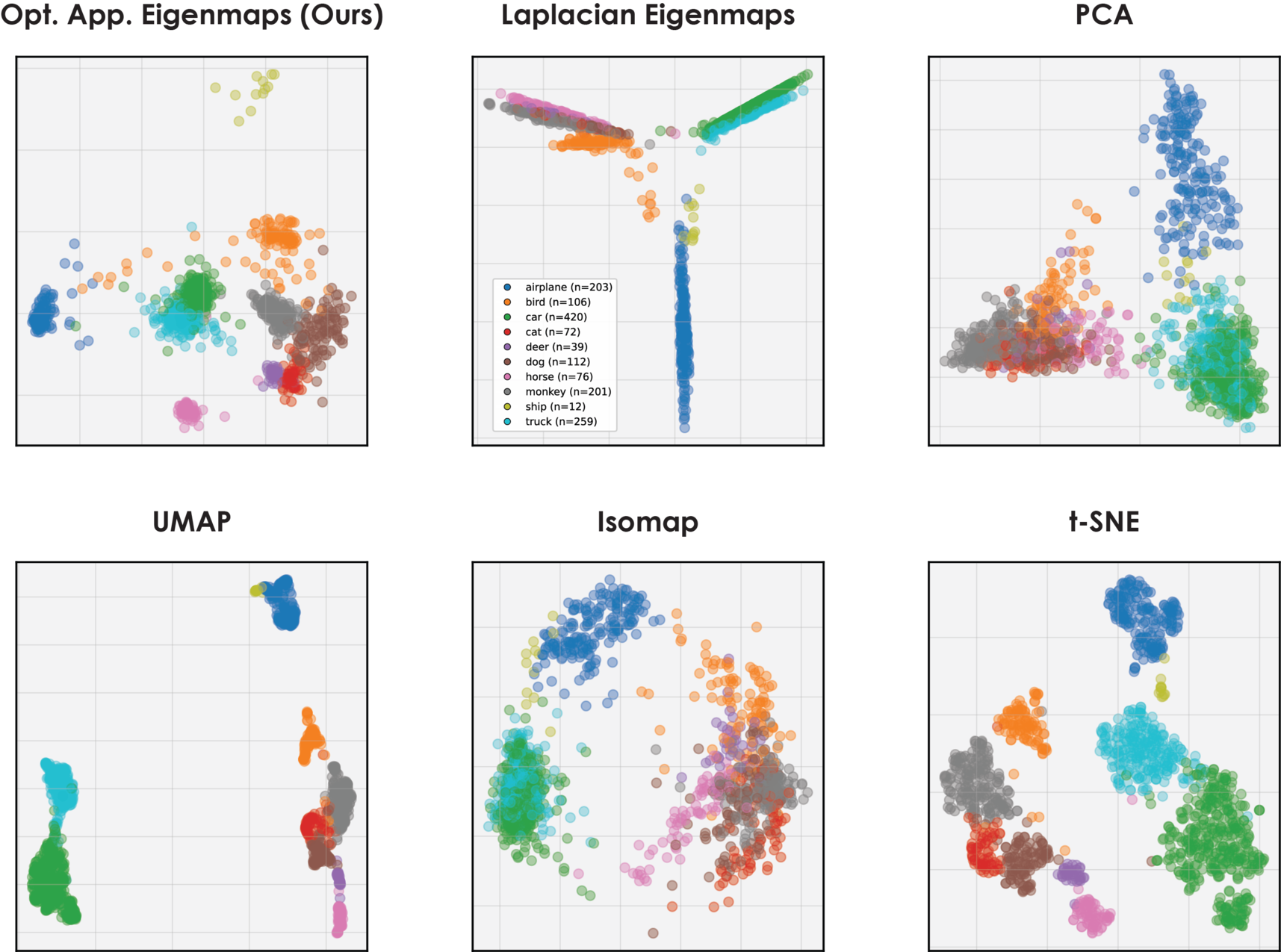}
    \caption{2D visualization of another random subset of the STL-10 dataset using DINOv2 embeddings. 
    }
    \label{fig: supp_stl10_dino_cluster2}
\end{figure*}

\paragraph{Qualitative Analysis.}
In \cref{fig: supp_imagenette_clip_cluster1,fig: supp_imagenette_clip_cluster2,fig: supp_imagenette_dino_cluster1,fig: supp_imagenette_dino_cluster2,fig: supp_stl10_clip_cluster1,fig: supp_stl10_clip_cluster2,fig: supp_stl10_dino_cluster1,fig: supp_stl10_dino_cluster2}, we plot the 2D (i.e.\ $k=2$) clustering results for Imagenette and STL-10 datasets with both CLIP and DINOv2 embeddings.
These visualizations demonstrate that our learned spectral basis captures meaningful semantic structure in the image manifolds, with distinct clusters corresponding to different object categories that are better clustered than in traditional methods like PCA and Isomap. In addition to accurate clusters, our embeddings provide smooth transitions between clusters, indicating that the learned basis successfully preserves the underlying smooth manifold geometry, unlike methods artificially overclustering the data like t-SNE and UMAP. This important observation demonstrates that our method is in fact superior to modern references t-SNE and UMAP, which cluster very well as revealed by the quantitative results but poorly preserve the manifold structure. Our visualisations are significantly better than those of Laplacian Eigenmaps, the vanilla analog of our method using the Graph Laplacian eigenfunctions instead of our Optimal Approximation Operator's eigenfunctions. Indeed, in Laplacian Eigenmaps clusters are ill-shaped and artificially separated unlike in our method.


\newpage



\clearpage  

\clearpage  

\begin{table*}[b]
    \caption{
    Average cosine similarity between predicted and oracle
    eigenfunctions at different truncation levels $k$ on each evaluation dataset (generalization setting).
    }
    \label{tab: 3D generalisation inference results cosine sim}
    \centering
    \resizebox{\textwidth}{!}{%
    \begin{tabular}{l l cccccccccc}
        \toprule
        \textbf{Dataset} & \textbf{\#Shapes} & {$\boldsymbol{k\leq 5}$} & {$\boldsymbol{k\leq 10}$} & {$\boldsymbol{k\leq 15}$} & {$\boldsymbol{k\leq 20}$} & {$\boldsymbol{k\leq 25}$} & {$\boldsymbol{k\leq 30}$} & {$\boldsymbol{k\leq 35}$} & {$\boldsymbol{k\leq 40}$} & {$\boldsymbol{k\leq 45}$} & {$\boldsymbol{k\leq 50}$} \\
        \midrule
        MPZ14 & 105 & 0.735 & 0.612 & 0.526 & 0.473 & 0.429 & 0.393 & 0.365 & 0.341 & 0.320 & 0.301 \\
        SHREC'07 & 380 & 0.729 & 0.610 & 0.535 & 0.480 & 0.438 & 0.403 & 0.374 & 0.349 & 0.327 & 0.308 \\
        SHREC'14 & 700 & 0.793 & 0.666 & 0.585 & 0.532 & 0.478 & 0.438 & 0.404 & 0.377 & 0.352 & 0.329 \\
        SHREC'15 & 1200 & 0.693 & 0.570 & 0.511 & 0.458 & 0.417 & 0.384 & 0.356 & 0.331 & 0.310 & 0.291 \\
        SHREC'19 & 50 & 0.750 & 0.633 & 0.556 & 0.495 & 0.448 & 0.409 & 0.381 & 0.357 & 0.334 & 0.315 \\
        SHREC'20-GR
        & 220 & 0.670 & 0.514 & 0.435 & 0.385 & 0.348 & 0.321 & 0.296 & 0.277 & 0.261 & 0.247 \\
        SHREC'20-NI & 14 & 0.598 & 0.536 & 0.495 & 0.446 & 0.404 & 0.375 & 0.348 & 0.326 & 0.307 & 0.291 \\
        DefTransfer & 278 & 0.621 & 0.500 & 0.425 & 0.386 & 0.352 & 0.326 & 0.303 & 0.285 & 0.269 & 0.254 \\
        TopKids & 26 & 0.788 & 0.656 & 0.551 & 0.486 & 0.438 & 0.402 & 0.374 & 0.352 & 0.331 & 0.312 \\
        \bottomrule
    \end{tabular}%
    }
\end{table*}

\clearpage

\begin{table*}[htbp]
\centering
\caption{Average manifold learning results on STL10 with DINOv2 embeddings. The mean and standard deviation over 32 runs is provided. The best and second best results are in bold and underlined respectively.}
\label{tab:dimred_stl10_dino}
\adjustbox{max width=\textwidth}{
\begin{tabular}{llccccccc}
\toprule
\multicolumn{9}{c}{\textbf{STL10 - DINOv2}} \\
\midrule
Method & $k$ & NMI & ARI & Comp. & AMI & Homo. & V-Meas. & FMI \\
\midrule
Opt. App. Eigenmaps (Ours) & 2 & \underline{0.790 $\pm$ 0.083} & \textbf{0.712 $\pm$ 0.144} & \textbf{0.752 $\pm$ 0.094} & \underline{0.787 $\pm$ 0.084} & 0.834 $\pm$ 0.080 & \underline{0.790 $\pm$ 0.083} & \textbf{0.775 $\pm$ 0.110} \\
Laplacian Eigenmaps & 2 & 0.594 $\pm$ 0.075 & 0.382 $\pm$ 0.096 & 0.568 $\pm$ 0.058 & 0.589 $\pm$ 0.076 & 0.627 $\pm$ 0.111 & 0.594 $\pm$ 0.075 & 0.510 $\pm$ 0.095 \\
PCA & 2 & 0.668 $\pm$ 0.046 & 0.475 $\pm$ 0.073 & 0.611 $\pm$ 0.041 & 0.664 $\pm$ 0.046 & 0.740 $\pm$ 0.075 & 0.668 $\pm$ 0.046 & 0.579 $\pm$ 0.076 \\
UMAP & 2 & \textbf{0.808 $\pm$ 0.058} & \underline{0.633 $\pm$ 0.119} & \underline{0.741 $\pm$ 0.082} & \textbf{0.806 $\pm$ 0.059} & \textbf{0.894 $\pm$ 0.038} & \textbf{0.808 $\pm$ 0.058} & \underline{0.717 $\pm$ 0.081} \\
t-SNE & 2 & 0.789 $\pm$ 0.058 & 0.595 $\pm$ 0.106 & 0.718 $\pm$ 0.080 & 0.786 $\pm$ 0.058 & \underline{0.879 $\pm$ 0.033} & 0.789 $\pm$ 0.058 & 0.686 $\pm$ 0.070 \\
Isomap & 2 & 0.684 $\pm$ 0.052 & 0.520 $\pm$ 0.092 & 0.632 $\pm$ 0.062 & 0.681 $\pm$ 0.053 & 0.751 $\pm$ 0.066 & 0.684 $\pm$ 0.052 & 0.618 $\pm$ 0.074 \\
\midrule
Opt. App. Eigenmaps (Ours) & 5 & \textbf{0.859 $\pm$ 0.061} & \textbf{0.812 $\pm$ 0.120} & \textbf{0.823 $\pm$ 0.081} & \textbf{0.858 $\pm$ 0.061} & \textbf{0.902 $\pm$ 0.047} & \textbf{0.859 $\pm$ 0.061} & \textbf{0.856 $\pm$ 0.088} \\
Laplacian Eigenmaps & 5 & 0.815 $\pm$ 0.054 & \underline{0.734 $\pm$ 0.130} & \underline{0.793 $\pm$ 0.063} & 0.813 $\pm$ 0.055 & 0.842 $\pm$ 0.065 & 0.815 $\pm$ 0.054 & \underline{0.792 $\pm$ 0.104} \\
PCA & 5 & 0.746 $\pm$ 0.053 & 0.565 $\pm$ 0.092 & 0.682 $\pm$ 0.073 & 0.743 $\pm$ 0.053 & 0.828 $\pm$ 0.043 & 0.746 $\pm$ 0.053 & 0.658 $\pm$ 0.061 \\
UMAP & 5 & \underline{0.821 $\pm$ 0.051} & 0.666 $\pm$ 0.109 & 0.757 $\pm$ 0.076 & \underline{0.819 $\pm$ 0.051} & \underline{0.901 $\pm$ 0.031} & \underline{0.821 $\pm$ 0.051} & 0.743 $\pm$ 0.074 \\
t-SNE & 5 & 0.796 $\pm$ 0.060 & 0.616 $\pm$ 0.112 & 0.727 $\pm$ 0.083 & 0.793 $\pm$ 0.060 & 0.884 $\pm$ 0.037 & 0.796 $\pm$ 0.060 & 0.704 $\pm$ 0.075 \\
Isomap & 5 & 0.772 $\pm$ 0.047 & 0.626 $\pm$ 0.104 & 0.711 $\pm$ 0.071 & 0.769 $\pm$ 0.048 & 0.849 $\pm$ 0.037 & 0.772 $\pm$ 0.047 & 0.709 $\pm$ 0.069 \\
\midrule
Opt. App. Eigenmaps (Ours) & 10 & \textbf{0.892 $\pm$ 0.047} & \textbf{0.863 $\pm$ 0.098} & \textbf{0.862 $\pm$ 0.071} & \textbf{0.891 $\pm$ 0.048} & \textbf{0.927 $\pm$ 0.029} & \textbf{0.892 $\pm$ 0.047} & \textbf{0.896 $\pm$ 0.071} \\
Laplacian Eigenmaps & 10 & \underline{0.828 $\pm$ 0.077} & \underline{0.768 $\pm$ 0.124} & \underline{0.835 $\pm$ 0.068} & \underline{0.826 $\pm$ 0.077} & 0.827 $\pm$ 0.101 & \underline{0.828 $\pm$ 0.077} & \underline{0.829 $\pm$ 0.073} \\
PCA & 10 & 0.788 $\pm$ 0.045 & 0.620 $\pm$ 0.097 & 0.721 $\pm$ 0.068 & 0.785 $\pm$ 0.046 & 0.873 $\pm$ 0.036 & 0.788 $\pm$ 0.045 & 0.706 $\pm$ 0.065 \\
UMAP & 10 & 0.821 $\pm$ 0.051 & 0.665 $\pm$ 0.109 & 0.758 $\pm$ 0.076 & 0.819 $\pm$ 0.051 & \underline{0.901 $\pm$ 0.031} & 0.821 $\pm$ 0.051 & 0.742 $\pm$ 0.074 \\
t-SNE & 10 & 0.810 $\pm$ 0.051 & 0.643 $\pm$ 0.108 & 0.743 $\pm$ 0.077 & 0.808 $\pm$ 0.051 & 0.896 $\pm$ 0.030 & 0.810 $\pm$ 0.051 & 0.726 $\pm$ 0.071 \\
Isomap & 10 & 0.794 $\pm$ 0.050 & 0.656 $\pm$ 0.101 & 0.733 $\pm$ 0.072 & 0.791 $\pm$ 0.050 & 0.870 $\pm$ 0.037 & 0.794 $\pm$ 0.050 & 0.733 $\pm$ 0.068 \\
\bottomrule
\end{tabular}
}
\end{table*}

\begin{table*}[htbp]
\centering
\caption{Average manifold learning results on STL10 with CLIP embeddings. The mean and standard deviation over 32 runs is provided. The best and second best results are in bold and underlined respectively.}
\label{tab:dimred_stl10_clip}
\adjustbox{max width=\textwidth}{
\begin{tabular}{llccccccc}
\toprule
\multicolumn{9}{c}{\textbf{STL10 - CLIP}} \\
\midrule
Method & $k$ & NMI & ARI & Comp. & AMI & Homo. & V-Meas. & FMI \\
\midrule
Opt. App. Eigenmaps (Ours) & 2 & 0.691 $\pm$ 0.071 & 0.525 $\pm$ 0.108 & 0.653 $\pm$ 0.071 & 0.687 $\pm$ 0.072 & 0.736 $\pm$ 0.085 & 0.691 $\pm$ 0.071 & 0.622 $\pm$ 0.093 \\
Laplacian Eigenmaps & 2 & 0.585 $\pm$ 0.071 & 0.365 $\pm$ 0.109 & 0.558 $\pm$ 0.056 & 0.580 $\pm$ 0.073 & 0.621 $\pm$ 0.112 & 0.585 $\pm$ 0.071 & 0.493 $\pm$ 0.110 \\
PCA & 2 & 0.673 $\pm$ 0.041 & 0.515 $\pm$ 0.071 & 0.615 $\pm$ 0.043 & 0.669 $\pm$ 0.042 & 0.746 $\pm$ 0.063 & 0.673 $\pm$ 0.041 & 0.613 $\pm$ 0.069 \\
UMAP & 2 & \textbf{0.835 $\pm$ 0.050} & \textbf{0.676 $\pm$ 0.111} & \textbf{0.770 $\pm$ 0.078} & \textbf{0.833 $\pm$ 0.051} & \textbf{0.915 $\pm$ 0.028} & \textbf{0.835 $\pm$ 0.050} & \textbf{0.752 $\pm$ 0.075} \\
t-SNE & 2 & \underline{0.801 $\pm$ 0.056} & \underline{0.614 $\pm$ 0.110} & \underline{0.730 $\pm$ 0.080} & \underline{0.798 $\pm$ 0.057} & \underline{0.892 $\pm$ 0.032} & \underline{0.801 $\pm$ 0.056} & \underline{0.703 $\pm$ 0.072} \\
Isomap & 2 & 0.718 $\pm$ 0.053 & 0.590 $\pm$ 0.113 & 0.667 $\pm$ 0.065 & 0.715 $\pm$ 0.053 & 0.781 $\pm$ 0.060 & 0.718 $\pm$ 0.053 & 0.675 $\pm$ 0.095 \\
\midrule
Opt. App. Eigenmaps (Ours) & 5 & \underline{0.820 $\pm$ 0.060} & \textbf{0.741 $\pm$ 0.120} & \textbf{0.795 $\pm$ 0.072} & \underline{0.818 $\pm$ 0.060} & 0.850 $\pm$ 0.063 & \underline{0.820 $\pm$ 0.060} & \textbf{0.797 $\pm$ 0.094} \\
Laplacian Eigenmaps & 5 & 0.796 $\pm$ 0.042 & \underline{0.716 $\pm$ 0.107} & 0.774 $\pm$ 0.063 & 0.794 $\pm$ 0.043 & 0.823 $\pm$ 0.045 & 0.796 $\pm$ 0.042 & \underline{0.775 $\pm$ 0.095} \\
PCA & 5 & 0.760 $\pm$ 0.043 & 0.593 $\pm$ 0.091 & 0.694 $\pm$ 0.061 & 0.757 $\pm$ 0.044 & 0.844 $\pm$ 0.041 & 0.760 $\pm$ 0.043 & 0.682 $\pm$ 0.066 \\
UMAP & 5 & \textbf{0.838 $\pm$ 0.049} & 0.694 $\pm$ 0.103 & \underline{0.777 $\pm$ 0.074} & \textbf{0.837 $\pm$ 0.050} & \textbf{0.914 $\pm$ 0.028} & \textbf{0.838 $\pm$ 0.049} & 0.765 $\pm$ 0.070 \\
t-SNE & 5 & 0.803 $\pm$ 0.064 & 0.632 $\pm$ 0.118 & 0.735 $\pm$ 0.088 & 0.801 $\pm$ 0.064 & \underline{0.890 $\pm$ 0.036} & 0.803 $\pm$ 0.064 & 0.717 $\pm$ 0.080 \\
Isomap & 5 & 0.796 $\pm$ 0.041 & 0.672 $\pm$ 0.102 & 0.741 $\pm$ 0.063 & 0.794 $\pm$ 0.042 & 0.865 $\pm$ 0.032 & 0.796 $\pm$ 0.041 & 0.745 $\pm$ 0.077 \\
\midrule
Opt. App. Eigenmaps (Ours) & 10 & \textbf{0.863 $\pm$ 0.047} & \textbf{0.818 $\pm$ 0.086} & \underline{0.845 $\pm$ 0.063} & \textbf{0.861 $\pm$ 0.047} & 0.884 $\pm$ 0.047 & \textbf{0.863 $\pm$ 0.047} & \textbf{0.858 $\pm$ 0.067} \\
Laplacian Eigenmaps & 10 & 0.796 $\pm$ 0.131 & \underline{0.720 $\pm$ 0.166} & \textbf{0.850 $\pm$ 0.055} & 0.794 $\pm$ 0.133 & 0.770 $\pm$ 0.161 & 0.796 $\pm$ 0.131 & \underline{0.803 $\pm$ 0.086} \\
PCA & 10 & 0.788 $\pm$ 0.051 & 0.628 $\pm$ 0.103 & 0.723 $\pm$ 0.072 & 0.785 $\pm$ 0.051 & 0.870 $\pm$ 0.035 & 0.788 $\pm$ 0.051 & 0.711 $\pm$ 0.073 \\
UMAP & 10 & \underline{0.833 $\pm$ 0.049} & 0.685 $\pm$ 0.103 & 0.772 $\pm$ 0.074 & \underline{0.832 $\pm$ 0.049} & \textbf{0.910 $\pm$ 0.029} & \underline{0.833 $\pm$ 0.049} & 0.758 $\pm$ 0.070 \\
t-SNE & 10 & 0.811 $\pm$ 0.066 & 0.650 $\pm$ 0.118 & 0.745 $\pm$ 0.090 & 0.809 $\pm$ 0.067 & \underline{0.894 $\pm$ 0.038} & 0.811 $\pm$ 0.066 & 0.731 $\pm$ 0.080 \\
Isomap & 10 & 0.814 $\pm$ 0.047 & 0.697 $\pm$ 0.111 & 0.759 $\pm$ 0.069 & 0.812 $\pm$ 0.048 & 0.881 $\pm$ 0.031 & 0.814 $\pm$ 0.047 & 0.765 $\pm$ 0.082 \\
\bottomrule
\end{tabular}
}
\end{table*}

\clearpage

\begin{table*}[htbp]
\centering
\caption{Average manifold learning results on Imagenette with DINOv2 embeddings. The mean and standard deviation over 32 runs is provided. The best and second best results are in bold and underlined respectively.}
\label{tab:dimred_imagenette_dino}
\adjustbox{max width=\textwidth}{
\begin{tabular}{llccccccc}
\toprule
\multicolumn{9}{c}{\textbf{Imagenette - DINOv2}} \\
\midrule
Method & $k$ & NMI & ARI & Comp. & AMI & Homo. & V-Meas. & FMI \\
\midrule
Opt. App. Eigenmaps (Ours) & 2 & 0.684 $\pm$ 0.061 & 0.602 $\pm$ 0.114 & 0.655 $\pm$ 0.070 & 0.680 $\pm$ 0.062 & 0.716 $\pm$ 0.053 & 0.684 $\pm$ 0.061 & 0.667 $\pm$ 0.098 \\
Laplacian Eigenmaps & 2 & 0.642 $\pm$ 0.062 & 0.456 $\pm$ 0.102 & 0.627 $\pm$ 0.048 & 0.638 $\pm$ 0.063 & 0.659 $\pm$ 0.082 & 0.642 $\pm$ 0.062 & 0.550 $\pm$ 0.091 \\
PCA & 2 & 0.645 $\pm$ 0.075 & 0.486 $\pm$ 0.126 & 0.617 $\pm$ 0.062 & 0.641 $\pm$ 0.077 & 0.677 $\pm$ 0.092 & 0.645 $\pm$ 0.075 & 0.568 $\pm$ 0.113 \\
UMAP & 2 & \textbf{0.875 $\pm$ 0.035} & \textbf{0.802 $\pm$ 0.076} & \textbf{0.836 $\pm$ 0.053} & \textbf{0.873 $\pm$ 0.036} & \textbf{0.918 $\pm$ 0.023} & \textbf{0.875 $\pm$ 0.035} & \textbf{0.839 $\pm$ 0.060} \\
t-SNE & 2 & \underline{0.829 $\pm$ 0.036} & \underline{0.716 $\pm$ 0.072} & \underline{0.782 $\pm$ 0.045} & \underline{0.828 $\pm$ 0.036} & \underline{0.884 $\pm$ 0.036} & \underline{0.829 $\pm$ 0.036} & \underline{0.767 $\pm$ 0.058} \\
Isomap & 2 & 0.679 $\pm$ 0.045 & 0.599 $\pm$ 0.090 & 0.646 $\pm$ 0.041 & 0.676 $\pm$ 0.046 & 0.717 $\pm$ 0.058 & 0.679 $\pm$ 0.045 & 0.665 $\pm$ 0.080 \\
\midrule
Opt. App. Eigenmaps (Ours) & 5 & \underline{0.842 $\pm$ 0.043} & \textbf{0.844 $\pm$ 0.051} & \underline{0.818 $\pm$ 0.058} & \underline{0.840 $\pm$ 0.044} & 0.868 $\pm$ 0.032 & \underline{0.842 $\pm$ 0.043} & \textbf{0.872 $\pm$ 0.041} \\
Laplacian Eigenmaps & 5 & 0.800 $\pm$ 0.069 & 0.703 $\pm$ 0.137 & 0.787 $\pm$ 0.071 & 0.798 $\pm$ 0.070 & 0.817 $\pm$ 0.081 & 0.800 $\pm$ 0.069 & 0.754 $\pm$ 0.117 \\
PCA & 5 & 0.759 $\pm$ 0.048 & 0.629 $\pm$ 0.080 & 0.718 $\pm$ 0.049 & 0.756 $\pm$ 0.048 & 0.806 $\pm$ 0.056 & 0.759 $\pm$ 0.048 & 0.691 $\pm$ 0.070 \\
UMAP & 5 & \textbf{0.881 $\pm$ 0.038} & \underline{0.816 $\pm$ 0.086} & \textbf{0.845 $\pm$ 0.058} & \textbf{0.880 $\pm$ 0.039} & \textbf{0.922 $\pm$ 0.021} & \textbf{0.881 $\pm$ 0.038} & \underline{0.850 $\pm$ 0.068} \\
t-SNE & 5 & 0.825 $\pm$ 0.029 & 0.725 $\pm$ 0.059 & 0.779 $\pm$ 0.040 & 0.823 $\pm$ 0.029 & \underline{0.879 $\pm$ 0.030} & 0.825 $\pm$ 0.029 & 0.775 $\pm$ 0.046 \\
Isomap & 5 & 0.836 $\pm$ 0.039 & 0.790 $\pm$ 0.077 & 0.802 $\pm$ 0.049 & 0.834 $\pm$ 0.040 & 0.875 $\pm$ 0.045 & 0.836 $\pm$ 0.039 & 0.827 $\pm$ 0.065 \\
\midrule
Opt. App. Eigenmaps (Ours) & 10 & 0.858 $\pm$ 0.046 & \textbf{0.856 $\pm$ 0.055} & \underline{0.832 $\pm$ 0.060} & 0.857 $\pm$ 0.046 & 0.888 $\pm$ 0.037 & 0.858 $\pm$ 0.046 & \textbf{0.881 $\pm$ 0.044} \\
Laplacian Eigenmaps & 10 & 0.752 $\pm$ 0.092 & 0.577 $\pm$ 0.145 & 0.820 $\pm$ 0.061 & 0.749 $\pm$ 0.092 & 0.706 $\pm$ 0.122 & 0.752 $\pm$ 0.092 & 0.685 $\pm$ 0.091 \\
PCA & 10 & 0.795 $\pm$ 0.047 & 0.676 $\pm$ 0.080 & 0.754 $\pm$ 0.052 & 0.793 $\pm$ 0.047 & 0.841 $\pm$ 0.052 & 0.795 $\pm$ 0.047 & 0.732 $\pm$ 0.067 \\
UMAP & 10 & \textbf{0.882 $\pm$ 0.038} & \underline{0.818 $\pm$ 0.086} & \textbf{0.846 $\pm$ 0.058} & \textbf{0.881 $\pm$ 0.039} & \textbf{0.923 $\pm$ 0.021} & \textbf{0.882 $\pm$ 0.038} & 0.851 $\pm$ 0.068 \\
t-SNE & 10 & 0.820 $\pm$ 0.036 & 0.721 $\pm$ 0.068 & 0.776 $\pm$ 0.047 & 0.818 $\pm$ 0.037 & 0.871 $\pm$ 0.034 & 0.820 $\pm$ 0.036 & 0.771 $\pm$ 0.055 \\
Isomap & 10 & \underline{0.862 $\pm$ 0.043} & 0.818 $\pm$ 0.093 & 0.828 $\pm$ 0.063 & \underline{0.860 $\pm$ 0.043} & \underline{0.901 $\pm$ 0.030} & \underline{0.862 $\pm$ 0.043} & \underline{0.851 $\pm$ 0.073} \\
\bottomrule
\end{tabular}
}
\end{table*}

\begin{table*}[htbp]
\centering
\caption{Average manifold learning results on Imagenette with CLIP embeddings. The mean and standard deviation over 32 runs is provided. The best and second best results are in bold and underlined respectively.}
\label{tab:dimred_imagenette_clip}
\adjustbox{max width=\textwidth}{
\begin{tabular}{llccccccc}
\toprule
\multicolumn{9}{c}{\textbf{Imagenette - CLIP}} \\
\midrule
Method & $k$ & NMI & ARI & Comp. & AMI & Homo. & V-Meas. & FMI \\
\midrule
Opt. App. Eigenmaps (Ours) & 2 & 0.746 $\pm$ 0.055 & 0.680 $\pm$ 0.083 & 0.712 $\pm$ 0.061 & 0.743 $\pm$ 0.056 & 0.784 $\pm$ 0.056 & 0.746 $\pm$ 0.055 & 0.733 $\pm$ 0.069 \\
Laplacian Eigenmaps & 2 & 0.519 $\pm$ 0.109 & 0.281 $\pm$ 0.130 & 0.532 $\pm$ 0.083 & 0.513 $\pm$ 0.112 & 0.515 $\pm$ 0.140 & 0.519 $\pm$ 0.109 & 0.421 $\pm$ 0.104 \\
PCA & 2 & 0.682 $\pm$ 0.043 & 0.551 $\pm$ 0.073 & 0.642 $\pm$ 0.034 & 0.678 $\pm$ 0.044 & 0.728 $\pm$ 0.060 & 0.682 $\pm$ 0.043 & 0.623 $\pm$ 0.071 \\
UMAP & 2 & \textbf{0.895 $\pm$ 0.042} & \textbf{0.818 $\pm$ 0.095} & \textbf{0.855 $\pm$ 0.065} & \textbf{0.894 $\pm$ 0.042} & \textbf{0.941 $\pm$ 0.017} & \textbf{0.895 $\pm$ 0.042} & \textbf{0.853 $\pm$ 0.074} \\
t-SNE & 2 & \underline{0.876 $\pm$ 0.041} & \underline{0.782 $\pm$ 0.091} & \underline{0.831 $\pm$ 0.063} & \underline{0.875 $\pm$ 0.042} & \underline{0.929 $\pm$ 0.022} & \underline{0.876 $\pm$ 0.041} & \underline{0.823 $\pm$ 0.071} \\
Isomap & 2 & 0.755 $\pm$ 0.040 & 0.683 $\pm$ 0.074 & 0.721 $\pm$ 0.039 & 0.752 $\pm$ 0.040 & 0.794 $\pm$ 0.054 & 0.755 $\pm$ 0.040 & 0.735 $\pm$ 0.066 \\
\midrule
Opt. App. Eigenmaps (Ours) & 5 & \underline{0.882 $\pm$ 0.034} & \textbf{0.874 $\pm$ 0.055} & \underline{0.858 $\pm$ 0.046} & \underline{0.880 $\pm$ 0.034} & 0.908 $\pm$ 0.026 & \underline{0.882 $\pm$ 0.034} & \textbf{0.896 $\pm$ 0.044} \\
Laplacian Eigenmaps & 5 & 0.820 $\pm$ 0.058 & 0.707 $\pm$ 0.141 & 0.809 $\pm$ 0.053 & 0.818 $\pm$ 0.058 & 0.833 $\pm$ 0.074 & 0.820 $\pm$ 0.058 & 0.757 $\pm$ 0.119 \\
PCA & 5 & 0.818 $\pm$ 0.040 & 0.746 $\pm$ 0.080 & 0.774 $\pm$ 0.050 & 0.816 $\pm$ 0.040 & 0.869 $\pm$ 0.039 & 0.818 $\pm$ 0.040 & 0.792 $\pm$ 0.064 \\
UMAP & 5 & \textbf{0.898 $\pm$ 0.043} & 0.830 $\pm$ 0.094 & \textbf{0.859 $\pm$ 0.066} & \textbf{0.897 $\pm$ 0.043} & \textbf{0.942 $\pm$ 0.020} & \textbf{0.898 $\pm$ 0.043} & 0.862 $\pm$ 0.073 \\
t-SNE & 5 & 0.875 $\pm$ 0.042 & 0.795 $\pm$ 0.093 & 0.832 $\pm$ 0.065 & 0.874 $\pm$ 0.043 & \underline{0.925 $\pm$ 0.022} & 0.875 $\pm$ 0.042 & 0.834 $\pm$ 0.072 \\
Isomap & 5 & 0.876 $\pm$ 0.030 & \underline{0.844 $\pm$ 0.064} & 0.842 $\pm$ 0.046 & 0.874 $\pm$ 0.030 & 0.914 $\pm$ 0.027 & 0.876 $\pm$ 0.030 & \underline{0.872 $\pm$ 0.052} \\
\midrule
Opt. App. Eigenmaps (Ours) & 10 & \textbf{0.915 $\pm$ 0.028} & \textbf{0.917 $\pm$ 0.045} & \underline{0.895 $\pm$ 0.042} & \textbf{0.914 $\pm$ 0.029} & \underline{0.937 $\pm$ 0.017} & \textbf{0.915 $\pm$ 0.028} & \textbf{0.932 $\pm$ 0.036} \\
Laplacian Eigenmaps & 10 & 0.819 $\pm$ 0.167 & 0.707 $\pm$ 0.193 & \textbf{0.899 $\pm$ 0.141} & 0.817 $\pm$ 0.168 & 0.761 $\pm$ 0.174 & 0.819 $\pm$ 0.167 & 0.790 $\pm$ 0.108 \\
PCA & 10 & 0.855 $\pm$ 0.034 & 0.780 $\pm$ 0.077 & 0.811 $\pm$ 0.051 & 0.854 $\pm$ 0.034 & 0.907 $\pm$ 0.025 & 0.855 $\pm$ 0.034 & 0.821 $\pm$ 0.060 \\
UMAP & 10 & \underline{0.899 $\pm$ 0.041} & 0.830 $\pm$ 0.091 & 0.860 $\pm$ 0.064 & \underline{0.898 $\pm$ 0.042} & \textbf{0.943 $\pm$ 0.018} & \underline{0.899 $\pm$ 0.041} & 0.863 $\pm$ 0.071 \\
t-SNE & 10 & 0.876 $\pm$ 0.039 & 0.793 $\pm$ 0.084 & 0.832 $\pm$ 0.060 & 0.874 $\pm$ 0.040 & 0.926 $\pm$ 0.022 & 0.876 $\pm$ 0.039 & 0.832 $\pm$ 0.064 \\
Isomap & 10 & 0.895 $\pm$ 0.038 & \underline{0.852 $\pm$ 0.080} & 0.860 $\pm$ 0.059 & 0.894 $\pm$ 0.038 & 0.934 $\pm$ 0.020 & 0.895 $\pm$ 0.038 & \underline{0.880 $\pm$ 0.063} \\
\bottomrule
\end{tabular}
}
\end{table*}

\clearpage

\begin{table*}[htbp]
\centering
\caption{Average manifold learning results on CALTECH256 with DINOv2 embeddings. The mean and standard deviation over 32 runs is provided. The best and second best results are in bold and underlined respectively.}
\label{tab:dimred_caltech256_dino}
\adjustbox{max width=\textwidth}{
\begin{tabular}{llccccccc}
\toprule
\multicolumn{9}{c}{\textbf{CALTECH256 - DINOv2}} \\
\midrule
Method & $k$ & NMI & ARI & Comp. & AMI & Homo. & V-Meas. & FMI \\
\midrule
Opt. App. Eigenmaps (Ours) & 2 & 0.611 $\pm$ 0.027 & 0.078 $\pm$ 0.026 & 0.594 $\pm$ 0.030 & 0.196 $\pm$ 0.053 & 0.629 $\pm$ 0.024 & 0.611 $\pm$ 0.027 & 0.092 $\pm$ 0.031 \\
Laplacian Eigenmaps & 2 & 0.650 $\pm$ 0.036 & 0.121 $\pm$ 0.037 & 0.645 $\pm$ 0.038 & 0.311 $\pm$ 0.061 & 0.655 $\pm$ 0.036 & 0.650 $\pm$ 0.036 & 0.134 $\pm$ 0.036 \\
PCA & 2 & 0.603 $\pm$ 0.029 & 0.063 $\pm$ 0.020 & 0.583 $\pm$ 0.032 & 0.165 $\pm$ 0.050 & 0.626 $\pm$ 0.027 & 0.603 $\pm$ 0.029 & 0.077 $\pm$ 0.026 \\
UMAP & 2 & \underline{0.834 $\pm$ 0.014} & \textbf{0.448 $\pm$ 0.084} & \textbf{0.810 $\pm$ 0.020} & \underline{0.654 $\pm$ 0.048} & \underline{0.860 $\pm$ 0.013} & \underline{0.834 $\pm$ 0.014} & \textbf{0.476 $\pm$ 0.073} \\
t-SNE & 2 & \textbf{0.840 $\pm$ 0.015} & \underline{0.419 $\pm$ 0.071} & \underline{0.809 $\pm$ 0.021} & \textbf{0.660 $\pm$ 0.036} & \textbf{0.873 $\pm$ 0.011} & \textbf{0.840 $\pm$ 0.015} & \underline{0.457 $\pm$ 0.059} \\
Isomap & 2 & 0.655 $\pm$ 0.021 & 0.138 $\pm$ 0.045 & 0.633 $\pm$ 0.024 & 0.273 $\pm$ 0.063 & 0.678 $\pm$ 0.019 & 0.655 $\pm$ 0.021 & 0.155 $\pm$ 0.051 \\
\midrule
Opt. App. Eigenmaps (Ours) & 5 & 0.714 $\pm$ 0.020 & 0.246 $\pm$ 0.062 & 0.696 $\pm$ 0.024 & 0.411 $\pm$ 0.058 & 0.733 $\pm$ 0.019 & 0.714 $\pm$ 0.020 & 0.263 $\pm$ 0.067 \\
Laplacian Eigenmaps & 5 & 0.724 $\pm$ 0.024 & 0.209 $\pm$ 0.044 & 0.712 $\pm$ 0.026 & 0.445 $\pm$ 0.053 & 0.737 $\pm$ 0.026 & 0.724 $\pm$ 0.024 & 0.223 $\pm$ 0.042 \\
PCA & 5 & 0.696 $\pm$ 0.019 & 0.184 $\pm$ 0.039 & 0.671 $\pm$ 0.023 & 0.355 $\pm$ 0.054 & 0.722 $\pm$ 0.015 & 0.696 $\pm$ 0.019 & 0.206 $\pm$ 0.043 \\
UMAP & 5 & \textbf{0.843 $\pm$ 0.013} & \textbf{0.465 $\pm$ 0.081} & \textbf{0.820 $\pm$ 0.020} & \textbf{0.675 $\pm$ 0.039} & \textbf{0.868 $\pm$ 0.011} & \textbf{0.843 $\pm$ 0.013} & \textbf{0.493 $\pm$ 0.068} \\
t-SNE & 5 & \underline{0.832 $\pm$ 0.015} & \underline{0.394 $\pm$ 0.066} & \underline{0.800 $\pm$ 0.020} & \underline{0.642 $\pm$ 0.039} & \underline{0.866 $\pm$ 0.011} & \underline{0.832 $\pm$ 0.015} & \underline{0.432 $\pm$ 0.056} \\
Isomap & 5 & 0.754 $\pm$ 0.017 & 0.319 $\pm$ 0.067 & 0.732 $\pm$ 0.021 & 0.486 $\pm$ 0.056 & 0.778 $\pm$ 0.015 & 0.754 $\pm$ 0.017 & 0.341 $\pm$ 0.067 \\
\midrule
Opt. App. Eigenmaps (Ours) & 10 & 0.759 $\pm$ 0.016 & 0.322 $\pm$ 0.060 & 0.743 $\pm$ 0.021 & 0.508 $\pm$ 0.047 & 0.775 $\pm$ 0.014 & 0.759 $\pm$ 0.016 & 0.339 $\pm$ 0.062 \\
Laplacian Eigenmaps & 10 & 0.758 $\pm$ 0.016 & 0.257 $\pm$ 0.042 & 0.743 $\pm$ 0.021 & 0.508 $\pm$ 0.045 & 0.774 $\pm$ 0.015 & 0.758 $\pm$ 0.016 & 0.272 $\pm$ 0.039 \\
PCA & 10 & 0.747 $\pm$ 0.015 & 0.267 $\pm$ 0.049 & 0.721 $\pm$ 0.020 & 0.466 $\pm$ 0.052 & 0.776 $\pm$ 0.013 & 0.747 $\pm$ 0.015 & 0.293 $\pm$ 0.050 \\
UMAP & 10 & \textbf{0.843 $\pm$ 0.013} & \textbf{0.462 $\pm$ 0.083} & \textbf{0.819 $\pm$ 0.021} & \textbf{0.673 $\pm$ 0.040} & \textbf{0.868 $\pm$ 0.011} & \textbf{0.843 $\pm$ 0.013} & \textbf{0.490 $\pm$ 0.070} \\
t-SNE & 10 & \underline{0.823 $\pm$ 0.015} & 0.379 $\pm$ 0.066 & \underline{0.792 $\pm$ 0.021} & \underline{0.623 $\pm$ 0.039} & \underline{0.857 $\pm$ 0.011} & \underline{0.823 $\pm$ 0.015} & 0.416 $\pm$ 0.055 \\
Isomap & 10 & 0.799 $\pm$ 0.015 & \underline{0.409 $\pm$ 0.072} & 0.777 $\pm$ 0.019 & 0.582 $\pm$ 0.050 & 0.823 $\pm$ 0.012 & 0.799 $\pm$ 0.015 & \underline{0.433 $\pm$ 0.068} \\
\bottomrule
\end{tabular}
}
\end{table*}

\begin{table*}[htbp]
\centering
\caption{Average manifold learning results on CALTECH256 with CLIP embeddings. The mean and standard deviation over 32 runs is provided. The best and second best results are in bold and underlined respectively.}
\label{tab:dimred_caltech256_clip}
\adjustbox{max width=\textwidth}{
\begin{tabular}{llccccccc}
\toprule
\multicolumn{9}{c}{\textbf{CALTECH256 - CLIP}} \\
\midrule
Method & $k$ & NMI & ARI & Comp. & AMI & Homo. & V-Meas. & FMI \\
\midrule
Opt. App. Eigenmaps (Ours) & 2 & 0.596 $\pm$ 0.031 & 0.069 $\pm$ 0.030 & 0.578 $\pm$ 0.033 & 0.159 $\pm$ 0.050 & 0.615 $\pm$ 0.029 & 0.596 $\pm$ 0.031 & 0.083 $\pm$ 0.036 \\
Laplacian Eigenmaps & 2 & 0.600 $\pm$ 0.055 & 0.075 $\pm$ 0.032 & 0.599 $\pm$ 0.055 & 0.225 $\pm$ 0.076 & 0.602 $\pm$ 0.057 & 0.600 $\pm$ 0.055 & 0.088 $\pm$ 0.032 \\
PCA & 2 & 0.599 $\pm$ 0.031 & 0.062 $\pm$ 0.022 & 0.577 $\pm$ 0.034 & 0.153 $\pm$ 0.046 & 0.622 $\pm$ 0.028 & 0.599 $\pm$ 0.031 & 0.076 $\pm$ 0.028 \\
UMAP & 2 & \underline{0.844 $\pm$ 0.015} & \textbf{0.472 $\pm$ 0.094} & \underline{0.819 $\pm$ 0.021} & \underline{0.673 $\pm$ 0.051} & \underline{0.869 $\pm$ 0.013} & \underline{0.844 $\pm$ 0.015} & \underline{0.501 $\pm$ 0.081} \\
t-SNE & 2 & \textbf{0.862 $\pm$ 0.011} & \underline{0.465 $\pm$ 0.079} & \textbf{0.832 $\pm$ 0.019} & \textbf{0.709 $\pm$ 0.031} & \textbf{0.895 $\pm$ 0.007} & \textbf{0.862 $\pm$ 0.011} & \textbf{0.503 $\pm$ 0.064} \\
Isomap & 2 & 0.643 $\pm$ 0.020 & 0.115 $\pm$ 0.035 & 0.622 $\pm$ 0.023 & 0.249 $\pm$ 0.065 & 0.666 $\pm$ 0.018 & 0.643 $\pm$ 0.020 & 0.132 $\pm$ 0.041 \\
\midrule
Opt. App. Eigenmaps (Ours) & 5 & 0.685 $\pm$ 0.022 & 0.226 $\pm$ 0.068 & 0.666 $\pm$ 0.026 & 0.348 $\pm$ 0.055 & 0.704 $\pm$ 0.020 & 0.685 $\pm$ 0.022 & 0.243 $\pm$ 0.073 \\
Laplacian Eigenmaps & 5 & 0.688 $\pm$ 0.024 & 0.160 $\pm$ 0.029 & 0.685 $\pm$ 0.026 & 0.390 $\pm$ 0.051 & 0.692 $\pm$ 0.025 & 0.688 $\pm$ 0.024 & 0.173 $\pm$ 0.029 \\
PCA & 5 & 0.677 $\pm$ 0.020 & 0.165 $\pm$ 0.044 & 0.653 $\pm$ 0.022 & 0.315 $\pm$ 0.064 & 0.704 $\pm$ 0.019 & 0.677 $\pm$ 0.020 & 0.186 $\pm$ 0.052 \\
UMAP & 5 & \textbf{0.857 $\pm$ 0.013} & \textbf{0.501 $\pm$ 0.089} & \textbf{0.835 $\pm$ 0.020} & \textbf{0.704 $\pm$ 0.040} & \underline{0.881 $\pm$ 0.010} & \textbf{0.857 $\pm$ 0.013} & \textbf{0.528 $\pm$ 0.075} \\
t-SNE & 5 & \underline{0.853 $\pm$ 0.011} & \underline{0.430 $\pm$ 0.070} & \underline{0.821 $\pm$ 0.019} & \underline{0.686 $\pm$ 0.033} & \textbf{0.887 $\pm$ 0.007} & \underline{0.853 $\pm$ 0.011} & \underline{0.469 $\pm$ 0.056} \\
Isomap & 5 & 0.742 $\pm$ 0.017 & 0.291 $\pm$ 0.056 & 0.721 $\pm$ 0.021 & 0.463 $\pm$ 0.059 & 0.765 $\pm$ 0.015 & 0.742 $\pm$ 0.017 & 0.313 $\pm$ 0.056 \\
\midrule
Opt. App. Eigenmaps (Ours) & 10 & 0.734 $\pm$ 0.018 & 0.309 $\pm$ 0.071 & 0.718 $\pm$ 0.022 & 0.456 $\pm$ 0.050 & 0.750 $\pm$ 0.015 & 0.734 $\pm$ 0.018 & 0.325 $\pm$ 0.073 \\
Laplacian Eigenmaps & 10 & 0.735 $\pm$ 0.020 & 0.223 $\pm$ 0.038 & 0.728 $\pm$ 0.020 & 0.475 $\pm$ 0.048 & 0.742 $\pm$ 0.025 & 0.735 $\pm$ 0.020 & 0.236 $\pm$ 0.038 \\
PCA & 10 & 0.724 $\pm$ 0.014 & 0.237 $\pm$ 0.051 & 0.698 $\pm$ 0.018 & 0.414 $\pm$ 0.062 & 0.752 $\pm$ 0.013 & 0.724 $\pm$ 0.014 & 0.262 $\pm$ 0.056 \\
UMAP & 10 & \textbf{0.856 $\pm$ 0.014} & \textbf{0.494 $\pm$ 0.088} & \textbf{0.833 $\pm$ 0.021} & \textbf{0.702 $\pm$ 0.040} & \textbf{0.881 $\pm$ 0.011} & \textbf{0.856 $\pm$ 0.014} & \textbf{0.522 $\pm$ 0.074} \\
t-SNE & 10 & \underline{0.844 $\pm$ 0.012} & \underline{0.417 $\pm$ 0.070} & \underline{0.812 $\pm$ 0.019} & \underline{0.667 $\pm$ 0.036} & \underline{0.879 $\pm$ 0.008} & \underline{0.844 $\pm$ 0.012} & \underline{0.456 $\pm$ 0.057} \\
Isomap & 10 & 0.788 $\pm$ 0.013 & 0.381 $\pm$ 0.062 & 0.767 $\pm$ 0.018 & 0.560 $\pm$ 0.054 & 0.810 $\pm$ 0.012 & 0.788 $\pm$ 0.013 & 0.403 $\pm$ 0.059 \\
\bottomrule
\end{tabular}
}
\end{table*}

\clearpage

\begin{table*}[htbp]
\centering
\caption{Average manifold learning results on CIFAR100 with DINOv2 embeddings. The mean and standard deviation over 32 runs is provided. The best and second best results are in bold and underlined respectively.}
\label{tab:dimred_cifar100_dino}
\adjustbox{max width=\textwidth}{
\begin{tabular}{llccccccc}
\toprule
\multicolumn{9}{c}{\textbf{CIFAR100 - DINOv2}} \\
\midrule
Method & $k$ & NMI & ARI & Comp. & AMI & Homo. & V-Meas. & FMI \\
\midrule
Opt. App. Eigenmaps (Ours) & 2 & 0.568 $\pm$ 0.016 & 0.164 $\pm$ 0.036 & 0.576 $\pm$ 0.016 & 0.412 $\pm$ 0.046 & 0.561 $\pm$ 0.020 & 0.568 $\pm$ 0.016 & 0.191 $\pm$ 0.038 \\
Laplacian Eigenmaps & 2 & 0.590 $\pm$ 0.032 & 0.186 $\pm$ 0.037 & 0.577 $\pm$ 0.032 & 0.426 $\pm$ 0.050 & 0.603 $\pm$ 0.034 & 0.590 $\pm$ 0.032 & 0.206 $\pm$ 0.038 \\
PCA & 2 & 0.492 $\pm$ 0.017 & 0.098 $\pm$ 0.019 & 0.472 $\pm$ 0.020 & 0.273 $\pm$ 0.046 & 0.515 $\pm$ 0.015 & 0.492 $\pm$ 0.017 & 0.118 $\pm$ 0.024 \\
UMAP & 2 & \textbf{0.745 $\pm$ 0.016} & \textbf{0.433 $\pm$ 0.040} & \textbf{0.716 $\pm$ 0.018} & \textbf{0.635 $\pm$ 0.040} & \textbf{0.776 $\pm$ 0.021} & \textbf{0.745 $\pm$ 0.016} & \textbf{0.457 $\pm$ 0.043} \\
t-SNE & 2 & \underline{0.740 $\pm$ 0.014} & \underline{0.412 $\pm$ 0.038} & \underline{0.708 $\pm$ 0.016} & \underline{0.626 $\pm$ 0.038} & \underline{0.775 $\pm$ 0.018} & \underline{0.740 $\pm$ 0.014} & \underline{0.439 $\pm$ 0.041} \\
Isomap & 2 & 0.533 $\pm$ 0.018 & 0.142 $\pm$ 0.032 & 0.512 $\pm$ 0.020 & 0.332 $\pm$ 0.051 & 0.556 $\pm$ 0.018 & 0.533 $\pm$ 0.018 & 0.163 $\pm$ 0.036 \\
\midrule
Opt. App. Eigenmaps (Ours) & 5 & 0.583 $\pm$ 0.017 & 0.176 $\pm$ 0.029 & 0.602 $\pm$ 0.023 & 0.442 $\pm$ 0.034 & 0.565 $\pm$ 0.016 & 0.583 $\pm$ 0.017 & 0.208 $\pm$ 0.030 \\
Laplacian Eigenmaps & 5 & 0.640 $\pm$ 0.019 & 0.239 $\pm$ 0.036 & 0.622 $\pm$ 0.018 & 0.492 $\pm$ 0.047 & 0.659 $\pm$ 0.024 & 0.640 $\pm$ 0.019 & 0.258 $\pm$ 0.039 \\
PCA & 5 & 0.601 $\pm$ 0.013 & 0.206 $\pm$ 0.031 & 0.576 $\pm$ 0.015 & 0.428 $\pm$ 0.049 & 0.629 $\pm$ 0.015 & 0.601 $\pm$ 0.013 & 0.228 $\pm$ 0.036 \\
UMAP & 5 & \textbf{0.750 $\pm$ 0.016} & \textbf{0.447 $\pm$ 0.041} & \textbf{0.722 $\pm$ 0.018} & \textbf{0.644 $\pm$ 0.039} & \textbf{0.781 $\pm$ 0.020} & \textbf{0.750 $\pm$ 0.016} & \textbf{0.470 $\pm$ 0.043} \\
t-SNE & 5 & \underline{0.736 $\pm$ 0.015} & \underline{0.403 $\pm$ 0.036} & \underline{0.705 $\pm$ 0.017} & \underline{0.621 $\pm$ 0.037} & \underline{0.771 $\pm$ 0.018} & \underline{0.736 $\pm$ 0.015} & \underline{0.430 $\pm$ 0.039} \\
Isomap & 5 & 0.647 $\pm$ 0.017 & 0.291 $\pm$ 0.048 & 0.622 $\pm$ 0.018 & 0.495 $\pm$ 0.049 & 0.673 $\pm$ 0.020 & 0.647 $\pm$ 0.017 & 0.313 $\pm$ 0.052 \\
\midrule
Opt. App. Eigenmaps (Ours) & 10 & 0.613 $\pm$ 0.013 & 0.214 $\pm$ 0.039 & 0.622 $\pm$ 0.016 & 0.477 $\pm$ 0.036 & 0.605 $\pm$ 0.017 & 0.613 $\pm$ 0.013 & 0.238 $\pm$ 0.040 \\
Laplacian Eigenmaps & 10 & 0.669 $\pm$ 0.018 & 0.273 $\pm$ 0.038 & 0.649 $\pm$ 0.017 & 0.531 $\pm$ 0.048 & 0.690 $\pm$ 0.024 & 0.669 $\pm$ 0.018 & 0.292 $\pm$ 0.041 \\
PCA & 10 & 0.649 $\pm$ 0.018 & 0.268 $\pm$ 0.042 & 0.622 $\pm$ 0.017 & 0.495 $\pm$ 0.054 & 0.679 $\pm$ 0.023 & 0.649 $\pm$ 0.018 & 0.292 $\pm$ 0.047 \\
UMAP & 10 & \textbf{0.751 $\pm$ 0.015} & \textbf{0.448 $\pm$ 0.038} & \textbf{0.722 $\pm$ 0.017} & \textbf{0.644 $\pm$ 0.036} & \textbf{0.781 $\pm$ 0.018} & \textbf{0.751 $\pm$ 0.015} & \textbf{0.472 $\pm$ 0.040} \\
t-SNE & 10 & \underline{0.734 $\pm$ 0.019} & \underline{0.400 $\pm$ 0.044} & \underline{0.703 $\pm$ 0.020} & \underline{0.617 $\pm$ 0.044} & \underline{0.768 $\pm$ 0.023} & \underline{0.734 $\pm$ 0.019} & \underline{0.426 $\pm$ 0.047} \\
Isomap & 10 & 0.689 $\pm$ 0.018 & 0.355 $\pm$ 0.055 & 0.664 $\pm$ 0.017 & 0.556 $\pm$ 0.050 & 0.717 $\pm$ 0.023 & 0.689 $\pm$ 0.018 & 0.376 $\pm$ 0.059 \\
\bottomrule
\end{tabular}
}
\end{table*}

\begin{table*}[htbp]
\centering
\caption{Average manifold learning results on CIFAR100 with CLIP embeddings. The mean and standard deviation over 32 runs is provided. The best and second best results are in bold and underlined respectively.}
\label{tab:dimred_cifar100_clip}
\adjustbox{max width=\textwidth}{
\begin{tabular}{llccccccc}
\toprule
\multicolumn{9}{c}{\textbf{CIFAR100 - CLIP}} \\
\midrule
Method & $k$ & NMI & ARI & Comp. & AMI & Homo. & V-Meas. & FMI \\
\midrule
Opt. App. Eigenmaps (Ours) & 2 & 0.499 $\pm$ 0.021 & 0.111 $\pm$ 0.024 & 0.488 $\pm$ 0.024 & 0.300 $\pm$ 0.040 & 0.511 $\pm$ 0.020 & 0.499 $\pm$ 0.021 & 0.132 $\pm$ 0.027 \\
Laplacian Eigenmaps & 2 & 0.392 $\pm$ 0.064 & 0.055 $\pm$ 0.028 & 0.414 $\pm$ 0.068 & 0.208 $\pm$ 0.080 & 0.373 $\pm$ 0.063 & 0.392 $\pm$ 0.064 & 0.088 $\pm$ 0.035 \\
PCA & 2 & 0.423 $\pm$ 0.028 & 0.052 $\pm$ 0.012 & 0.406 $\pm$ 0.031 & 0.177 $\pm$ 0.036 & 0.442 $\pm$ 0.025 & 0.423 $\pm$ 0.028 & 0.072 $\pm$ 0.016 \\
UMAP & 2 & \underline{0.656 $\pm$ 0.022} & \textbf{0.320 $\pm$ 0.045} & \underline{0.628 $\pm$ 0.024} & \underline{0.506 $\pm$ 0.049} & \underline{0.686 $\pm$ 0.023} & \underline{0.656 $\pm$ 0.022} & \underline{0.345 $\pm$ 0.048} \\
t-SNE & 2 & \textbf{0.668 $\pm$ 0.021} & \underline{0.320 $\pm$ 0.038} & \textbf{0.639 $\pm$ 0.024} & \textbf{0.523 $\pm$ 0.044} & \textbf{0.701 $\pm$ 0.021} & \textbf{0.668 $\pm$ 0.021} & \textbf{0.346 $\pm$ 0.041} \\
Isomap & 2 & 0.464 $\pm$ 0.019 & 0.083 $\pm$ 0.019 & 0.448 $\pm$ 0.022 & 0.240 $\pm$ 0.044 & 0.481 $\pm$ 0.017 & 0.464 $\pm$ 0.019 & 0.103 $\pm$ 0.024 \\
\midrule
Opt. App. Eigenmaps (Ours) & 5 & 0.555 $\pm$ 0.019 & 0.159 $\pm$ 0.027 & 0.552 $\pm$ 0.022 & 0.388 $\pm$ 0.040 & 0.557 $\pm$ 0.020 & 0.555 $\pm$ 0.019 & 0.180 $\pm$ 0.030 \\
Laplacian Eigenmaps & 5 & 0.487 $\pm$ 0.051 & 0.101 $\pm$ 0.031 & 0.520 $\pm$ 0.047 & 0.333 $\pm$ 0.059 & 0.459 $\pm$ 0.056 & 0.487 $\pm$ 0.051 & 0.140 $\pm$ 0.032 \\
PCA & 5 & 0.504 $\pm$ 0.020 & 0.116 $\pm$ 0.026 & 0.481 $\pm$ 0.023 & 0.286 $\pm$ 0.048 & 0.528 $\pm$ 0.019 & 0.504 $\pm$ 0.020 & 0.138 $\pm$ 0.030 \\
UMAP & 5 & \textbf{0.667 $\pm$ 0.021} & \textbf{0.342 $\pm$ 0.041} & \textbf{0.641 $\pm$ 0.023} & \textbf{0.525 $\pm$ 0.044} & \underline{0.696 $\pm$ 0.021} & \textbf{0.667 $\pm$ 0.021} & \textbf{0.365 $\pm$ 0.045} \\
t-SNE & 5 & \underline{0.665 $\pm$ 0.021} & \underline{0.310 $\pm$ 0.039} & \underline{0.636 $\pm$ 0.023} & \underline{0.519 $\pm$ 0.044} & \textbf{0.698 $\pm$ 0.021} & \underline{0.665 $\pm$ 0.021} & \underline{0.336 $\pm$ 0.042} \\
Isomap & 5 & 0.561 $\pm$ 0.019 & 0.176 $\pm$ 0.036 & 0.541 $\pm$ 0.021 & 0.375 $\pm$ 0.054 & 0.582 $\pm$ 0.020 & 0.561 $\pm$ 0.019 & 0.197 $\pm$ 0.041 \\
\midrule
Opt. App. Eigenmaps (Ours) & 10 & 0.566 $\pm$ 0.021 & 0.173 $\pm$ 0.037 & 0.577 $\pm$ 0.023 & 0.415 $\pm$ 0.042 & 0.555 $\pm$ 0.023 & 0.566 $\pm$ 0.021 & 0.200 $\pm$ 0.037 \\
Laplacian Eigenmaps & 10 & 0.551 $\pm$ 0.031 & 0.137 $\pm$ 0.037 & 0.579 $\pm$ 0.025 & 0.407 $\pm$ 0.049 & 0.526 $\pm$ 0.039 & 0.551 $\pm$ 0.031 & 0.174 $\pm$ 0.036 \\
PCA & 10 & 0.557 $\pm$ 0.018 & 0.172 $\pm$ 0.035 & 0.532 $\pm$ 0.021 & 0.363 $\pm$ 0.047 & 0.584 $\pm$ 0.017 & 0.557 $\pm$ 0.018 & 0.195 $\pm$ 0.039 \\
UMAP & 10 & \textbf{0.667 $\pm$ 0.021} & \textbf{0.343 $\pm$ 0.045} & \textbf{0.641 $\pm$ 0.024} & \textbf{0.525 $\pm$ 0.044} & \textbf{0.696 $\pm$ 0.021} & \textbf{0.667 $\pm$ 0.021} & \textbf{0.366 $\pm$ 0.048} \\
t-SNE & 10 & \underline{0.653 $\pm$ 0.027} & \underline{0.292 $\pm$ 0.049} & \underline{0.625 $\pm$ 0.028} & \underline{0.501 $\pm$ 0.056} & \underline{0.685 $\pm$ 0.028} & \underline{0.653 $\pm$ 0.027} & \underline{0.318 $\pm$ 0.053} \\
Isomap & 10 & 0.588 $\pm$ 0.019 & 0.211 $\pm$ 0.039 & 0.567 $\pm$ 0.021 & 0.414 $\pm$ 0.051 & 0.611 $\pm$ 0.021 & 0.588 $\pm$ 0.019 & 0.232 $\pm$ 0.044 \\
\bottomrule
\end{tabular}
}
\end{table*}

\clearpage

\end{document}